%% file: main.tex
\pdfoutput=1
\documentclass{article}

\usepackage[dblblindworkshop, final]{neurips_2026}
\workshoptitle{PTA: From Pretrained Representations to Acting Agents}

\usepackage[utf8]{inputenc}
\usepackage[T1]{fontenc}
\input{math_commands.tex}

\usepackage{hyperref}
\usepackage{url}
\usepackage{booktabs}
\usepackage{amsfonts}
\usepackage{nicefrac}
\usepackage{microtype}
\usepackage{xcolor}
\usepackage{verbatim}
\usepackage{algorithm}
\usepackage{algorithmic}
\usepackage{graphicx}
\usepackage{subcaption}
\usepackage{amsmath}
\usepackage{wrapfig}
\usepackage{amssymb}
\usepackage{mathtools}
\usepackage{amsthm}
\usepackage{bm}
\usepackage{multirow}
\usepackage{tikz}

\definecolor{skyblue}{RGB}{30,60,130}

\newtheorem{proposition}{Proposition}
\newtheorem{theorem}{Theorem}
\newtheorem{lemma}{Lemma}

\title{Value-Guided Adaptive Action Chunking for Reinforcement Learning}

\author{%
  Yongjae Shin \quad Jongeui Park \quad Jongseong Chae \quad Seongmin Kim \quad Youngchul Sung \\
  KAIST \\
  \texttt{\{yongjae.shin, ycsung\}@kaist.ac.kr} \\
}

\begin{document}

\maketitle

\begin{abstract}
Action chunking emerged as a pivotal technique in imitation learning, enabling policies to predict cohesive action sequences rather than single actions.
Recently, this approach has expanded to reinforcement learning (RL), enhancing behavioral consistency and reducing bootstrapping errors in value function estimation.
However, existing methods rely on a fixed chunk length, creating a performance bottleneck as the optimal length varies across states and tasks.
In this paper, we propose \textbf{A}daptive Action \textbf{CH}unking (ACH), a novel offline-to-online RL algorithm that dynamically modulates chunk length during both training and inference. 
For each state, ACH takes the action value of each candidate chunk length as its optimality criterion.
We estimate the action values of all candidate chunk lengths simultaneously in a single forward pass by employing a causal Transformer-based value function.
Based on these values, the agent adaptively selects the most effective chunk length for the current state.
Evaluated on 34 challenging tasks, ACH outperforms fixed-length baselines in most environments, demonstrating enhanced learning efficiency.
Moreover, ACH  improves the performance of vision-language-action (VLA) models, demonstrating its applicability to large-scale models.
\end{abstract}

\section{Introduction}

Action chunking \citep{act} is a technique introduced in imitation learning (IL), where a policy predicts a sequence of consecutive actions, called an action chunk, instead of a single action per step.
Since an action chunk is executed in an open-loop manner, the policy makes decisions once every several steps rather than at every step, enabling efficient inference.
Moreover, while individually sampled single actions often result in redundant or inconsistent behaviors, an action chunk is generated as one cohesive sequence, thereby promoting behavioral consistency.
Driven by these advantages, action chunking has emerged as a pivotal technique within the imitation domain \citep{act,pi_0,diffusionpolicy,groot}.

\begin{figure}[t]
    \centering    \includegraphics[width=0.75\linewidth]{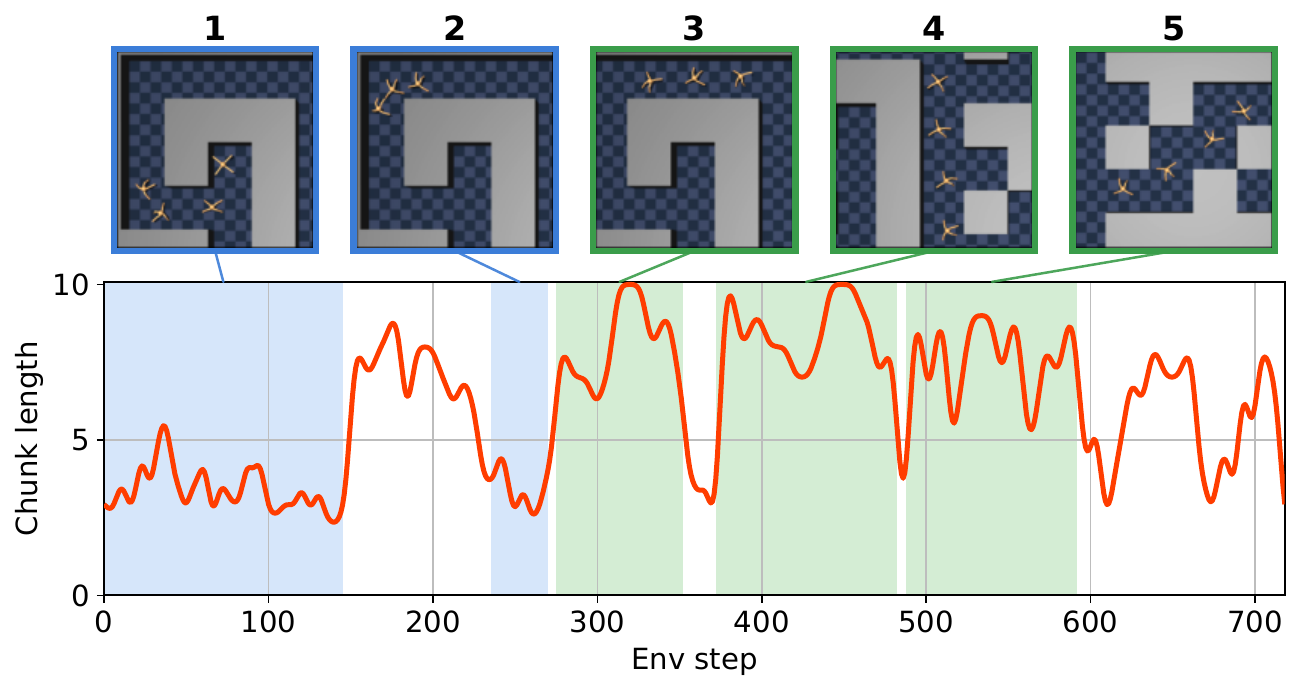}
    \vspace{-2ex}
    \caption{Adaptive Action Chunking: Chunk lengths selected by our proposed algorithm, ACH, over environment timesteps during an \texttt{antmaze-giant} episode. Shaded regions correspond to the subfigures above. ACH selects shorter chunks (blue) in segments where the required action changes rapidly, such as around corners, and longer chunks (green) where the trajectory is predictable. A similar pattern in a manipulation task is shown in Appendix \ref{appendix:manipulation}.}
    \label{fig:intro}
    \vspace{-3ex}
\end{figure}

Recently, action chunking has expanded beyond IL into the realm of reinforcement learning (RL) \citep{rl}.
Specifically, in offline RL \citep{offlinerl}, several methods have successfully integrated action chunking \citep{qc,dqc,deas,cgq}.
These approaches focus on effective value function training for policies with a fixed chunk length.
By aligning value estimation with action sequences rather than individual actions, these methods effectively mitigate the bootstrapping errors inherent in single-action value estimation.
Consequently, they demonstrate the advantage of action chunking in offline RL, particularly in long-horizon tasks.

Building on these advantages, action chunking has been further extended to offline-to-online (Off2On) RL \citep{off2on,pex,cal-ql,qc}.
In Off2On RL, a policy pre-trained via offline RL is fine-tuned through online interaction with the environment, allowing it to surpass the performance limit imposed by the quality of the given dataset.
In this setting, the behavioral consistency of action chunking enables efficient exploration during online fine-tuning, complementing its benefits in value function learning \citep{qc}.

However, these methods keep the chunk length fixed throughout training and inference, so their performance hinges on a single value that is rarely optimal for every task, or even for every state within a single task.
Consider, for example, the antmaze navigation task \citep{ogbench}, where an ant must traverse a complex, large-scale maze to reach a goal, as illustrated in Figure \ref{fig:intro}.
When the ant is in the segment shown in Subfigure 1, the appropriate action changes rapidly, so an excessively large chunk length may cause frequent collisions or make the agent overshoot turning points, thereby degrading performance. In such scenarios, a shorter chunk length is preferred.
Conversely, in the segment shown in Subfigure 4, where the trajectory is more predictable, the agent should ideally maintain its momentum to reach the goal efficiently.
In this case, a larger chunk length is more advantageous.
These variations underscore the need to move beyond rigid, fixed-length approaches toward a more adaptive action chunking strategy.

In this paper, we propose \textbf{A}daptive Action \textbf{CH}unking (ACH), a novel Off2On RL algorithm designed to address these limitations.
Our method enables the policy to optimally adjust its chunk length based on the current state, as illustrated in Figure \ref{fig:intro}.
As the criterion for this adjustment, we use the action value of each candidate chunk length, which naturally reflects the effectiveness of the corresponding chunk.
These values are estimated simultaneously in a single forward pass by employing a causal Transformer-based value function, allowing ACH to select the most suitable chunk length efficiently.
We evaluate our approach on 34 diverse and challenging tasks, where experimental results demonstrate that ACH successfully modulates chunk length across various environments, outperforming existing Off2On RL methods in most environments.
We further apply ACH to a vision-language-action (VLA) model, where it outperforms all baselines on two robot embodiments, showing that ACH extends to large-scale models.

\section{Preliminaries}
\label{sec:prelim}

\textbf{Offline-to-online RL} ~~
In this paper, we consider a Markov Decision Process (MDP) \citep{rl} defined as $\mathcal{M}=(\mathcal{S, A}, r, \rho, \mathcal{P}, \gamma)$, where $\mathcal{S}$ denotes the state space, $\mathcal{A}$ denotes the action space, $r(s,a):\mathcal{S}\times\mathcal{A}\rightarrow\mathbb{R}$ represents the reward function, $\rho\in\Delta(\mathcal{S})$ is the initial state distribution, $\mathcal{P}(s'|s,a):\mathcal{S}\times\mathcal{A}\rightarrow\Delta(\mathcal{S})$ is the transition kernel, and $\gamma\in[0, 1]$ is the discount factor.
The objective of RL is to find a policy $\pi(a|s)$ that maximizes the expected return: $\mathbb{E}_\pi\left[\sum_k\gamma^k r(s_k,a_k)\right]$.

Off2On RL is a two-stage learning framework that performs offline pre-training and subsequent online learning \citep{off2on,cal-ql,wsrl,fino}.
In the offline pre-training stage, the agent is trained on a pre-collected dataset $D=\{(s,a,r,s')\}$ without any interaction with the environment.
During the subsequent online learning stage, the trained policy interacts directly with the environment, further refining its behavior.

\textbf{Action Chunking in RL} ~
Action Chunking, originally introduced in IL to improve robotic control efficiency \citep{act}, has become a key technique in robotics research \citep{pi_0,diffusionpolicy,groot}.
By grouping a sequence of actions into a single unit, referred to as an action chunk, this approach reduces decision-making frequency while promoting consistent execution.
Recently, the action chunking approach has been extended to RL \citep{top-erl,qc,dqc,ac3}.
Instead of learning the value of a single action $a_t$ as in conventional approaches, 
the value of an action chunk $\bm{a}_{t:t+h}:=(a_t,a_{t+1},\cdots,a_{t+h})$ is learned to enable faster value propagation and unbiased value estimation, thereby facilitating effective and stable temporal-difference (TD) learning \citep{qc,dqc,deas}. 

In the context of Off2On RL, action chunking can be further leveraged to enhance online exploration through consistent behavior.
Q-Chunking (QC) \citep{qc} combines this exploration benefit with the value learning advantages of chunking, employing a flow-based policy as its backbone.
To this end, a flow-based behavior policy $\beta_{\psi}$ is trained by learning a velocity field $v_\psi$ in a chunk-wise manner:
\begin{equation}
    \mathcal{L}_{\beta}(\psi)=\mathbb{E}_{\substack{(s_t,\bm{a}_{t:t+h})\sim D,\\ u\sim\text{Unif}([0,1]),\\\mathbf{z}\sim \mathcal{N}(0,I_{A(h+1)})}}\Big[\big\|v_\psi(s_t,\bm{a}_u,u)-(\bm{a}_{t:t+h}-\mathbf{z})\big\|_2^2\Big]
    \label{eq:flow}
\end{equation}
where $\bm{a}_u=(1-u)\mathbf{z}+u\bm{a}_{t:t+h}$ and $A$ denotes the action space dimension.
Unlike conventional single-action learning, the velocity field here is defined over action chunks. By integrating the trained velocity field $v_\psi$ from initial noise  $\mathbf{z}\sim\mathcal{N}(0,I_{A(h+1)})$, the behavior policy samples an action chunk $\bm{a}_{t:t+h}$ for a given state $s_t$.
Subsequently, behavior-regularized Q-maximization is applied to derive a one-step flow policy \citep{fql}:
\begin{equation}
\mathcal{L}_\pi(\theta)=\mathbb{E}_{\substack{s_t\sim D,\\\mathbf{z}\sim\mathcal{N}(0,I_{A(h+1)})}}
    \Big[-Q_\phi(s_t,\pi_\theta(s_t,\mathbf{z}))+\alpha\,\big\|\pi_\theta(s_t,\mathbf{z})-\beta_\psi(s_t,\mathbf{z})\big\|_2^2\Big]
    \label{eq:onestep}
\end{equation}

where $\alpha$ is a hyperparameter controlling the strength of the regularization.
Beyond policy action generation, the action-value function is also learned in a chunk-wise manner via temporal-difference (TD) learning: {\small
\begin{equation}
\mathcal{L}_Q(\phi)=\mathbb{E}_{\substack{(s_t,\bm{a}_{t:t+h},\bm{r}_{t:t+h},s_{t+h+1})\sim D,\\\bm{a}^\prime_{t+h+1:t+2h+1}\sim\pi_\theta(\cdot|s_{t+h+1})}}\Big[\big(Q_\phi(s_t,\bm{a}_{t:t+h})-R_{t:t+h}-\gamma^{h+1}Q_{\bar{\phi}}(s_{t+h+1},\bm{a}^\prime_{t+h+1:t+2h+1})\big)^2\Big]
    \label{eq:qc}
\end{equation}}where $R_{t:t+h}=\sum_{k=0}^{h}\gamma^kr_{t+k}$ and $\bar{\phi}$ denotes the target network parameters.
In this way, both the policy and the value function operate on action chunks rather than individual actions.

\section{Method}
\label{sec:method}

In this section, we present ACH, a novel framework for Off2On RL.
We first introduce a value-based criterion for selecting the chunk length (Section~\ref{subsec:value_based_optimality}), then describe a Transformer-based value function that estimates the values of all candidate chunk lengths in parallel (Section~\ref{subsec:transformer}), and finally present the adaptive chunk length sampling mechanism (Section~\ref{subsec:sampling}).

\subsection{Value-based Chunk Length Selection}
\label{subsec:value_based_optimality}

To adapt the chunk length at each state $s_t$, we need a criterion that reflects how the chunk length affects performance.
Intuitively, the chunk length involves a trade-off between two opposing factors.
The first is \emph{the uncertainty of predicted future states}: since an action chunk is sampled conditioned solely on $s_t$ and executed in an open-loop manner, its actions become suboptimal when the actual trajectory deviates from the states implicitly predicted at time $t$, which favors shorter chunks.
The second is \emph{policy stochasticity jittering loss}: when future states are highly correlated with $s_t$, resampling at every step interrupts the continuity of behaviors that a longer chunk would provide \citep{qc}, and in stochastic policies such as flow-based models, each independent sampling introduces jitter that hinders the agent's progress toward the goal, which favors longer chunks.
Thus, a chunk length of one is not always optimal even when computational cost is disregarded, and the best chunk length depends on how these two factors balance at each state.

One way to balance this trade-off is to design heuristic criteria, such as bounding the estimated entropy of predicted future states or actions, possibly combined with a penalty on action jittering.
A recent study by \citet{liang2026adaptive} falls into this category, adjusting the chunk length at inference time based on the average entropy of the predicted actions and its maximum gap.
However, such criteria affect the return only indirectly, so the resulting chunk length may be suboptimal, and they require multiple hand-designed functions and thresholds.

Instead of relying on such heuristics, we take a different perspective and directly employ the values of action chunks as our optimality criterion, aligning with the fundamental goal of RL: return maximization.
Concretely, we construct a set $\mathcal{A}_h$ of action chunks with lengths up to $h$, all originating from $s_t$, as follows:
\begin{equation} \label{eq:ChunkSet}
    \mathcal{A}_h =\{ (s_t,a_t), (s_t,a_t,a_{t+1}), \cdots, (s_t, a_t,a_{t+1},\cdots, a_{t+h})  \},
\end{equation}
where each element includes the preceding element as its prefix.
To compare these chunks, we evaluate each of them by its value under a policy $\pi$, i.e., the expected return of executing the chunk $(a_t,\cdots,a_{t+n})$ from $s_t$ and following $\pi$ afterward:
\begin{equation} \label{eq:chunkQ}
    Q^\pi(s_t, a_t,\cdots,a_{t+n}) = \mathbb{E}_{a_{t+n+1}, a_{t+n+2},\cdots \sim \pi}\Big[\textstyle\sum_{k} \gamma^{k} r_{t+k} \,\Big|\, s_t, a_t, \cdots, a_{t+n} \Big]
\end{equation}
Evaluating every chunk in $\mathcal{A}_h$ in this way yields the value sequence
\begin{equation}  \label{eq:QvalueSeq}
Q^\pi(s_t,a_t), Q^\pi(s_t,a_t,a_{t+1}), \cdots, Q^\pi(s_t,a_t,a_{t+1},\cdots,a_{t+h}).
\end{equation}
This sequence progressively evaluates the value of action chunks $(a_t,\cdots,a_{t+n})$ for given $s_t$, with  $n$ ranging up to $h$.
In other words, the sequence describes how the expected return changes as the agent commits to a longer chunk at $s_t$ before making its next decision, as illustrated in Appendix~\ref{appendix:motivation}.

The key of this approach lies in the prefix structure of the chunk sequence.
Specifically, if we were to compare $Q^\pi(s_t,a_t,a_{t+1})$ and $Q^\pi(s_t, a_t^\prime, a_{t+1}^\prime, a_{t+2})$  where  $(a_t,a_{t+1}) \ne (a_t^\prime, a_{t+1}^\prime)$, the difference in Q-values could stem from the initial action mismatch rather than the chunk length itself.
By maintaining a prefix-consistent sequence, we can isolate the effect of length and determine the optimal chunk length within the range $[1,h+1]$ by selecting the index that maximizes the value in the sequence in (\ref{eq:QvalueSeq}).

\subsection{Transformer-based Multi-Chunk Value Estimation}
\label{subsec:transformer}

\begin{figure}[t]
    \centering
    \includegraphics[width=0.85\linewidth]{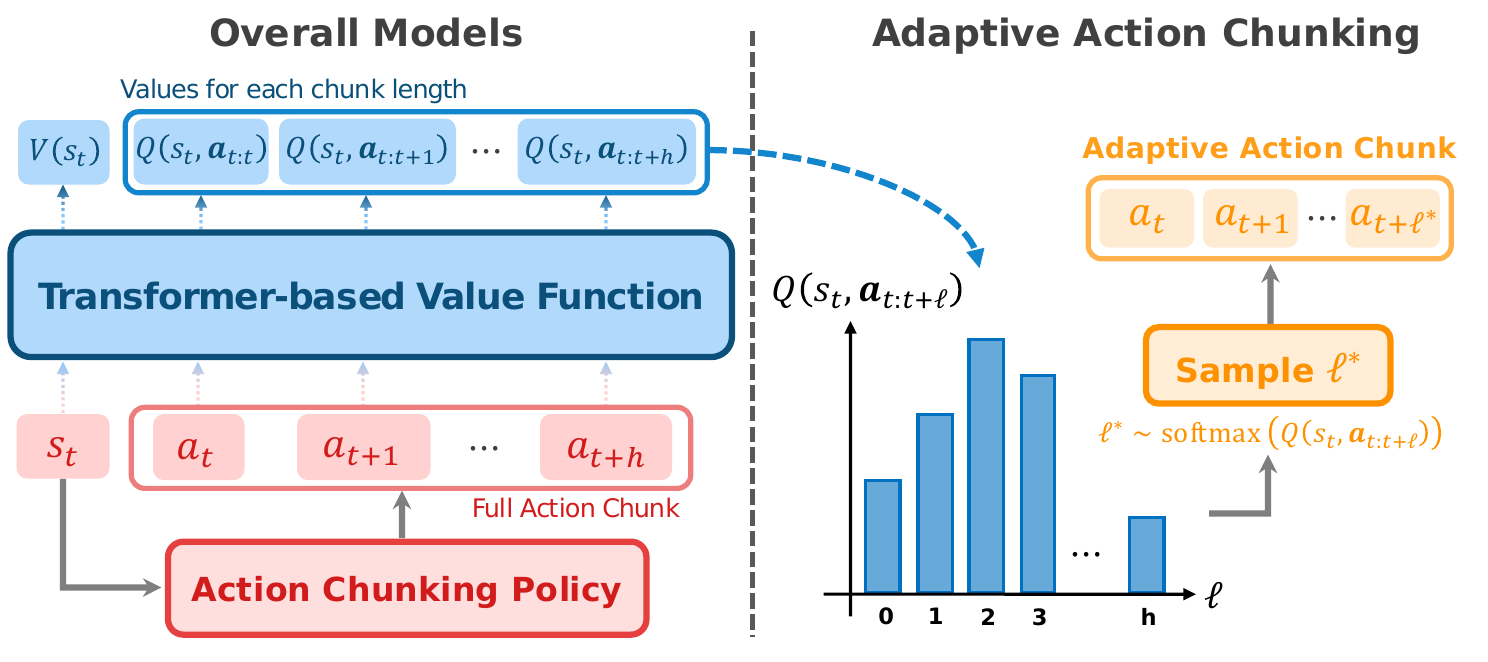}
    \vspace{-2ex}
    \caption{The overall architecture of ACH. (left) Causal Transformer-based joint value learning: from $(s_t,\bm{a}_{t:t+h})$, the value function outputs $V(s_t)$ and the action values of all prefix chunks in a single forward pass. (right) Chunk length sampling: the chunk length is drawn from the softmax distribution over these action values.}
    \label{fig:method}
    \vspace{-3ex}
\end{figure}

The main hurdle to realizing our adaptive chunk length selection described above is how to learn the Q-values of multiple increasing chunks in (\ref{eq:QvalueSeq})  simultaneously.
To solve this problem,  we adopt a {\em causal Transformer} as our value function architecture.
Building on previous work \citep{top-erl}, which utilized a Transformer-based value function for episodic RL, we design a value function that takes $(s_t, \bm{a}_{t:t+h})$ as input and outputs $(V(s_t), Q(s_t, \bm{a}_{t:t}), Q(s_t,\bm{a}_{t:t+1}),\cdots,Q(s_t,\bm{a}_{t:t+h}))$,  as illustrated in Figure \ref{fig:method}, where the state value function $V(s)$ is additionally included for efficient bootstrapping.  With a causal Transformer as the value function, we can estimate the values of all possible action chunks in a single forward pass, while ensuring that action chunks do not attend to future steps.

To train the value function under multiple chunk lengths, we follow the training scheme of the previous work \citep{top-erl} and modify it for adaptive chunk lengths, which yields the following objective that trains the entire value sequence in (\ref{eq:QvalueSeq}) at once under the current policy:
\begin{equation} 
    \begin{aligned}
        \mathcal{L}_Q(\phi) = \mathbb{E}_{\substack{(\bm{s}_{t:t+h}, \bm{a}_{t:t+h}, \\ \bm{r}_{t:t+h}, s_{t+h+1}) \sim D, \\ \bm{a}^\prime_{t:t+h} \sim \pi_\theta(\cdot|s_t)}} 
         \Biggl[ \frac{1}{h+1} \sum_{n=0}^{h} &\biggl( Q_\phi(s_t, \bm{a}_{t:t+n}) - \big(R_t^n + \gamma^{n+1} V_{\bar{\phi}}(s_{t+n+1})\big) \biggr)^2 \\
        & + \biggl( V_\phi(s_t) - \mathbb{E}_{\ell\sim \pi_L(\cdot|s_t,\bm{a}^\prime_{t:t+h})}\big[Q_{\bar{\phi}}(s_t, \bm{a}^\prime_{t:t+\ell})\big] \biggr)^2 \Biggr]
    \end{aligned}
    \label{eq:transformer}
\end{equation}
where $R_t^n=\sum_{k=0}^n\gamma^k r_{t+k}$, ~~$\bm{a}^\prime_{t:t+\ell}$ is a truncation (up to $\ell$) of full chunk $\bm{a}^\prime_{t:t+h}$ sampled from $\pi_\theta(\cdot|s_t)$, and $\pi_L(\ell|s_t,\bm{a}^\prime_{t:t+h})$ denotes the  chunk length selection policy over $\ell\in\{0,1,\dots,h\}$, where $\ell$ corresponds to executing the prefix $\bm{a}^\prime_{t:t+\ell}$, i.e., a chunk length of $\ell+1$. This policy will be explained shortly in Section \ref{subsec:sampling}.

The loss above consists of two terms.
The first term learns the action value for each chunk length $n$, which corresponds to the $n$-th element of (\ref{eq:QvalueSeq}), using the $n$-step target.
By estimating the value of the action chunk $\bm{a}_{t:t+n}$ covering the $n$-step reward $\bm{r}_{t:t+n}$, the proposed method yields less biased value estimation.
Moreover, unlike the previous approach \citep{qc} that requires the state-action value of the next $n$-step chunk $\bm{a}_{t+n+1:t+2n+1}$ when applying bootstrapping, our method instead builds the target from the value of the state $s_{t+n+1}$, thereby avoiding the variance from action-based bootstrapping.
The second term learns the state value from the current policy's behavior.
In doing so, it performs value estimation according to the chunk length selection distribution.
We further discuss the connections of Eq.~(\ref{eq:transformer}) to multi-step TD learning and the options framework in Appendix~\ref{appendix:related}.

Based on the learned value function, we train the policy $\pi_\theta$ by jointly optimizing distillation from the behavior policy $\beta_\psi$ learned via flow matching and action-value maximization:
\begin{equation}    \mathcal{L}_\pi(\theta)=\mathbb{E}_{\substack{s_t\sim D,\\\mathbf{z}\sim\mathcal{N}(0,I_{A(h+1)}),\\\bm{a}^\prime_{t:t+h}\sim\pi_\theta(s_t,\mathbf{z})}}
    \Big[-\mathbb{E}_{\ell\sim \pi_L(\cdot|s_t,\bm{a}^\prime_{t:t+h})}\big[Q_\phi(s_t,\bm{a}^\prime_{t:t+\ell})\big]+\alpha\,\big\|\bm{a}^\prime_{t:t+h}-\beta_\psi(s_t,\mathbf{z})\big\|_2^2\Big],
    \label{eq:policy}
\end{equation}
where $\alpha$ is a behavior cloning coefficient.
Note that the value-maximization term is taken in expectation over $\ell\sim\pi_L$, so the policy is optimized for the chunk lengths that it will actually execute rather than for a single fixed length.
This keeps the training objective aligned with the adaptive execution, guiding the policy to learn action chunks with high action values at the lengths that $\pi_L$ selects during environment interaction.

\subsection{Adaptive Chunk Length Sampling}
\label{subsec:sampling}

We generate a full action chunk $\bm{a}^\prime_{t:t+h}$ for current state $s_t$ using the policy $\pi_\theta(\cdot|s_t)$, and feed the full sequence $(s_t,\bm{a}^\prime_{t:t+h})$ to the transformer-based value function. This yields the Q-values of all $h+1$ prefix chunks: $(s_t,\bm{a}^\prime_{t:t}),(s_t,\bm{a}^\prime_{t:t+1}),\cdots,(s_t, \bm{a}^\prime_{t:t+h})$.  One could then select the optimal chunk length for  $s_t$ as $h^* = 1+\arg\max_{n\in\{0,1,\cdots,h\}} Q_\phi(s_t,\bm{a}^\prime_{t:t+n})$, 
resulting in a greedy selection policy. However, inspired by 
$\epsilon$-greedy strategies in Q-learning, we further refine this selection process to incorporate exploration capabilities during online learning. Consequently, we define the chunk length selection policy, $\pi_L$
, as a soft policy following a sampling distribution:
\begin{equation}
\pi_L(\ell)=\frac{\exp(\beta \cdot Q_\phi(s_t,\bm{a}^\prime_{t:t+\ell}))}{\textstyle\sum_{n=0}^h\exp(\beta \cdot Q_\phi(s_t,\bm{a}^\prime_{t:t+n}))}, \quad \forall\ell\in\{0, 1, \cdots, h\}
    \label{eq:sampling}
\end{equation}
where $\beta$ denotes the inverse temperature parameter.
Since both the policy and the value function are trained to prioritize chunks with high action values in the offline dataset via Eqs. (\ref{eq:transformer}) and (\ref{eq:policy}), this method naturally selects action chunks that align with the training dynamics.
We sample lengths during training to encourage exploration and select the highest-valued prefix at evaluation.

To provide a theoretical basis for adaptive chunk length selection, we analyze the prefix values learned by ACH in Appendix~\ref{appendix:theory}. Under deterministic transitions and exact policy evaluation, each prefix value is the expected return from executing that prefix and then resuming the current policy. Extending a prefix by one step increases its value precisely when the next planned action yields a higher expected return than replanning at that state. We also characterize the optimal values attained by exact policy iteration with and without a fixed behavior cloning constraint.

\section{Experiments}
\label{sec:experiments}

In this section, we empirically evaluate the proposed method, ACH, across challenging environments involving long-horizon and diverse tasks in comparison with several key baselines.

\newcommand{\res}[2]{#1 & #2}
\begin{table}[t!]
    \centering
    \footnotesize
    \vskip 0.1in
    \renewcommand{\arraystretch}{1.1}
    \caption{Aggregated results of ACH and baselines on OGBench. All methods are evaluated after 1M steps of offline learning followed by 1M steps of online learning. We report the mean and 95\% confidence intervals across 10 seeds. Results for each environment are averaged over five tasks.}
    \resizebox{\textwidth}{!}{%
    \begin{tabular}{l|*{7}{r@{\,$\rightarrow$\,}l}}
    \toprule
    Task
      & \multicolumn{2}{c}{RLPD}
      & \multicolumn{2}{c}{ReBRAC}
      & \multicolumn{2}{c}{BFN}
      & \multicolumn{2}{c}{FQL}
      & \multicolumn{2}{c}{FINO}
      & \multicolumn{2}{c}{QC}
      & \multicolumn{2}{c}{ACH} \\
    \midrule
    antmaze-giant-navigate & \res{0{\tiny{$\pm$0}}}{69{\tiny{$\pm$9}}} & \res{30{\tiny{$\pm$6}}}{\textbf{99}{\tiny{$\pm$0}}} & \res{0{\tiny{$\pm$0}}}{22{\tiny{$\pm$6}}} & \res{10{\tiny{$\pm$5}}}{74{\tiny{$\pm$7}}} & \res{2{\tiny{$\pm$2}}}{76{\tiny{$\pm$4}}} & \res{0{\tiny{$\pm$0}}}{47{\tiny{$\pm$8}}} & \res{1{\tiny{$\pm$1}}}{72{\tiny{$\pm$2}}} \\
    antsoccer-arena-navigate & \res{0{\tiny{$\pm$0}}}{64{\tiny{$\pm$4}}} & \res{1{\tiny{$\pm$1}}}{0{\tiny{$\pm$0}}} & \res{51{\tiny{$\pm$2}}}{74{\tiny{$\pm$7}}} & \res{35{\tiny{$\pm$2}}}{70{\tiny{$\pm$5}}} & \res{40{\tiny{$\pm$3}}}{71{\tiny{$\pm$5}}} & \res{10{\tiny{$\pm$2}}}{67{\tiny{$\pm$2}}} & \res{0{\tiny{$\pm$0}}}{\textbf{84}{\tiny{$\pm$1}}} \\
    cube-quadruple-play-10M & \res{0{\tiny{$\pm$0}}}{0{\tiny{$\pm$0}}} & \res{0{\tiny{$\pm$0}}}{0{\tiny{$\pm$0}}} & \res{2{\tiny{$\pm$1}}}{10{\tiny{$\pm$2}}} & \res{0{\tiny{$\pm$0}}}{2{\tiny{$\pm$1}}} & \res{1{\tiny{$\pm$0}}}{7{\tiny{$\pm$3}}} & \res{1{\tiny{$\pm$1}}}{33{\tiny{$\pm$3}}} & \res{0{\tiny{$\pm$0}}}{\textbf{52}{\tiny{$\pm$6}}} \\
    cube-triple-play & \res{0{\tiny{$\pm$0}}}{0{\tiny{$\pm$0}}} & \res{0{\tiny{$\pm$0}}}{0{\tiny{$\pm$0}}} & \res{4{\tiny{$\pm$1}}}{22{\tiny{$\pm$3}}} & \res{2{\tiny{$\pm$1}}}{19{\tiny{$\pm$2}}} & \res{2{\tiny{$\pm$1}}}{24{\tiny{$\pm$3}}} & \res{4{\tiny{$\pm$2}}}{57{\tiny{$\pm$4}}} & \res{1{\tiny{$\pm$1}}}{\textbf{67}{\tiny{$\pm$3}}} \\
    humanoidmaze-medium-navigate & \res{0{\tiny{$\pm$0}}}{29{\tiny{$\pm$9}}} & \res{34{\tiny{$\pm$4}}}{4{\tiny{$\pm$2}}} & \res{59{\tiny{$\pm$3}}}{87{\tiny{$\pm$6}}} & \res{36{\tiny{$\pm$3}}}{56{\tiny{$\pm$3}}} & \res{38{\tiny{$\pm$5}}}{\textbf{92}{\tiny{$\pm$4}}} & \res{15{\tiny{$\pm$2}}}{32{\tiny{$\pm$4}}} & \res{45{\tiny{$\pm$2}}}{82{\tiny{$\pm$1}}} \\
    puzzle-4x4-play & \res{0{\tiny{$\pm$0}}}{64{\tiny{$\pm$12}}} & \res{14{\tiny{$\pm$2}}}{32{\tiny{$\pm$6}}} & \res{32{\tiny{$\pm$2}}}{60{\tiny{$\pm$7}}} & \res{14{\tiny{$\pm$2}}}{54{\tiny{$\pm$6}}} & \res{15{\tiny{$\pm$3}}}{53{\tiny{$\pm$7}}} & \res{11{\tiny{$\pm$2}}}{38{\tiny{$\pm$4}}} & \res{13{\tiny{$\pm$3}}}{\textbf{90}{\tiny{$\pm$8}}} \\
    \bottomrule
    \end{tabular}%
    }
    \label{tab:main}
\par\medskip
\begin{minipage}{\linewidth}
  \centering
  \vspace{-2ex}
  \includegraphics[width=\linewidth]{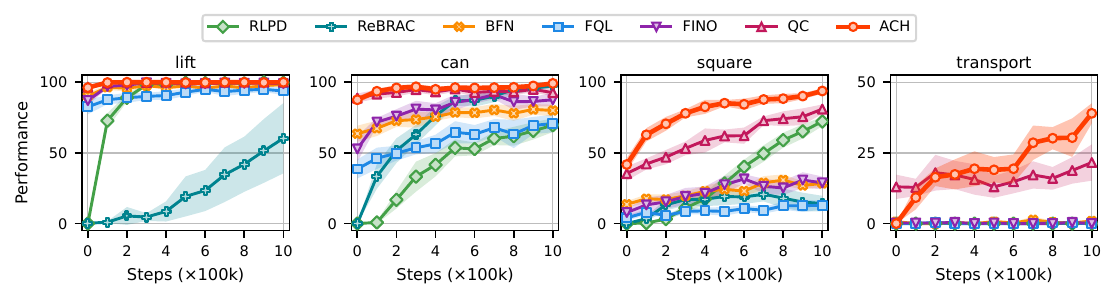}
  \vspace{-3ex}
  \captionof{figure}{Learning curves over 1M online steps across robomimic tasks. The performance is  averaged over 10 random seeds, with the 95\% confidence interval as  the shaded region.}
  \vspace{-6.5ex}
  \label{fig:robomimics}
\end{minipage}
\end{table}

\textbf{Benchmarks and Tasks} ~
We evaluate ACH on 30 tasks from six environments of OGBench \citep{ogbench}, spanning locomotion and manipulation, and on four robotic manipulation tasks from robomimic \citep{robomimic}.
Details of the tasks and datasets are provided in Appendix~\ref{appendix:benchmarks}.

\textbf{Baselines} ~
We compare ACH against the following Off2On RL methods:
(1) RLPD \citep{rlpd}, an online RL algorithm that starts with an offline dataset and achieves high performance without an offline learning phase.
(2) ReBRAC \citep{rebrac}, an offline RL method built on a Gaussian policy, which has shown strong performance in both offline and Off2On RL.
(3) BFN is a method introduced in the QC paper \citep{qc}, which draws multiple action candidates from a behavior policy and selects the one with the highest estimated action value.
(4) FQL \citep{fql}, an offline RL method that learns a behavior policy via flow matching and distills it into a one-step policy, demonstrating strong performance across offline and Off2On RL.
(5) FINO \citep{fino}, an algorithm targeting Off2On RL with a flow-based policy, which builds upon FQL and injects noise into the flow matching for strong Off2On performance.
(6) QC \citep{qc}, as discussed in Section \ref{sec:prelim}, extends the conventional single-action approach to action chunks, and serves as our backbone algorithm.

\textbf{Implementation Details} ~
We build ACH upon QC \citep{qc}, using its variant that trains a one-step policy for efficient inference, and set the maximum chunk length to 10 ($h=9$) in most environments.
Further implementation details, hyperparameters, and the pseudo-code are provided in Appendix~\ref{appendix:ach}.

\textbf{Evaluations} ~
For all baselines, we apply the same protocol to obtain experimental results.
Training consists of 1M steps for offline learning followed by 1M steps for online learning, and the final results are based on the performance evaluated at the last step.
All experiments are conducted with 10 random seeds per task, and we report the mean and 95\% confidence intervals.
To fairly compare the performance improvement during online learning, we also report the offline performance.
We bold all results within 95\% of the top-performing algorithm after online learning.

\textbf{Results} ~Table \ref{tab:main} summarizes the performance across 30 OGBench tasks, aggregated by environment.
The results demonstrate that ACH achieves superior performance across diverse domains.
Specifically, RLPD, which conducts online RL without a prior offline phase, exhibits clear limitations inherent in its learning paradigm.
Although ReBRAC achieves strong performance in \texttt{antmaze}, it struggles in more complex environments such as \texttt{antsoccer} and \texttt{humanoidmaze}, where it fails to improve or even collapses during online learning, which we attribute to the limited expressiveness of its unimodal policy.
BFN employs a sampling strategy similar to the proposed method, yet fails to achieve strong performance due to the inherent limitation of single-action prediction.
Moreover, the performance of FQL suggests that simply extending offline RL methods to online fine-tuning is insufficient for addressing the challenges of the Off2On transition.
While FINO performs competitively in navigation-heavy tasks such as \texttt{humanoidmaze}, its performance in manipulation tasks (e.g., \texttt{cube}) reveals the limitations of single-action prediction in scenarios requiring precise manipulation.
Lastly, while QC demonstrates that action chunking yields competitive results in most tasks, its comparison with ACH underscores that our proposed method facilitates more efficient learning.
This gain is primarily attributed to ACH’s ability to employ adaptive chunk lengths, as opposed to the rigid, fixed-length approach of the baseline.

Figure \ref{fig:robomimics} illustrates the learning curves for four environments within robomimic during the online fine-tuning.
The results demonstrate that ACH consistently achieves the highest performance across all tasks.
This performance gain is particularly pronounced in the \texttt{square} and \texttt{transport} tasks, where the backbone algorithm with a fixed chunk length exhibits only marginal improvements.
These findings, together with the results in Table \ref{tab:main}, validate ACH as a robust framework that outperforms conventional methods by adaptively modulating chunk lengths across challenging domains.

\subsection{Scaling to Vision-Language-Action Models}
\label{subsec:vla}
Vision-language-action (VLA) models have been actively studied in robotics, and many of them adopt action chunking \citep{pi_0,groot}.
To examine whether ACH also improves such large-scale models, we apply it to a VLA model.
We evaluate on SimplerEnv \citep{simplerenv}, covering seven tasks with the WidowX robot and six tasks with the Google Robot, and all methods build upon the same GR00T N1.7 model \citep{groot}, which serves as the base policy.
We compare ACH against the following methods:
(1) Base, the base policy without any additional training;
(2) QC \citep{qc}, the backbone algorithm of ACH, which is trained and executed with a fixed chunk length;
(3) AAC \citep{liang2026adaptive}, which adapts the chunk length of the base policy at inference time based on the entropy of the predicted actions;
(4) QC+AAC, which trains the policy with QC and additionally applies AAC at inference time.

\begin{figure}[t]
    \centering
    \begin{minipage}[t]{0.485\textwidth}
        \centering
        \includegraphics[width=\linewidth]{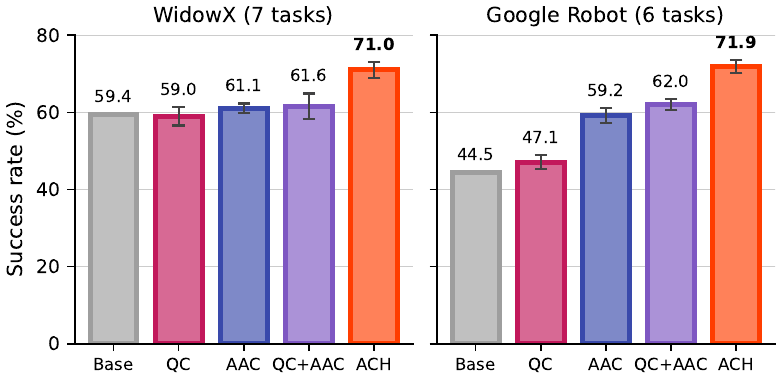}
        \vspace{-4ex}
        \caption{Success rates on SimplerEnv, averaged over 7 WidowX tasks and 6 Google Robot tasks. Error bars denote 95\% confidence intervals. Per-task results are reported in Appendix \ref{appendix:vla_results}.}
        \label{fig:vla}
    \end{minipage}\hfill
    \begin{minipage}[t]{0.485\textwidth}
        \centering
        \includegraphics[width=\linewidth]{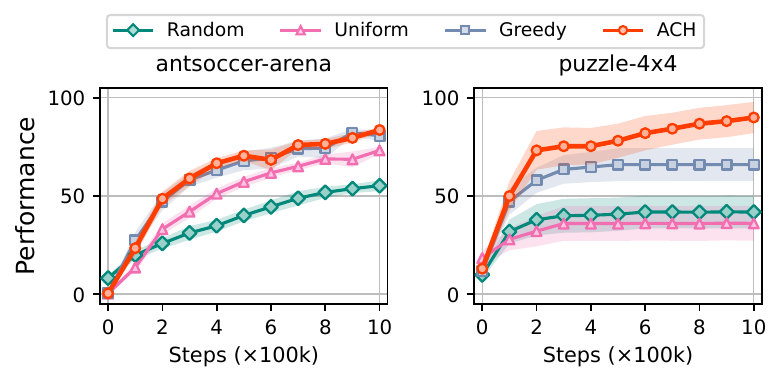}
        \vspace{-4ex}
        \caption{Experimental comparison of ACH against alternative training and sampling strategies. Each plot presents the aggregated performance across five distinct tasks.}
        \label{fig:ablation}
    \end{minipage}
    \vspace{-3ex}
\end{figure}

Following the base policy, which predicts action chunks of up to 8 steps, we set the chunk length to 8 for all methods, which serves as the maximum chunk length for the adaptive methods.
For QC and ACH, we keep the VLA model frozen and train only lightweight one-step policy and value networks through offline-to-online RL.
For offline training, we collect 1,000 episodes per task with the base policy and use them to train the one-step policy and the value function of each method, followed by online training for 100K environment steps.
Each task is evaluated over 200 episodes, and we report the mean and 95\% confidence intervals across 4 seeds.
Further details are provided in Appendix \ref{appendix:vla}.

As shown in Figure \ref{fig:vla}, ACH achieves the highest success rates on both robots, improving the average success rate of the base policy from 59.4\% to 71.0\% on WidowX and from 44.5\% to 71.9\% on the Google Robot.
Although QC is trained through offline-to-online RL, it performs similarly to the base policy, indicating that training alone cannot compensate for a rigid execution horizon.
AAC improves the base policy on the Google Robot without additional training, but its gain on WidowX is marginal, as its entropy-based heuristic reflects the resulting return only indirectly.
QC+AAC yields only modest gains over AAC, indicating that merely coupling a policy trained under a fixed chunk length with an inference-time heuristic cannot fully exploit adaptive chunking, since the chunk length is never taken into account during training.
In contrast, ACH adapts the chunk length during both training and inference, aligning the training objective with the chunk lengths actually executed, and thus outperforms all baselines.
These results demonstrate that ACH is also applicable to large-scale models that take visual observations as input.

\subsection{Further Analysis}
\label{subsec:analysis}
In this section, we present an ablation study on the core components of ACH and a behavior analysis that shows how adaptive action chunking shapes the behavior of the agent.
Additional analyses are provided in Appendix~\ref{appendix:additional}.

\par\noindent\textbf{Ablation Study}\quad\ignorespaces
Section \ref{sec:method} presents the rationale for the value-based chunk length selection criterion (Section \ref{subsec:value_based_optimality}), describes the training of the value function and policy (Section \ref{subsec:transformer}), and details the adaptive chunk length sampling procedure (Section \ref{subsec:sampling}).
To assess the impact of these core components, we conduct an ablation study by replacing each with an alternative approach.
For the chunk length selection criterion, we examine a variant that randomly selects a length among all candidates instead of value-based sampling (denoted as \texttt{Random}).
Regarding the training procedure, we evaluate a variant that assigns a uniform weight to all chunk length candidates, i.e., $\ell \sim \mbox{Unif}\{0,1,\dots,h\}$ in  Eqs. (\ref{eq:transformer}) and (\ref{eq:policy}) rather than prioritizing the distribution $\pi_L$ within the training losses (denoted as \texttt{Uniform}).
To evaluate the sampling mechanism, we replace our soft sampling strategy Eq. (\ref{eq:sampling}) with a greedy approach that selects the chunk length with the highest estimated value during online interaction (denoted as \texttt{Greedy}).

As shown in Figure \ref{fig:ablation}, replacing any of these components leads to a performance decline, confirming that no substitution matches the effectiveness of our proposed method.
Simply selecting the chunk length at random does not consider the expected return of executing the corresponding action chunk, resulting in inferior performance across all environments.
Training uniformly across all chunk lengths results in significant degradation, which we attribute to the misalignment between the training objective and the action selection policy used during online learning.
The greedy alternative achieves competitive performance in some environments due to its behavior similarity to our method; however, since it solely exploits the Q-values, it fails to explore sufficiently.
Consequently, it exhibits suboptimal sample efficiency compared to ACH.
These findings confirm that the synergy between our training and sampling mechanisms is essential to enabling sample-efficient learning, which is consistently observed on robomimic (Appendix~\ref{appendix:ablation_robomimic}).

\par\noindent\textbf{Behavior Analysis}\quad\ignorespaces
\begin{figure}[t]
  \centering
  \includegraphics[width=\linewidth]{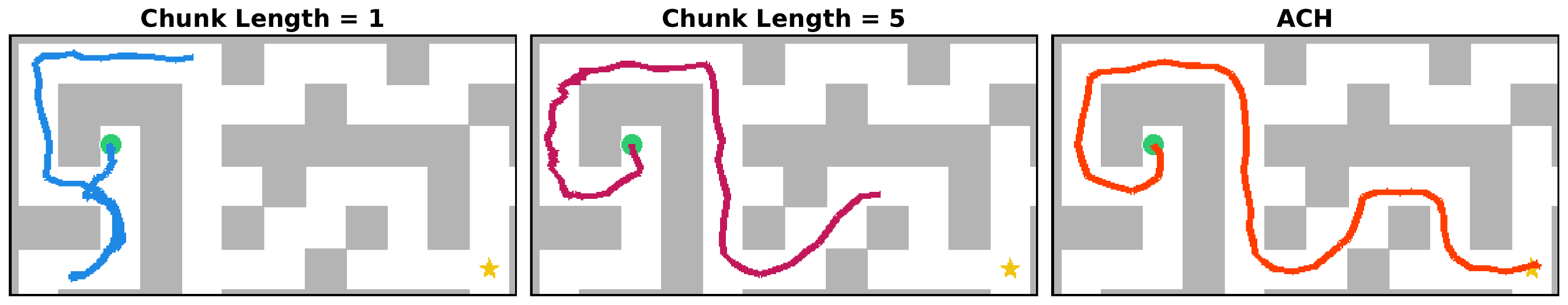}
  \vspace{-4ex}
  \caption{Behavioral comparison of ACH and fixed-length baselines in the \texttt{antmaze-giant} environment. Starting from a shared initial state, all algorithms execute the same number of total action steps. Agent trajectories are constructed by connecting sequential positions at each timestep. The initial state and goal state are marked by a green circle and a yellow star, respectively.}
  \vspace{-4ex}
  \label{fig:analysis}
\end{figure}
To verify the efficacy of the adaptive chunk length selection, we conducted a comparative analysis in the challenging \texttt{antmaze-giant} environment.
In Figure \ref{fig:analysis}, we compare the trajectories of our proposed method against two baselines: a model that does not utilize action chunking (employing a fixed chunk length of 1) and a backbone model using a static, fixed chunk length of 5.
All baselines are trained within the same learning timestep, and the trajectories are recorded by sampling from the final trained policies.

The results show that the model without action chunking exhibits redundant, oscillatory (i.e., jittering) behaviors near the initial state.
This is because the policy predicts only a single action at each step, leading to frequent inconsistencies in decision making and highlighting the necessity of action chunking.
While the fixed-length action chunking demonstrates improved behavior compared to single-action prediction, it still suffers from suboptimal efficiency.
For instance, it fails to execute efficient maneuvers in areas requiring precise control, such as navigating corners, and ultimately fails to reach the goal within the timestep limit.
In contrast, our method selects the most appropriate chunk length for every state, resulting in efficient and smooth trajectories that successfully reach the goal.
These behavioral patterns are accumulated as high-quality transitions during online learning, which facilitates sample-efficient training and enhanced performance.
We present the results on the full map of Figure \ref{fig:analysis} and additional behavioral patterns discovered for chunk length 5 in Appendix \ref{appendix:analysis}.

\section{Related Work}

\textbf{Offline-to-online RL} ~
Off2On RL is a framework that addresses the inability of offline RL to learn beyond the given dataset and the safety and cost issues of applying online RL from scratch \citep{off2on,pex,cal-ql,aca,famo2o,cft,so2,opt,wsrl,fino}.
The most straightforward approach is to extend existing offline RL methods \citep{cql,iql,spot,rebrac} to online learning, but such approaches fail to fully exploit the interaction due to the inherited conservatism of offline RL.
Prior works \citep{off2on,cal-ql,aca,famo2o,cft,so2,opt} proposed easing conservatism, yet these methods still retain it, limiting sample efficiency.
Recently, several works \citep{idql,fql,expo,fino,fac} have sought to leverage the multimodality of expressive generative models \citep{diffusion2,diffusion1,flow2,flow1} as policies, but since they rely on single-action prediction, they tend to produce redundant or inconsistent behaviors, resulting in limited sample efficiency.
A recently proposed action chunking-based method \citep{qc} achieves strong performance through consistent behavior and value learning benefits, yet remains suboptimal due to its fixed chunk length.
Our method employs adaptive action chunking to select the optimal chunk length for each state, enabling sample-efficient learning within a limited online budget.

\textbf{Action Chunking for RL} ~~~
Action chunking has become a key technique in imitation learning by addressing compounding errors and dataset non-stationarity arising from single-action prediction \citep{act,pi_0,diffusionpolicy,groot,pi0.5,openvla-oft}.
This technique has been extended to reinforcement learning: in offline RL, various studies \citep{qc,cqn,dqc,cgq,deas,mac} have leveraged action chunking, as the resulting horizon reduction facilitates value learning, particularly for long-horizon tasks, while in online RL, prior works \citep{top-erl,chunkingthecritic,sear} have proposed Transformer-based value functions using action chunks to mitigate bias from $n$-step returns.
However, these approaches suffer from limited sample efficiency due to fixed chunk length.
Although prior work \citep{liang2026adaptive} in the imitation domain introduces adaptive chunk length selection, it targets only inference-time adjustment without incorporating adaptation into training.
In contrast, our method leverages action chunking while adaptively adjusting its length during both training and inference, leading to strong sample efficiency in Off2On RL.
We further discuss the connections to multi-step TD learning, the options framework, and hierarchical RL in Appendix \ref{appendix:related}.

\section{Conclusion}

In this paper, we introduced ACH, a novel offline-to-online RL algorithm that dynamically modulates chunk length during both training and inference.
To determine optimal chunk length for a current state, we adopt chunk value as an optimality criterion and employ a causal Transformer-based critic to simultaneously evaluate all candidate chunk lengths in a single forward pass.
To enhance exploration during online fine-tuning, we developed a soft policy that adaptively samples chunk lengths based on these values.
Our evaluation across 34 challenging tasks demonstrates that ACH outperforms fixed-length baselines in most environments with enhanced learning efficiency.
Furthermore, experiments with a vision-language-action model show that ACH is applicable to large-scale models.
We believe that our work opens a new avenue for leveraging adaptive action chunking in reinforcement learning, and that this direction will play an important role in advancing efficient learning in sequential decision making.

\newpage

\clearpage
\section{Acknowledgments}
This work was supported by the Institute of Information \& Communications Technology Planning \& Evaluation (IITP) grant funded by the Korea government (MSIT) (No.~RS-2024-00457882, AI Research Hub Project), the National Research Foundation of Korea (NRF) grant funded by the Korea government (MSIT) (No.~RS-2026-25510069), and the ``Advanced GPU Utilization Support Program'' funded by the Government of the Republic of Korea (Ministry of Science and ICT) (No.~02-26-01-0110). The authors declare no competing interests.

\bibliography{references}
\bibliographystyle{plainnat}

\appendix

\newpage

\section{Limitations}
\label{sec:limitation}
Since the proposed method employs a Transformer as the value function, which is relatively heavier than the MLP used in prior methods, training requires comparatively more time.
We report this in Appendix \ref{appendix:compute}.
A potential direction for future work is to identify a lighter alternative to the Transformer that still enables multiple chunk value estimation.
Additionally, as with existing methods, the proposed method executes the sampled action chunk in an open-loop manner, which may limit its applicability in environments with rapidly changing dynamics.
In such cases, replanning during action chunk execution \citep{a2c2,rtc} would be necessary, and this constitutes another promising direction for future research.
{Finally, our VLA experiments are conducted in simulation, and validating ACH on physical robots remains an important direction for future work.}

\section{Additional Related Work\texorpdfstring{ {and Theoretical Analysis}}{ and Theoretical Analysis}}
\label{appendix:related}
In this section, we discuss the connections between ACH and prior research on temporally extended decision making, including multi-step temporal-difference (TD) learning, the options framework, action repetition, and hierarchical RL.
{We then formalize what the objectives of ACH learn in Appendix~\ref{appendix:theory}.}

\par\noindent\textbf{Multi-step TD learning}\quad\ignorespaces
Multi-step TD methods construct the learning target from the rewards over the next $n$ steps and the value estimate of the state reached after $n$ steps.
The choice of $n$ controls a trade-off between the bias of one-step bootstrapping and the variance of Monte Carlo returns \citep{sutton1988learning,rl,kearns2000bias}.
However, the $n$-step return becomes biased when computed from off-policy data, since the actions taken within the $n$ steps are not drawn from the policy being evaluated.
Prior works correct this bias through importance sampling or truncated importance weights \citep{precup2000eligibility,munos2016safe}, while many practical algorithms simply use uncorrected $n$-step returns for faster learning \citep{hessel2018rainbow,barth2018distributed}.
More recently, several works avoid this bias by conditioning the critic on the executed action sequence \citep{top-erl,qc,chunkingthecritic,sear}.
ACH follows this line of work.

From the perspective of multi-step TD learning, the first term of Eq.~(\ref{eq:transformer}) applies $n$-step TD policy evaluation to all $h+1$ prefixes of an action chunk simultaneously, where each $Q(s_t,\bm{a}_{t:t+n})$ is regressed onto the realized $n$-step return plus the bootstrapped value $\gamma^{n+1}V(s_{t+n+1})$.
Since each $Q$ is conditioned on the executed actions, this target avoids the off-policy bias of standard $n$-step returns {when the data are strongly open-loop consistent \citep{dqc}, e.g., under deterministic transitions (Proposition~\ref{prop:evaluation})}.
The bootstrapped value $V$ in this target is learned by the second term as the expected $Q$ under the current policy and $\pi_L$, analogous to Expected Sarsa \citep{van2009theoretical}.
Combined with the policy improvement step in Eq.~(\ref{eq:policy}), Eq.~(\ref{eq:transformer}) forms an actor-critic scheme that generalizes the chunk-wise TD objective of QC \citep{qc} to all chunk lengths.

\par\noindent\textbf{Options and semi-Markov decision processes}\quad\ignorespaces
The options framework formalizes temporally extended actions as options, each consisting of an initiation set, an intra-option policy, and a termination condition \citep{sutton1999between,precup2000temporal}.
Decision making over options is then modeled as a semi-Markov decision process (SMDP), in which each decision can span multiple time steps \citep{bradtke1994reinforcement,sutton1999between}.
Many works learn options and their termination conditions end-to-end \citep{bacon2017option,harb2018waiting,harutyunyan2019termination,klissarov2021flexible}, typically over a small set of discrete options.

From the perspective of the options framework, an action chunk can be viewed as an option that executes a sequence of actions determined at its start for a fixed number of steps, without responding to the states encountered along the way.
Options of this kind are known as macro-actions \citep{vezhnevets2016strategic}.
Such an option is not a Markov option because the action it selects and whether it terminates depend on the number of steps elapsed since it was initiated rather than on the current state alone. It is instead a semi-Markov option, whose policy and termination condition may depend on the entire history since its initiation \citep{sutton1999between}.
The composition of $\pi_\theta$ and $\pi_L$ then plays the role of the policy over options.
For an option given by the chunk $\bm{a}_{t:t+n}$ of duration $n+1$, the only remaining randomness is the environment transition, and the SMDP Bellman expectation equation is
\begin{equation}
    Q(s_t,\bm{a}_{t:t+n}) = \mathbb{E}\Big[\textstyle\sum_{k=0}^{n}\gamma^k r_{t+k} + \gamma^{n+1} V(s_{t+n+1})\Big],
\end{equation}
where $V$ denotes the state value under the policy over options.
The first term of Eq.~(\ref{eq:transformer}) regresses $Q(s_t,\bm{a}_{t:t+n})$ onto exactly this target for every duration, and the second term corresponds to the SMDP relation between $V$ and $Q$ under the composite policy over options.
Thus, the two terms of Eq.~(\ref{eq:transformer}) jointly instantiate the SMDP Bellman expectation equations for $Q$ and $V$.
Moreover, since every prefix of an executed chunk is itself a valid shorter option, a single trajectory provides valid targets for all durations at once, which is analogous to intra-option value learning \citep{sutton1998intra}.
Unlike prior option learning methods, the options in ACH are not chosen from a predefined set; instead, the policy generates a new continuous action sequence at every state, so the option space is uncountable.
Even in this setting, ACH efficiently determines how long to execute each generated sequence by comparing the values of all its prefixes, which the causal Transformer-based value function evaluates in a single forward pass.

\par\noindent\textbf{Action repetition}\quad\ignorespaces
Action repetition methods study how long an agent should keep executing the same action, which is closely related to selecting the chunk length.
Frame skipping, which repeats each action for a fixed number of steps, is a common practice \citep{mnih2015human}, and subsequent works learn the repetition length adaptively \citep{lakshminarayanan2017dynamic,sharma2017learning,biedenkapp2021temporl}, select a persistence level for offline RL \citep{metelli2020control}, or decide at each step whether to keep the previous action \citep{yu2021taac}.
Other works generate multi-step action plans and decide when to replace them, for example by learning when to update the plan \citep{vezhnevets2016strategic} or by switching to a newly generated plan when its value is higher than that of the current plan \citep{zhang2022generative}.
In contrast to these methods, which decide whether to keep or replace the current plan, ACH compares the values of all prefixes of a generated chunk to select how long to execute it.

\par\noindent\textbf{Hierarchical RL}\quad\ignorespaces
Hierarchical RL decomposes decision making into multiple levels of temporal abstraction \citep{dayan1992feudal,dietterich2000hierarchical}.
Typically, a high-level policy sets subgoals or latent goals every few steps, and a low-level policy selects primitive actions to reach them \citep{vezhnevets2017feudal,nachum2018data}.
Such hierarchies have been shown to benefit mainly from improved exploration \citep{nachum2019does}.
Other works extract reusable skills from offline data \citep{ajay2020opal,pertsch2021accelerating}, and hierarchical methods that decompose tasks into subgoals achieve strong performance on long-horizon offline goal-conditioned tasks \citep{park2023hiql,horizon}.
More recently, several works have combined hierarchy with action chunking, for example by pairing a high-level planner that predicts latent subgoals with a low-level critic conditioned on fixed-length action chunks \citep{jawaid2026offline}, or by predicting action chunks at multiple frequencies and balancing them during execution with an entropy-guided mechanism in imitation learning \citep{zhang2026hipolicy}.
Like ACH, these approaches aim to combine long-horizon structure with fine-grained control, but they either fix the chunk length or select it with an uncertainty-based heuristic rather than value estimation.
In contrast, ACH keeps a single-level policy and uses the values of all chunk prefixes to select the chunk length at each state, adapting it during both training and inference.

\subsection{Theoretical Analysis}
\label{appendix:theory}
We formalize what the objectives of ACH learn in the idealized setting of exact (tabular) policy evaluation and improvement.
Because an action chunk is a semi-Markov option (Appendix~\ref{appendix:related}), decision making over chunks forms a Semi-MDP (SMDP) \citep[Theorem~1]{sutton1999between}, and the policy evaluation and improvement results below (Propositions~\ref{prop:evaluation} and~\ref{prop:improvement}) are instances of standard SMDP theory \citep{puterman94} tailored to our setting. We state and prove them in our notation for completeness, and then characterize the relation to the optimal value of the underlying MDP (Theorem~\ref{thm:pi} and Theorem~\ref{thm:opt_chunk}), when longer chunks are preferred, and what is learned under the behavior constraint (Theorem~\ref{thm:bc_pi}).

We restrict the analysis to deterministic transitions $s_{t+1}=f(s_t,a_t)$, under which the states and rewards along an action sequence are determined by the initial state and the actions alone, and discuss stochastic transitions at the end of this section.
Throughout the analysis, we assume bounded rewards and $0<\gamma<1$, yielding bounded value functions and the existence of a deterministic optimal policy $\pi^*$ of the corresponding MDP, and assume that the maximum over action chunks is attained. We denote the optimal value function of the MDP by $V^*$, which can be achieved by $\pi^*$.
For a state $s_t$ and an action sequence $\bm{a}_{t:t+n}=(a_t,\dots,a_{t+n})$,
we have $s_{t+k+1}=f(s_{t+k},a_{t+k})$ for $k=0, 1, \dotsc, n$,
and $R_t^n:=\sum_{k=0}^{n}\gamma^k r(s_{t+k},a_{t+k})$ for each  $0\le n \le h$, as in Eq.~(\ref{eq:transformer}).

\par\noindent\textbf{Composite policy}\quad\ignorespaces The overall procedure of ACH is captured by a composite policy $\mu:=(\pi_\theta,\pi_L)$, which samples a full chunk $\bm{a}'_{t:t+h}\sim\pi_\theta(\cdot|s_t)$ and an execution length $\ell\sim\pi_L(\cdot|s_t,\bm{a}'_{t:t+h})$ at each decision state $s_t$, executes $\bm{a}'_{t:t+\ell}$ open-loop for $\ell+1$ time steps, and samples its next chunk at $s_{t+\ell+1}$.
Assuming that both $\pi_\theta$ and $\pi_L$ are fixed, the value functions $Q^\mu$ and $V^\mu$ of the policy $\mu$ (see Eq.~(\ref{eq:chunkQ})) satisfy
\begin{equation}
\begin{aligned}
    Q^\mu(s_t,\bm{a}_{t:t+n}) &= R_t^n + \gamma^{n+1}V^\mu(s_{t+n+1}), ~~\forall n \in \{0,1,\cdots,h\}\\
    V^\mu(s_t) &= \mathbb{E}_{\bm{a}'_{t:t+h}\sim\pi_\theta(\cdot|s_t),\,\ell\sim\pi_L(\cdot|s_t,\bm{a}'_{t:t+h})}\big[Q^\mu(s_t,\bm{a}'_{t:t+\ell})\big],
\end{aligned}
    \label{eq:smdp_det}
\end{equation}
which are the SMDP Bellman expectation equations for $Q$ and $V$ discussed above, specialized to deterministic transitions.
Substituting the first equation of (\ref{eq:smdp_det}) into the second gives the operator
\begin{equation}
    (T^\mu V)(s_t)=\mathbb{E}_{(\bm{a}'_{t:t+h},\ell) \sim \mu}\big[R_t^\ell+\gamma^{\ell+1}V(s_{t+\ell+1})\big].
    \label{eq:operator}
\end{equation}
Note that the difference between Eq.(\ref{eq:operator}) and the standard MDP Bellman expectation operator is that we have an expectation over $\ell$ instead of $\ell$ being fixed to 0, i.e., a single action.

\vspace{0.5em}

\begin{proposition}[Policy evaluation]
\label{prop:evaluation} The SMDP Bellman equation (\ref{eq:smdp_det}) has a unique bounded solution $(Q^\mu,V^\mu)$, and this corresponds to the unique fixed point of the iterations:
\begin{equation*}
    Q(s_t,\bm{a}_{t:t+n})\leftarrow R_t^n+\gamma^{n+1}V(s_{t+n+1}),\qquad V(s_t)\leftarrow\mathbb{E}_{\bm{a}'_{t:t+h},\ell}\big[Q(s_t,\bm{a}'_{t:t+\ell})\big],
\end{equation*}
which corresponds to the targets of Eq.~(\ref{eq:transformer}) in the mean square error (MSE) loss form. Applying these two updates over the full state space in an alternating fashion converges to $(Q^\mu,V^\mu)$.

\end{proposition}
\begin{proof}
For any bounded $V$ and $V'$, the reward terms in (\ref{eq:operator}) cancel, and since $\ell\ge0$,
\begin{equation*}
\begin{aligned}
    \big|(T^\mu V)(s_t)-(T^\mu V')(s_t)\big|
    &=\Big|\mathbb{E}\big[\gamma^{\ell+1}\big(V(s_{t+\ell+1})-V'(s_{t+\ell+1})\big)\big]\Big|\\
    &\le\mathbb{E}\big[\gamma^{\ell+1}\big|V(s_{t+\ell+1})-V'(s_{t+\ell+1})\big|\big]
    \le\gamma\|V-V'\|_\infty.
\end{aligned}
\end{equation*}
Thus $T^\mu$ is a $\gamma$-contraction, and by the Banach fixed-point theorem it has a unique fixed point, which is $V^\mu$. The first equation of (\ref{eq:smdp_det}) then determines $Q^\mu$ uniquely. Applying the two updates is equivalent to applying $T^\mu$ to $V$, so the iteration converges to $V^\mu$ and, in turn, $Q$ converges to $Q^\mu$.

\end{proof}

\begin{proposition}[Policy improvement]
\label{prop:improvement}
Let $\mu'=(\pi_\theta',\pi_L')$ be a composite policy such that
\begin{equation}
    \mathbb{E}_{\bm{a}'_{t:t+h}\sim\pi_\theta'(\cdot|s_t),\,\ell\sim\pi_L'(\cdot|s_t,\bm{a}'_{t:t+h})}\big[Q^\mu(s_t,\bm{a}'_{t:t+\ell})\big]\ge V^\mu(s_t)
    \label{eq:improve_cond}
\end{equation}
for all $s_t$. Then, $V^{\mu'}(s_t)\ge V^\mu(s_t)$ for all $s_t$. In particular, (\ref{eq:improve_cond}) holds (i) when $\pi_\theta'=\pi_\theta$ and $\pi_L'$ is the greedy selection used at evaluation, which selects $\ell=\arg\max_n Q^\mu(s_t,\bm{a}'_{t:t+n})$ with probability one, and (ii) when $\pi_L'$ is this greedy selection and $\pi_\theta'$ generates a chunk that maximizes $\max_n Q^\mu(s_t,\bm{a}'_{t:t+n})$ over $\bm{a}'_{t:t+h}$.
\end{proposition}
\begin{proof}
By (\ref{eq:smdp_det}), $(T^{\mu'}V^\mu)(s_t)$ equals the left-hand side (LHS) of (\ref{eq:improve_cond}), so $T^{\mu'}V^\mu\ge V^\mu$. Since $T^{\mu'}$ is monotone, i.e., $V\ge V'$ implies $T^{\mu'}V\ge T^{\mu'}V'$, as (\ref{eq:operator}) adds rewards to a nonnegative multiple of $V$, applying it repeatedly gives $(T^{\mu'})^kV^\mu\ge V^\mu$ for all $k$, and the LHS converges to $V^{\mu'}$ by Proposition~\ref{prop:evaluation}.
The right-hand side (RHS) of (\ref{eq:improve_cond}) is the average of $Q^\mu(s_t,\bm{a}'_{t:t+\ell})$ over $\bm{a}'_{t:t+h}\sim\pi_\theta$ and $\ell\sim\pi_L$. In case (i), the LHS is $\mathbb{E}_{\bm{a}'_{t:t+h}\sim\pi_\theta}[\max_n Q^\mu(s_t,\bm{a}'_{t:t+n})]$, and in case (ii) it is $\max_{\bm{a}'_{t:t+h}}\max_n Q^\mu(s_t,\bm{a}'_{t:t+n})$. Both are at least as large as the average.
\end{proof}

\par\noindent\textbf{Optimality operator}\quad\ignorespaces To capture the SMDP situation, we recursively define the prefix transition function
$\tilde{f}\colon \mathcal{S}\times\mathcal{A}^{h+1}\times \{0, 1, \dotsc, h\}
\to \mathcal{S}$ based on the transition function $s_{t+1}=f(s_t,a_t)$ as
\begin{equation}
    \tilde{f}(s, \va_{t:t+h}, n)=\begin{cases}
        f(s, a_t)
        & \textnormal{if } n=0,\\
        f(\tilde{f}(s, \va_{t:t+h}, n-1), a_{t+n})
        & \textnormal{otherwise,}
    \end{cases}
\end{equation}
and the discounted prefix return function
$\tilde{R}\colon \mathcal{S}\times \mathcal{A}^{h+1}\times \{0, 1, \dotsc, h\}
\to \R$ as
\begin{equation}
    \tilde{R}(s, \va_{t:t+h}, n)=\begin{cases}
        r(s, a_t)
        & \textnormal{if } n=0,\\
        \tilde{R}(s, \va_{t:t+h}, n-1)
        +\gamma^n r(\tilde{f}(s, \va_{t:t+h}, n-1), a_{t+n})
        & \textnormal{otherwise.}
    \end{cases}
\end{equation}
Note that $s_{t+n+1}=\tilde{f}(s_t, \va_{t:t+h}, n)$
and $R_t^n=\tilde{R}(s_t, \va_{t:t+h}, n)$.
Define the composite optimality operator
\begin{equation}%
    (T^*V)(s_t)=\max_{\bm{a}'_{t:t+h}}\ \max_{n\in\{0,\dots,h\}}
    \left[R_t^n+\gamma^{n+1}V(s_{t+n+1})\right],
    \label{eq:opt_operator}
\end{equation}
which replaces the expectation over $(\bm{a}'_{t:t+h},\ell)$ in (\ref{eq:operator}) with maximum.
We can see that the expression inside the maximum
is a function of $s_t$, $\va_{t:t+h}$, and $n$,
which we will denote by
$\tilde{V}\colon \mathcal{S}\times \mathcal{A}^{h+1}\times \{0, 1, \dotsc, h\}\to\R$:
\begin{equation}
    \tilde{V}(s_t, \va_{t:t+h}, n)
    :=\tilde{R}(s_t, \va_{t:t+h}, n)
    +\gamma^{n+1}(V\circ \tilde{f})(s_t, \va_{t:t+h}, n).
    \label{eq:tilde_v}
\end{equation}
Using $\tilde{V}$, from  Eqs. (\ref{eq:opt_operator}) and (\ref{eq:operator}) we have 
\[
    (T^*V)(s_t)=\max_{\va'_{t:t+h},\,n}\tilde{V}(s_t, \va'_{t:t+h}, n),
\]
and
\[
    (T^\mu V)(s_t)=\mathbb{E}_{\va'_{t:t+h},\ell}\big[\tilde{V}(s_t, \va'_{t:t+h}, \ell)\big].
\]
We write $\tilde{V}'$, $\tilde{V}^*$, and $\tilde{V}^\mu$ for the functions defined in the same way from $V'$, $V^*$, and $V^\mu$, respectively.
Note that $\tilde{V}^\mu(s_t, \va_{t:t+h}, n)=Q^\mu(s_t,\va_{t:t+n})$ by (\ref{eq:smdp_det}).

\vspace{0.5em}

\begin{lemma}[Optimality operator]
\label{lem:opt_operator}
$T^*$ is a monotone $\gamma$-contraction whose unique bounded fixed point is $V^*$. Moreover, $V^\mu\le V^*$ for every composite policy $\mu$.
\end{lemma}
\begin{proof}
For any bounded $V$ and $V'$, by (\ref{eq:opt_operator}) and $|\max_x F(x)-\max_x G(x)|\le\sup_x|F(x)-G(x)|$ with $x=(\va'_{t:t+h},n)$,
\begin{equation*}
\begin{aligned}
    \big|(T^*V)(s_t)-(T^*V')(s_t)\big|
    &=\left|
    \max_{\va'_{t:t+h},\,n}\tilde{V}(s_t, \va'_{t:t+h}, n)
    -\max_{\va'_{t:t+h},\,n}\tilde{V}'(s_t, \va'_{t:t+h}, n)
    \right|\\
    &\le\sup_{\va'_{t:t+h},\,n}\left|
    \tilde{V}(s_t, \va'_{t:t+h}, n) -\tilde{V}'(s_t, \va'_{t:t+h}, n)
    \right|\\
    &=\sup_{\bm{a}'_{t:t+h},\,n}\Big|\big[R_t^n+\gamma^{n+1}V(s_{t+n+1})\big]-\big[R_t^n+\gamma^{n+1}V'(s_{t+n+1})\big]\Big|\\
    &=\sup_{\bm{a}'_{t:t+h},\,n}\gamma^{n+1}\big|V(s_{t+n+1})-V'(s_{t+n+1})\big|\\
    &\le\gamma\|V-V'\|_\infty,
\end{aligned}
\end{equation*}
where the last inequality uses $\gamma^{n+1}\le\gamma$ for $n\ge0$ and $|V(s)-V'(s)|\le\|V-V'\|_\infty$ for every state $s$. Thus, $T^*$ is a $\gamma$-contraction with a unique bounded fixed point. It is also monotone, since $V\ge V'$ implies $\tilde{V}\ge\tilde{V}'$ by (\ref{eq:tilde_v}).

We next show that the fixed point of $T^*$ is $V^*$.
We begin by showing that $\tilde{V}^*(s_t,\va'_{t:t+h},n)\le V^*(s_t)$ for any chunk $\va'_{t:t+h}$ and any $n$.
Since $V^*$ satisfies the Bellman optimality equation of the MDP, $V^*(s)=\max_a\left[r(s,a)+\gamma V^*(f(s,a))\right]$, we have
\begin{equation}
    r(s,a)+\gamma V^*(f(s,a))\le V^*(s) \label{eq:one_step_bound}
\end{equation}
for every state $s$ and action $a$.
For $n=0$, the claim is (\ref{eq:one_step_bound}) with $s=s_t$ and $a=a'_t$.
For $n\ge1$, the recursive definitions of $\tilde{R}$ and $\tilde{f}$ give
\begin{equation*}
\begin{aligned}
    \tilde{V}^*(s_t,\va'_{t:t+h},n)
    &=\tilde{R}(s_t,\va'_{t:t+h},n-1)+\gamma^n\left[r(s_{t+n},a'_{t+n})+\gamma V^*(f(s_{t+n},a'_{t+n}))\right]\\
    &\le\tilde{R}(s_t,\va'_{t:t+h},n-1)+\gamma^nV^*(s_{t+n})\\
    &=\tilde{V}^*(s_t,\va'_{t:t+h},n-1)\\
    &\le V^*(s_t),
\end{aligned}
\end{equation*}
where the first inequality is (\ref{eq:one_step_bound}) with $s=s_{t+n}$ and $a=a'_{t+n}$ and the second inequality is the induction hypothesis.
By induction on $n$, $\tilde{V}^*(s_t,\va'_{t:t+h},n)\le V^*(s_t)$ for all $n$.
Because $(T^*V^*)(s_t)=\max_{\va'_{t:t+h},\,n}\tilde{V}^*(s_t,\va'_{t:t+h},n)$, taking the maximum over $\va'_{t:t+h}$ and $n$ gives $T^*V^*\le V^*$.

Conversely, the optimality of $\pi^*$ means that $V^*=V^{\pi^*}$, and hence
\begin{equation}
    r(s,\pi^*(s))+\gamma V^*(f(s,\pi^*(s)))=V^*(s) \label{eq:one_step_eq}
\end{equation}
for every state $s$, i.e., (\ref{eq:one_step_bound}) holds with equality for $a=\pi^*(s)$.
Thus, $\tilde{V}^*(s_t,\va'_{t:t+h},0)=V^*(s_t)$ for any chunk $\va'_{t:t+h}$ with $a'_t=\pi^*(s_t)$.
The maximum of $\tilde{V}^*(s_t,\cdot,\cdot)$ over all chunks and lengths, which is $(T^*V^*)(s_t)$, is therefore at least $V^*(s_t)$, i.e., $T^*V^*\ge V^*$. Together with $T^*V^*\le V^*$, this gives $T^*V^*=V^*$.

It follows that $V^\mu\le V^*$ for every composite policy $\mu$.
Note that a composite policy need not be Markov with respect to the state of the MDP, since the action it executes at time $t+p$ can be determined by the chunk generated at the earlier decision state $s_t$ rather than by the current state $s_{t+p}$. We therefore verify this bound directly.
Since $(T^\mu V^*)(s_t)$ is an expectation of $\tilde{V}^*(s_t,\cdot,\cdot)$ and $(T^*V^*)(s_t)$ is its maximum, $T^\mu V^*\le T^*V^*=V^*$.
By the monotonicity of $T^\mu$, applying it repeatedly gives $(T^\mu)^jV^*\le V^*$ for all $j$, and the LHS converges to $V^\mu$ by Proposition~\ref{prop:evaluation}.
Hence $V^\mu\le V^*$.
\end{proof}

We now consider policy iteration over composite policies, which starts from any $\mu_0$ and alternates computing $Q^{\mu_k}$ (Proposition~\ref{prop:evaluation}) and constructing $\mu_{k+1}$ that selects the chunk and its length maximizing $Q^{\mu_k}$, as in case (ii) of Proposition~\ref{prop:improvement}.

\begin{theorem}[Policy iteration]
\label{thm:pi}
Policy iteration satisfies $V^{\mu_0}\le V^{\mu_1}\le\cdots\le V^*$ and $V^{\mu_k}\to V^*$ in the uniform norm.
\end{theorem}
\begin{proof}
By the definition of case (ii), $\mu_{k+1}$ selects a chunk and a length that maximize $Q^{\mu_k}(s_t,\va'_{t:t+n})=\tilde{V}^{\mu_k}(s_t,\va'_{t:t+h},n)$, that is,
\begin{equation}
    T^{\mu_{k+1}}V^{\mu_k}=T^*V^{\mu_k}.
    \label{eq:greedy_step}
\end{equation}
Since $(T^{\mu_k}V)(s_t)$ is an expectation of $\tilde{V}(s_t,\cdot,\cdot)$ and $(T^*V)(s_t)$ is its maximum, $T^*V\ge T^{\mu_k}V$ for every $V$. Moreover, $V^{\mu_k}$ is the fixed point of $T^{\mu_k}$ by Proposition~\ref{prop:evaluation}. Together with (\ref{eq:greedy_step}),
\begin{equation*}
    T^{\mu_{k+1}}V^{\mu_k}\ =\ T^*V^{\mu_k}\ \ge\ T^{\mu_k}V^{\mu_k}\ =\ V^{\mu_k}.
\end{equation*}
Since $(T^{\mu_{k+1}}V^{\mu_k})(s_t)$ is the LHS of (\ref{eq:improve_cond}) for $\mu'=\mu_{k+1}$ and $\mu=\mu_k$, this shows that $\mu_{k+1}$ satisfies (\ref{eq:improve_cond}). As in the proof of Proposition~\ref{prop:improvement}, applying the monotone operator $T^{\mu_{k+1}}$ repeatedly to $T^{\mu_{k+1}}V^{\mu_k}\ge V^{\mu_k}$ gives a nondecreasing sequence $(T^{\mu_{k+1}})^jV^{\mu_k}$ that converges to $V^{\mu_{k+1}}$, and hence
\begin{equation*}
    V^{\mu_{k+1}}\ \ge\ T^{\mu_{k+1}}V^{\mu_k}\ =\ T^*V^{\mu_k}\ \ge\ V^{\mu_k}.
\end{equation*}
In particular, $V^{\mu_k}$ is nondecreasing in $k$. We next show by induction that $V^{\mu_k}\ge(T^*)^kV^{\mu_0}$. The case $k=0$ is trivial. If $V^{\mu_k}\ge(T^*)^kV^{\mu_0}$, then the monotonicity of $T^*$ (Lemma~\ref{lem:opt_operator}) and the inequality above give $V^{\mu_{k+1}}\ge T^*V^{\mu_k}\ge(T^*)^{k+1}V^{\mu_0}$. Combining this with $V^{\mu_k}\le V^*$ from Lemma~\ref{lem:opt_operator},
\begin{equation*}
    (T^*)^kV^{\mu_0}\ \le\ V^{\mu_k}\ \le\ V^*\quad\text{for all }k.
\end{equation*}
Subtracting each side from $V^*$ gives $0\le V^*-V^{\mu_k}\le V^*-(T^*)^kV^{\mu_0}$. Since $T^*$ is a $\gamma$-contraction with fixed point $V^*$ by Lemma~\ref{lem:opt_operator}, applying it $k$ times gives $\|V^*-(T^*)^kV^{\mu_0}\|_\infty=\|(T^*)^kV^*-(T^*)^kV^{\mu_0}\|_\infty\le\gamma^k\|V^*-V^{\mu_0}\|_\infty$. Hence $\|V^*-V^{\mu_k}\|_\infty\le\gamma^k\|V^*-V^{\mu_0}\|_\infty$, and $V^{\mu_k}\to V^*$.
\end{proof}

\begin{theorem}[Optimal chunks]
\label{thm:opt_chunk}
There exists a composite policy $\mu^*$ with $V^{\mu^*}=V^*$ whose chunk $\bm{a}^*_{t:t+h}$ at each state $s_t$ satisfies $Q^{\mu^*}(s_t,\bm{a}^*_{t:t+n})=V^*(s_t)$ for every $n=0, 1, \dotsc, h$.
\end{theorem}
\begin{proof}
For each state $s_t$, let $\va^*_{t:t+h}$ be the chunk obtained by rolling out $\pi^*$ from $s_t$, i.e., $a^*_t=\pi^*(s_t)$ and $a^*_{t+p+1}=\pi^*(f(s_{t+p}, a^*_{t+p}))$ for $p=0,1,\dotsc,h-1$, and let $\mu^*$ be the composite policy that generates this chunk at every decision state, with any $\pi_L$.
We show by induction on $n$ that
\begin{equation}
    \tilde{V}^*(s_t,\va^*_{t:t+h},n)=V^*(s_t)
    \label{eq:opt_chunk}
\end{equation}
for every $n=0, 1, \dotsc, h$.
The case $n=0$ is (\ref{eq:one_step_eq}) with $s=s_t$.
For $n\ge1$,
\begin{equation*}
\begin{aligned}
    \tilde{V}^*(s_t,\va^*_{t:t+h},n)
    &=\tilde{R}(s_t,\va^*_{t:t+h},n-1)+\gamma^n\left[r(s_{t+n},a^*_{t+n})+\gamma V^*(f(s_{t+n},a^*_{t+n}))\right]\\
    &=\tilde{R}(s_t,\va^*_{t:t+h},n-1)+\gamma^nV^*(s_{t+n})\\
    &=\tilde{V}^*(s_t,\va^*_{t:t+h},n-1)\\
    &=V^*(s_t),
\end{aligned}
\end{equation*}
where the second equality is (\ref{eq:one_step_eq}) with $s=s_{t+n}$ and $\pi^*(s_{t+n})=a_{t+n}^*$, and the last equality is the induction hypothesis.
By (\ref{eq:opt_chunk}), $(T^{\mu^*}V^*)(s_t)=\mathbb{E}_{\ell\sim\pi_L}\big[\tilde{V}^*(s_t,\va^*_{t:t+h},\ell)\big]=V^*(s_t)$, so $V^*$ is a bounded fixed point of $T^{\mu^*}$, and $V^{\mu^*}=V^*$ by the uniqueness in Proposition~\ref{prop:evaluation}.
Consequently, $\tilde{V}^{\mu^*}=\tilde{V}^*$, and $Q^{\mu^*}(s_t,\va^*_{t:t+n})=\tilde{V}^{\mu^*}(s_t,\va^*_{t:t+h},n)=V^*(s_t)$ for every $n$.
\end{proof}

Theorem~\ref{thm:pi} shows that alternating the policy evaluation of Proposition~\ref{prop:evaluation} and the greedy improvement of Proposition~\ref{prop:improvement}~(ii) converges to the optimal value.
This is consistent with the observation of \citet{sutton1999between} that introducing options does not change the optimal value function of the underlying MDP when the primitive actions remain available, which holds here since a prefix of length one is a primitive action, and that planning over a set of options converges to the optimal value function over that set. Theorem~\ref{thm:opt_chunk} shows that restricting the policy to macro-actions does not lose optimality, since every prefix of the chunk generated by $\mu^*$ attains the optimal value. For $\mu^*$, neither of the two factors in Section~\ref{subsec:value_based_optimality}, namely the uncertainty of predicted future states and the policy stochasticity jittering loss, is present: each chunk follows the trajectory of the optimal policy exactly, so the reached states never deviate from those predicted at the decision state, and the policy has no sampling noise, so replanning introduces no jitter. The chunk length matters for a stochastic or suboptimal policy, such as the policy being learned or a policy regularized toward the behavior policy, as in Eq.~(\ref{eq:policy}).

To see when a longer chunk is preferred, note that by (\ref{eq:smdp_det}), for any composite policy $\mu$, action sequence $\bm{a}_{t:t+n}$, and $n\ge1$, we have
\begin{eqnarray}
    Q^\mu(s_t,\bm{a}_{t:t+n})-Q^\mu(s_t,\bm{a}_{t:t+n-1})&=&\gamma^n\big(r(s_{t+n},a_{t+n})+\gamma V^\mu(s_{t+n+1})-V^\mu(s_{t+n})\big)\nonumber\\
    &=& \gamma^n(Q^\mu(s_{t+n},a_{t+n})- V^\mu(s_{t+n})),
    \label{eq:length_diff}
\end{eqnarray}
where $Q^\mu(s_{t+n},a_{t+n})=r(s_{t+n},a_{t+n})+\gamma V^\mu(s_{t+n+1})$ is the value of executing the single action $a_{t+n}$ at $s_{t+n}$ and following $\mu$ afterward.
Thus, extending a chunk by one step increases the value exactly when the pre-planned action $a_{t+n}$ is better than replanning at $s_{t+n}$, i.e., drawing a new chunk from $\pi_\theta$ and a new length from $\pi_L$ since $ V^\mu(s_{t+n}) = \mathbb{E}_{\bm{a}'_{t+n:t+n+h}\sim\pi_\theta(\cdot|s_{t+n}),\,\ell\sim\pi_L(\cdot|s_{t+n},\bm{a}'_{t+n:t+n+h})}\big[Q^\mu(s_{t+n},\bm{a}'_{t+n:t+n+\ell})\big]$. 
This comparison is analogous to the interruption of options \citep{sutton1999between}, where an option is terminated early when continuing it is valued less than selecting a new option. In our ACH case, the comparison is made in advance at the decision state $s_t$ across the prefixes of the generated chunk, rather than during execution.
When the reached state is consistent with the plan, replanning can introduce the sampling noise of the stochastic policy $\pi_\theta$ without improving the action, so the pre-planned action can be better and longer chunks are favored.
When the pre-planned action, which is generated from $s_t$ alone, no longer suits the reached state, replanning is better and shorter chunks are favored.
These two cases correspond to the policy stochasticity jittering loss and the uncertainty of predicted future states in Section~\ref{subsec:value_based_optimality}, respectively, and comparing the values of all prefixes allows ACH to balance them at each state, which is consistent with the improved sample efficiency observed in Section~\ref{sec:experiments}.

\par\noindent\textbf{Behavior-constrained policy iteration}\quad\ignorespaces
To match the actor objective in Eq.~(\ref{eq:policy}), which includes the behavior cloning (BC) term, we next analyze policy iteration in which the BC term takes the form of a constraint.
As in Eq.~(\ref{eq:policy}), let the chunk policy be a map $\bm{a}'_{t:t+h}=\pi_\theta(s_t,\mathbf{z})$ of the state and a noise $\mathbf{z}\sim p(\mathbf{z})$, so that the expectation over $\bm{a}'_{t:t+h}\sim\pi_\theta(\cdot|s_t)$ in (\ref{eq:operator}) is the expectation over $\mathbf{z}$, and let $\beta_\psi(s_t,\mathbf{z})$ denote the chunk generated by the behavior policy from the same noise.
For a level $\kappa\colon\mathcal{S}\to[0,\infty]$, we define the set of chunk policies that are feasible at $s_t$ as
\begin{equation}
    \Pi_\kappa(s_t)=\Big\{\pi(s_t,\cdot)\,:\,\mathbb{E}_{\mathbf{z}}\big[\|\pi(s_t,\mathbf{z})-\beta_\psi(s_t,\mathbf{z})\|_2^2\big]\le\kappa(s_t)\Big\},
    \label{eq:bc_set}
\end{equation}
which is the BC term of Eq.~(\ref{eq:policy}) in the form of a hard constraint.
Let $\mathcal{F}_\kappa$ be the set of composite policies $\mu=(\pi_\theta,\pi_L)$ with $\pi_\theta(s_t,\cdot)\in\Pi_\kappa(s_t)$ for all $s_t$, with no restriction on $\pi_L$.
The behavior policy satisfies the constraint with zero cost, so $\mathcal{F}_\kappa$ is nonempty.
We define the behavior-constrained optimality operator
\begin{equation}
    (T_\kappa V)(s_t)=\max_{\pi(s_t,\cdot)\in\Pi_\kappa(s_t)}\ \mathbb{E}_{\mathbf{z}}\Big[\max_{n\in\{0,\dots,h\}}\tilde{V}\big(s_t,\pi(s_t,\mathbf{z}),n\big)\Big],
    \label{eq:bc_operator}
\end{equation}
where we assume that the maximum over $\Pi_\kappa(s_t)$ is attained, as for $T^*$.
Behavior-constrained policy iteration keeps $\beta_\psi$ and $\kappa$ fixed, so that every iterate is selected from the same set $\mathcal{F}_\kappa$. It starts from any $\mu_0\in\mathcal{F}_\kappa$, e.g., the behavior policy with any $\pi_L$, and alternates two steps.
The evaluation step computes $Q^{\mu_k}$ (Proposition~\ref{prop:evaluation}).
The improvement step is case (ii) of Proposition~\ref{prop:improvement} with the chunk policy restricted to the BC constraint. It constructs $\mu_{k+1}$ by choosing, at every state, a chunk policy in $\Pi_\kappa(s_t)$ that maximizes the expected value of the best prefix, $\mathbb{E}_{\mathbf{z}}[\max_nQ^{\mu_k}(s_t,\va'_{t:t+n})]$ with $\va'_{t:t+h}=\pi(s_t,\mathbf{z})$, and by selecting the length greedily.
Since $\tilde{V}^{\mu_k}(s_t,\va'_{t:t+h},n)=Q^{\mu_k}(s_t,\va'_{t:t+n})$, this chunk policy attains the maximum in $(T_\kappa V^{\mu_k})(s_t)$.
The following theorem shows that this iteration improves monotonically and converges to the best value attainable under the BC constraint.

\begin{theorem}[Behavior-constrained policy iteration]
\label{thm:bc_pi}
$T_\kappa$ is a monotone $\gamma$-contraction, and its unique bounded fixed point $V^*_\kappa$ satisfies $V^*_\kappa(s_t)=\max_{\mu\in\mathcal{F}_\kappa}V^\mu(s_t)$ for all $s_t$, where the maximum is attained at every state by a single policy $\mu^*_\kappa\in\mathcal{F}_\kappa$.
Behavior-constrained policy iteration satisfies $V^{\mu_0}\le V^{\mu_1}\le\cdots\le V^*_\kappa$ and $V^{\mu_k}\to V^*_\kappa$ in the uniform norm.
\end{theorem}
\begin{proof}
For a bounded $V$ and $\pi(s_t,\cdot)\in\Pi_\kappa(s_t)$, let
\begin{equation*}
    G_V(\pi;s_t):=\mathbb{E}_{\mathbf{z}}\Big[\max_{n}\tilde{V}\big(s_t,\pi(s_t,\mathbf{z}),n\big)\Big]
\end{equation*}
denote the objective inside the maximum in (\ref{eq:bc_operator}), so that $(T_\kappa V)(s_t)=\max_{\pi(s_t,\cdot)\in\Pi_\kappa(s_t)}G_V(\pi;s_t)$.
For any bounded $V$ and $V'$, any $\pi(s_t,\cdot)\in\Pi_\kappa(s_t)$, and $\va'_{t:t+h}=\pi(s_t,\mathbf{z})$,
\begin{equation*}
\begin{aligned}
    \big|G_V(\pi;s_t)-G_{V'}(\pi;s_t)\big|
    &\le\mathbb{E}_{\mathbf{z}}\Big[\Big|\max_n\tilde{V}(s_t,\va'_{t:t+h},n)-\max_n\tilde{V}'(s_t,\va'_{t:t+h},n)\Big|\Big]\\
    &\le\mathbb{E}_{\mathbf{z}}\Big[\max_n\big|\tilde{V}(s_t,\va'_{t:t+h},n)-\tilde{V}'(s_t,\va'_{t:t+h},n)\big|\Big]\\
    &=\mathbb{E}_{\mathbf{z}}\Big[\max_n\gamma^{n+1}\big|V(s_{t+n+1})-V'(s_{t+n+1})\big|\Big]\\
    &\le\gamma\|V-V'\|_\infty,
\end{aligned}
\end{equation*}
where the second inequality uses $|\max_xF(x)-\max_xG(x)|\le\max_x|F(x)-G(x)|$, and the remaining steps are as in the proof of Lemma~\ref{lem:opt_operator}.
Applying $|\max_xF(x)-\max_xG(x)|\le\sup_x|F(x)-G(x)|$ once more, now with the maximum over $\Pi_\kappa(s_t)$, gives
\begin{equation*}
    \big|(T_\kappa V)(s_t)-(T_\kappa V')(s_t)\big|\le\sup_{\pi(s_t,\cdot)\in\Pi_\kappa(s_t)}\big|G_V(\pi;s_t)-G_{V'}(\pi;s_t)\big|\le\gamma\|V-V'\|_\infty
\end{equation*}
for every $s_t$. Thus, $T_\kappa$ is a $\gamma$-contraction, and by the Banach fixed-point theorem it has a unique bounded fixed point $V^*_\kappa$. It is also monotone, since $V\ge V'$ implies $\tilde{V}\ge\tilde{V}'$ by (\ref{eq:tilde_v}), hence $G_V(\pi;s_t)\ge G_{V'}(\pi;s_t)$ for every $\pi$, and hence $T_\kappa V\ge T_\kappa V'$.

For any $\mu=(\pi_\theta,\pi_L)\in\mathcal{F}_\kappa$ and any bounded $V$, replacing the expectation over $\ell\sim\pi_L$ with the maximum over $n$ and then the feasible $\pi_\theta(s_t,\cdot)$ with the maximum over $\Pi_\kappa(s_t)$ gives
\begin{equation}
    (T^\mu V)(s_t)=\mathbb{E}_{\mathbf{z}}\,\mathbb{E}_{\ell\sim\pi_L}\big[\tilde{V}(s_t,\pi_\theta(s_t,\mathbf{z}),\ell)\big]\ \le\ (T_\kappa V)(s_t).
    \label{eq:bc_dominate}
\end{equation}
In particular, $T^\mu V^*_\kappa\le T_\kappa V^*_\kappa=V^*_\kappa$. By the monotonicity of $T^\mu$, $(T^\mu)^jV^*_\kappa\le V^*_\kappa$ for all $j$, and the LHS converges to $V^\mu$ by Proposition~\ref{prop:evaluation}. Hence $V^\mu\le V^*_\kappa$ for every $\mu\in\mathcal{F}_\kappa$.
Let $\mu^*_\kappa\in\mathcal{F}_\kappa$ be the policy whose chunk policy attains the maximum in $(T_\kappa V^*_\kappa)(s_t)$ at every state and whose length maximizes $\tilde{V}^*_\kappa(s_t,\va'_{t:t+h},n)$ over $n$. Then $T^{\mu^*_\kappa}V^*_\kappa=T_\kappa V^*_\kappa=V^*_\kappa$, so $V^{\mu^*_\kappa}=V^*_\kappa$ by the uniqueness in Proposition~\ref{prop:evaluation}. Therefore, $V^*_\kappa=\max_{\mu\in\mathcal{F}_\kappa}V^\mu$.

For the iteration, the construction of $\mu_{k+1}$ gives $T^{\mu_{k+1}}V^{\mu_k}=T_\kappa V^{\mu_k}$, and since $\mu_k\in\mathcal{F}_\kappa$, (\ref{eq:bc_dominate}) gives $T_\kappa V^{\mu_k}\ge T^{\mu_k}V^{\mu_k}=V^{\mu_k}$.
These are the two properties of $T^*$ used in the proof of Theorem~\ref{thm:pi}. The rest of that proof applies with $T^*$ and $V^*$ replaced by $T_\kappa$ and $V^*_\kappa$, which gives $V^{\mu_{k+1}}\ge T_\kappa V^{\mu_k}\ge V^{\mu_k}$ and $(T_\kappa)^kV^{\mu_0}\le V^{\mu_k}\le V^*_\kappa$, and hence $\|V^*_\kappa-V^{\mu_k}\|_\infty\le\gamma^k\|V^*_\kappa-V^{\mu_0}\|_\infty$. Therefore, $V^{\mu_k}\to V^*_\kappa$.
\end{proof}

Theorem~\ref{thm:bc_pi} states what is learned when the BC constraint is present.
The values converge to $V^*_\kappa$, the best value among the composite policies whose chunks stay within the level $\kappa$ of the behavior chunks, and $Q^{\mu_k}(s_t,\bm{a}_{t:t+n})$ converges to $R_t^n+\gamma^{n+1}V^*_\kappa(s_{t+n+1})$, the value of executing the prefix and following $\mu^*_\kappa$ afterward.
With $\kappa=\infty$, $T_\kappa$ reduces to $T^*$ and Theorem~\ref{thm:pi} is recovered. With $\kappa=0$, the limit is the value of the behavior chunk policy whose length is selected greedily.

\par\noindent\textbf{Relation to the practical actor update}\quad\ignorespaces
Eq.~(\ref{eq:policy}) uses the BC term as a penalty with a fixed coefficient $\alpha$, rather than as a constraint with a level $\kappa$. The two forms are closely related.
Let $D_{\mathrm{BC}}(\pi;s_t):=\mathbb{E}_{\mathbf{z}}\big[\|\pi(s_t,\mathbf{z})-\beta_\psi(s_t,\mathbf{z})\|_2^2\big]$ denote the BC cost in (\ref{eq:bc_set}), and recall the objective $G_V(\pi;s_t)$ of $T_\kappa$ from the proof of Theorem~\ref{thm:bc_pi}.
A chunk policy that maximizes $G_{V^{\mu_k}}(\pi;s_t)-\alpha D_{\mathrm{BC}}(\pi;s_t)$ also maximizes $G_{V^{\mu_k}}(\pi;s_t)$ among the chunk policies whose BC cost does not exceed its own, since otherwise the penalized objective could be increased. Thus, each penalized step is a constrained step whose level $\kappa(s_t)$ is determined by $\alpha$.
Conversely, each constrained step is a penalized step with a suitable coefficient.

\begin{proposition}[Penalized form of the constrained step]
\label{prop:lagrange}
Assume that the noise distribution $p(\mathbf{z})$ has a density, as for the Gaussian noise in Eq.~(\ref{eq:policy}), and that the action space is bounded. Let the chunk policy at each state range over all measurable maps from the noise to chunks, rather than a fixed parameterization. For an iteration $k$ and a state $s_t$ with $0<\kappa(s_t)<\infty$, there exists $\alpha_k(s_t)\ge0$ such that any chunk policy that attains the maximum in $(T_\kappa V^{\mu_k})(s_t)$ also maximizes $G_{V^{\mu_k}}(\pi;s_t)-\alpha_k(s_t)D_{\mathrm{BC}}(\pi;s_t)$ over all such chunk policies $\pi(s_t,\cdot)$.
\end{proposition}
\begin{proof}
Fix $k$ and $s_t$, and write $G(\pi)=G_{V^{\mu_k}}(\pi;s_t)$, $D(\pi)=D_{\mathrm{BC}}(\pi;s_t)$, and $c=\kappa(s_t)$, which are expectations over $\mathbf{z}$ of bounded functions.

We first show that the set $\mathcal{S}$ of pairs $(G(\pi),D(\pi))$ over all chunk policies is convex. For chunk policies $\pi_1$ and $\pi_2$ and $\lambda\in[0,1]$, let $\pi_E$ follow $\pi_1$ for $\mathbf{z}\in E$ and $\pi_2$ otherwise. Since $p(\mathbf{z})$ has a density, no single noise value has positive probability, and the Lyapunov convexity theorem gives a set $E$ of noises with
\begin{equation*}
    G(\pi_E)=\lambda G(\pi_1)+(1-\lambda)G(\pi_2),\qquad D(\pi_E)=\lambda D(\pi_1)+(1-\lambda)D(\pi_2).
\end{equation*}

The constrained step thus maximizes $u$ over the convex set $\mathcal{S}$ subject to $v\le c$, and the behavior policy satisfies Slater's condition with $D=0<c$. Lagrangian duality therefore gives $\alpha_k(s_t)\ge0$ such that
\begin{equation*}
    \sup_\pi\big[G(\pi)-\alpha_k(s_t)(D(\pi)-c)\big]=G^\star,
\end{equation*}
where $G^\star$ is the optimal value of the constrained step \citep{paternain2019constrained}.
A chunk policy $\pi^\star$ that attains the maximum in $(T_\kappa V^{\mu_k})(s_t)$ satisfies $G(\pi^\star)=G^\star$ and $D(\pi^\star)\le c$, so $G(\pi^\star)-\alpha_k(s_t)(D(\pi^\star)-c)\ge G^\star$. Hence $\pi^\star$ maximizes $G-\alpha_k(s_t)D$.
\end{proof}

By Proposition~\ref{prop:lagrange}, a solution of each constrained improvement step of Theorem~\ref{thm:bc_pi} also solves a penalized best-prefix maximization with a coefficient $\alpha_k(s_t)$ that depends on the state and the iteration, as in primal-dual methods for constrained RL \citep{paternain2019constrained}.
With a single fixed $\alpha$, as in Eq.~(\ref{eq:policy}), however, the corresponding level is the BC cost of the penalized solution, which differs across states and changes as $Q^{\mu_k}$ changes over iterations, whereas Theorem~\ref{thm:bc_pi} assumes a fixed $\kappa$. Theorem~\ref{thm:bc_pi} thus describes what is learned when the BC coefficient is adapted for each state and iteration.

\par\noindent\textbf{Chunk lengths under the BC term}\quad\ignorespaces
Theorem~\ref{thm:bc_pi} characterizes the value learned under the BC term. We next examine which chunk lengths attain it. Unlike the unconstrained case of Theorem~\ref{thm:opt_chunk}, in which every prefix length attains the optimal value, the BC term can make the best chunk length depend on the state. We illustrate this with a greedy-length counterpart of Eq.~(\ref{eq:policy}), which uses the BC term as a penalty with a fixed coefficient $\alpha$ as in practice.

Consider the environment illustrated below. From a state $A$ or $B$, the agent always moves to a state $C$. At $C$, the episode ends with reward $1$ if the action is at most $0$ and with reward $0$ otherwise.
Actions are scalars in $[-1,1]$, and we write $\mathrm{L}=-1$ for the action that succeeds at $C$ and $\mathrm{R}=+1$ for the action that fails at $C$. Chunks have length two, so a chunk generated at $A$ or $B$ consists of an action at that state and an action at $C$.
The offline dataset contains two types of trajectories in equal proportion, $(\mathrm{L},\mathrm{L})$ from $A$ and $(\mathrm{L},\mathrm{R})$ from $B$. The behavior policy therefore generates $(\mathrm{L},\mathrm{L})$ at $A$ and $(\mathrm{L},\mathrm{R})$ at $B$. At $C$, where both types pass, it starts its chunk with $\mathrm{L}$ or $\mathrm{R}$ with probability $1/2$ each.

\begin{center}
\begin{tikzpicture}[>=stealth, st/.style={circle, draw, minimum size=7mm, inner sep=0pt}, ab/.style={rectangle, draw, rounded corners=2pt, minimum size=6mm, inner sep=2pt}]
\node[st] (A) at (0,0.8) {$A$};
\node[st] (B) at (0,-0.8) {$B$};
\node[st] (C) at (2.8,0) {$C$};
\node[ab] (G) at (6.0,1.0) {$r=1$};
\node[ab] (D) at (6.0,-1.0) {$r=0$};
\draw[->] (A) -- (C);
\draw[->] (B) -- (C);
\draw[->] (C) -- node[above, sloped, font=\small] {$a\le0$} (G);
\draw[->] (C) -- node[above, sloped, font=\small] {$a>0$} (D);
\node[anchor=east, font=\small] at ([xshift=-1mm]A.west) {$\beta_\psi(A)=(\mathrm{L},\mathrm{L})$};
\node[anchor=east, font=\small] at ([xshift=-1mm]B.west) {$\beta_\psi(B)=(\mathrm{L},\mathrm{R})$};
\end{tikzpicture}
\end{center}

At each state, the improvement step maximizes the expected best-prefix value minus $\alpha$ times the BC cost and selects the length greedily, with $\alpha>1$. Since this objective is an expectation over the behavior chunks, it suffices to compare, for each behavior chunk, keeping it with replacing some of its actions.

At $C$, replacing the first action $\mathrm{R}$ of a behavior chunk with an action $a\le0$ raises the reward from $0$ to $1$ but costs at least $\alpha>1$. The policy therefore keeps the behavior chunks at $C$, succeeds with probability $1/2$, and $V(C)=1/2$.
At $A$, the behavior chunk $(\mathrm{L},\mathrm{L})$ already succeeds, so the policy keeps it. Executing the full chunk gives $\gamma$, whereas executing only the first action and replanning at $C$ gives $\gamma V(C)=\gamma/2$, so the greedy length executes the full chunk.
At $B$, keeping the behavior chunk $(\mathrm{L},\mathrm{R})$ and replanning at $C$ gives $\gamma/2$, whereas replacing $\mathrm{R}$ and executing the full chunk gives at most $\gamma-\alpha<\gamma/2$. The policy therefore keeps the chunk, and the greedy length executes only the first action.

Hence the policy stays at the behavior chunks, but the best length differs between $A$ and $B$. Replanning at every step loses at $A$, and executing full chunks loses at $B$.
At $A$, the chunk commits to $\mathrm{L}$ at $C$, whereas replanning draws a new chunk from the policy at $C$, which mixes the two types of trajectories and remains stochastic because the BC term makes replacing $\mathrm{R}$ too costly. This illustrates the policy stochasticity jittering loss of Section~\ref{subsec:value_based_optimality}. At $B$, the pre-planned action $\mathrm{R}$ is worse than a new chunk.
This agrees with (\ref{eq:length_diff}), since $Q(C,\mathrm{L})-V(C)=1/2>0$ and $Q(C,\mathrm{R})-V(C)=-1/2<0$.

The above results assume exact policy evaluation and improvement under deterministic transitions, as in standard analyses of actor-critic methods, and the practical algorithm differs in using function approximation, the penalized BC term with a fixed $\alpha$, and soft length sampling. For stochastic transitions, the equations in Proposition~\ref{prop:evaluation} are recovered when, for every prefix in the data, the states that follow $(s_t,\bm{a}_{t:t+n})$ in the data have the same distribution as when the prefix is executed open-loop from $s_t$. This condition corresponds to the strong open-loop consistency of \citet{dqc} with zero error, and it holds for any data under deterministic transitions. We leave the analysis of the penalized update with a fixed $\alpha$ and a formal extension to stochastic transitions for future work.

\newpage
\section{Experimental Details}
\label{appendix:implementation}
In Section \ref{sec:experiments}, we validate the efficacy of the proposed method through comprehensive evaluations across 34 distinct tasks.
This section provides the specific details of the experiments used to obtain these results.
\subsection{Benchmarks and Tasks}
\label{appendix:benchmarks}
We conduct experiments on 30 tasks from OGBench \citep{ogbench} and 4 tasks from robomimic \citep{robomimic}.
Following standard protocols in offline-to-online RL~\citep{fql,fino}, we utilize the single-task variant across six environments: \texttt{antmaze-giant, antsoccer-arena, cube-quadruple, cube-triple, humanoidmaze-medium, puzzle-4x4}.
Each environment includes five distinct tasks.
\begin{itemize}
    \item Antmaze and Antsoccer: The objective is to control a robot with an 8-dimensional action space to reach a designated goal.
    \item Cube: A robotic arm with a 5-dimensional action space performs various pick-and-place tasks.
    \item Humanoidmaze: This task involves navigating a humanoid robot in a 21-dimensional action space to a goal.
    \item Puzzle: A robotic arm with a 5-dimensional action space presses buttons in a specific manner to solve the puzzle.
\end{itemize}
For the datasets, we employ the navigate dataset for antmaze, antsoccer, and humanoidmaze, which consist of 1M, 1M, and 2M transitions, respectively.
The play dataset is utilized for cube and puzzle, containing 3M transitions for cube-triple, 10M transitions for cube-quadruple, and 1M transitions for puzzle.

We further evaluate the method on four specific tasks within the robomimic benchmark.
All experiments in this domain utilize the multi-human dataset, which consists of 300 successful trajectories for each task.
\begin{itemize}
    \item lift: A single robotic arm with a 7-dimensional action space lifts a small cube.
    \item can: A single robotic arm with a 7-dimensional action space picks up a can and places it into a target bin.
    \item square: A single robotic arm with a 7-dimensional action space places a square nut onto a rod.
    \item transport: Two robotic arms with a combined 14-dimensional action space perform a collaborative maneuver. One arm lifts a hammer and passes it to the other arm, which then places the object into a target bin.
\end{itemize}

\subsection{Evaluation}
\label{appendix:evaluation}
We evaluate all baselines every 100K environment steps. Each evaluation measures the success rate over 50 episodes.
Regarding the training budget, 1M offline steps are sufficient for offline pre-training to converge on datasets with 1M or more transitions.
The 1M online steps match the scale of the offline data and are long enough for each method to meaningfully surpass the performance limited by the dataset.
This choice is also consistent with prior studies \citep{fql,qc}.

\subsection{Baselines}
\label{appendix:baselines}
For the comparative analysis, we implement the baselines by leveraging the experimental settings of FINO \citep{fino} and QC \citep{qc}.
Specifically, the implementations of RLPD, ReBRAC, FQL, and FINO are based on the paper and official code of FINO\footnote{https://github.com/CTID282/FINO}, while the BFN and QC follow the paper and official code of QC\footnote{https://github.com/ColinQiyangLi/qc}.
In accordance with previous studies \citep{fql,fino}, RLPD is configured with an update-to-data ratio of 1 and two value functions.
The number of sampled actions for BFN is fixed at 4.
The BC coefficients $\alpha$ for ReBRAC, FQL, FINO, and QC are summarized in Table \ref{tab:hyperparameter_alpha}, and the common hyperparameters utilized across all baselines are presented in Table \ref{tab:hyperparameter_shared}.

For hyperparameter tuning, we follow two rules.
First, when the original paper of a baseline covers the corresponding task, we use its reported hyperparameters.
Second, when a task is not covered, we construct three candidate values for the main hyperparameter by referring to the values used for similar tasks and report the best result.
In particular, since the original papers of RLPD, ReBRAC, and BFN do not cover OGBench \citep{ogbench}, we refer to the settings of FQL \citep{fql} and QC \citep{qc}.
The selection criterion is the final performance after online learning, averaged over seeds.
We apply the same protocol to ACH for a fair comparison.

\begin{table}[h!]
    \centering
    \vskip 0.1in
    \renewcommand{\arraystretch}{1.1}
    \vspace{-3ex}
    \caption{Shared Hyperparameters}
    \vspace{2ex}
    \resizebox{\textwidth}{!}{%
    \begin{tabular}{l|l}
    \toprule
    Hyperparameter     & Value                  \\ \midrule
    Learning rate                 & 0.0003    \\
    Optimizer                      & Adam \citep{adam}   \\
    Batch size                    & 256    \\
    Hidden dimensions                 & [512, 512, 512, 512]    \\
    Nonlinearity                      & GELU \citep{gelu}   \\
    Target network update rate                   & 0.005   \\
    Discount factor $\gamma$               & 0.99 (default), 0.995 (\texttt{antmaze, humanoidmaze, antsoccer})   \\
    Flow steps                    & 10   \\
    \bottomrule
    \end{tabular}%
    }
    \label{tab:hyperparameter_shared}
\end{table}

\begin{table}[h]
    \centering
    \vskip 0.1in
    \renewcommand{\arraystretch}{1.1}
    \vspace{-4ex}
    \caption{Task-specific BC coefficient $\alpha$ for each method. ReBRAC follows the loss normalization of its original implementation, so its coefficients are not directly comparable in scale to those of the other methods.}
    \vspace{2ex}
    \resizebox{0.65\textwidth}{!}{%
    \begin{tabular}{l|ccccc}
    \toprule
    Task & ReBRAC & FQL & FINO & QC & ACH \\
    \midrule
    antmaze-giant-navigate       & 0.003 & 10           & 10         & 30  & 30   \\
    antsoccer-arena-navigate     & 0.01  & 30           & 30         & 30  & 30   \\
    cube-quadruple-play-10M      & 0.1   & 300          & 300        & 100 & 300  \\
    cube-triple-play             & 0.1   & 300          & 300        & 100 & 300  \\
    humanoidmaze-medium-navigate & 0.01  & 100          & 100        & 100 & 100  \\
    puzzle-4x4-play              & 0.3   & 1000          & 1000       & 300 & 300  \\
    \midrule
    lift        & 0.03 & 10000 & 10000 & 10000 & 10000  \\
    can         & 0.03 & 10000 & 10000 & 10000 & 10000  \\
    square      & 0.1 & 10000 & 10000 & 10000 & 10000  \\
    transport   & 0.1 & 10000 & 10000 & 10000 & 1000 \\
    \bottomrule
    \end{tabular}%
    }
\label{tab:hyperparameter_alpha}
\end{table}

\subsection{ACH}
\label{appendix:ach}
The proposed method, ACH, is built upon QC as the backbone model and incorporates a transformer architecture into the value function.
This transformer configuration consists of an embedding dimension of 128, 4 heads, and 2 attention layers.
As with other baselines, ACH utilizes an environment-specific BC coefficient $\alpha$, which is provided in Table \ref{tab:hyperparameter_alpha}.
{For OGBench, we set the maximum chunk length to 10 except on \texttt{humanoidmaze}, where we use 5. The comparison in Appendix~\ref{appendix:max_chunk_length} shows that a maximum length of 10 yields lower performance on this environment. We conjecture that the relatively high-dimensional action space of \texttt{humanoidmaze} (21 dimensions) makes learning longer action chunks more challenging.}
For adaptive chunk length sampling at inference, the action chunk with the highest action value is selected to ensure consistent evaluation, thereby rendering the policy deterministic.
{In practice, $\pi_L$ in Eq.~(\ref{eq:sampling}) is computed with the normalized values $\tilde{Q}_\phi(s_t,\bm{a}^\prime_{t:t+\ell})=Q_\phi(s_t,\bm{a}^\prime_{t:t+\ell})/m_\phi(s_t,\bm{a}^\prime_{t:t+h})$ in place of $Q_\phi(s_t,\bm{a}^\prime_{t:t+\ell})$, where $m_\phi(s_t,\bm{a}^\prime_{t:t+h})=\frac{1}{h+1}\sum_{n=0}^{h}|Q_\phi(s_t,\bm{a}^\prime_{t:t+n})|$ is the average magnitude of the $h+1$ prefix values. This normalization reduces the dependence of the sampling distribution on the absolute Q-value scale while preserving the ranking of the prefixes, facilitating the use of a shared inverse temperature across tasks. We use $\beta=1$ in all experiments.}
The pseudo-code for ACH is presented in Algorithm \ref{alg}.

\begin{algorithm}[h!]
\caption{ACH: Adaptive Action CHunking for Offline-to-Online RL}
\label{alg}
\begin{algorithmic}[1]
    \STATE \textbf{Inputs:} offline dataset $D$, behavior policy $\beta_\psi$, one-step policy $\pi_\theta$, transformer-based value function $Q_{\phi}$, horizon length $h$, action chunk buffer $\mathcal{B} \gets \emptyset$
    \WHILE{in offline learning}
        \STATE Update $\psi, \theta, \phi$ based on Equations \ref{eq:flow}, \ref{eq:policy}, \ref{eq:transformer}.
    \ENDWHILE
    \WHILE{in online learning}
        \IF{{$\mathcal{B}$ is empty}}
            \STATE Sample full-length action chunk $\bm{a}_{t:t+h} \sim \pi_\theta(\cdot|s_t)$
            \STATE Compute values $\{Q_\phi(s_t,\bm{a}_{t:t}), \dots, Q_\phi(s_t, \bm{a}_{t:t+h})\}$
            \STATE Calculate sampling probability $\pi_L(\ell)$ from Equation \ref{eq:sampling}
            \STATE Select $\ell \sim \pi_L(\ell)$ \COMMENT{chunk length $\ell+1$}
            \STATE Store $\{a_t, a_{t+1}, \dots, a_{t+\ell}\}$ in $\mathcal{B}$
        \ENDIF
        \STATE Get $a_t$ from $\mathcal{B}$ and execute $a_t$
        \STATE Update $\psi, \theta, \phi$ based on Equations \ref{eq:flow}, \ref{eq:policy}, \ref{eq:transformer}.
    \ENDWHILE
\end{algorithmic}
\end{algorithm}

\subsection{VLA Experiments}
\label{appendix:vla}
\par\noindent\textbf{Benchmark}\quad\ignorespaces
We evaluate on SimplerEnv \citep{simplerenv} with two embodiments.
For the WidowX robot, we use seven tasks: \texttt{spoon\_on\_towel}, \texttt{carrot\_on\_plate}, \texttt{put\_eggplant\_in\_basket}, \texttt{stack\_cube}, \texttt{put\_eggplant\_in\_sink}, \texttt{close\_drawer}, and \texttt{open\_drawer}.
For the Google Robot, we use six tasks under the visual matching setting: \texttt{pick\_coke\_can}, \texttt{pick\_object}, \texttt{move\_near}, \texttt{open\_drawer}, \texttt{close\_drawer}, and \texttt{place\_in\_closed\_drawer}.

\par\noindent\textbf{Base policy}\quad\ignorespaces
We use GR00T N1.7 \citep{groot} as the base VLA, with the publicly released checkpoints for the WidowX\footnote{\url{https://huggingface.co/nvidia/GR00T-N1.7-SimplerEnv-Bridge}} and Google Robot\footnote{\url{https://huggingface.co/nvidia/GR00T-N1.7-SimplerEnv-Fractal}} embodiments.
The base VLA predicts action chunks of 8 steps by default, and we use this chunk length for all methods, which also serves as the maximum chunk length for the adaptive methods.
The VLA is kept frozen throughout training.

\par\noindent\textbf{QC and ACH}\quad\ignorespaces
For QC and ACH, we add a one-step policy on top of the frozen VLA.
The one-step policy takes the observation and the action chunk produced by the VLA as input and outputs a new action chunk.
Observations consist of an image and the proprioceptive state.
The image is processed by a four-layer convolutional network with 32, 64, 128, and 256 channels and a stride of 2, and its output is combined with the proprioceptive state into a 256-dimensional embedding.
This encoder is shared by the policy and the value function, which use the same network architecture.
QC is trained with Eqs.~(\ref{eq:onestep}) and (\ref{eq:qc}), and ACH is trained with Eqs.~(\ref{eq:transformer}) and (\ref{eq:policy}) and selects the chunk length with Eq.~(\ref{eq:sampling}) using $\beta=1$.
Both methods use a BC coefficient of $\alpha=1000$, a batch size of 128, and the Adam optimizer \citep{adam} with a learning rate of $3\times10^{-4}$.

We train a separate model for each task.
For offline training, we collect 1,000 episodes per task with the base policy and train for 100K gradient steps.
Online training then proceeds for 100K environment steps with an update-to-data ratio of 5, where each batch is sampled symmetrically from the offline dataset and the online replay buffer following RLPD \citep{rlpd}.
We use a discount factor of $\gamma=0.99$, a target network update rate of $\tau=0.005$, and an ensemble of two Q-functions.

\par\noindent\textbf{AAC and QC+AAC}\quad\ignorespaces
For AAC \citep{liang2026adaptive}, we use $N=20$ samples and select its movement floor hyperparameter for each embodiment from $\{0, 0.05, 0.1, 0.15, 0.3, 0.6\}$.
For QC+AAC, we sample 20 action chunks from the one-step policy of QC and determine the chunk length based on the rule of AAC.

\clearpage
\section{Additional Experiments}
\label{appendix:additional}
\subsection{Adaptive Chunk Length in Manipulation}
\label{appendix:manipulation}
\begin{figure}[h!]
    \centering
    \vspace{-2ex}
    \includegraphics[width=0.85\linewidth]{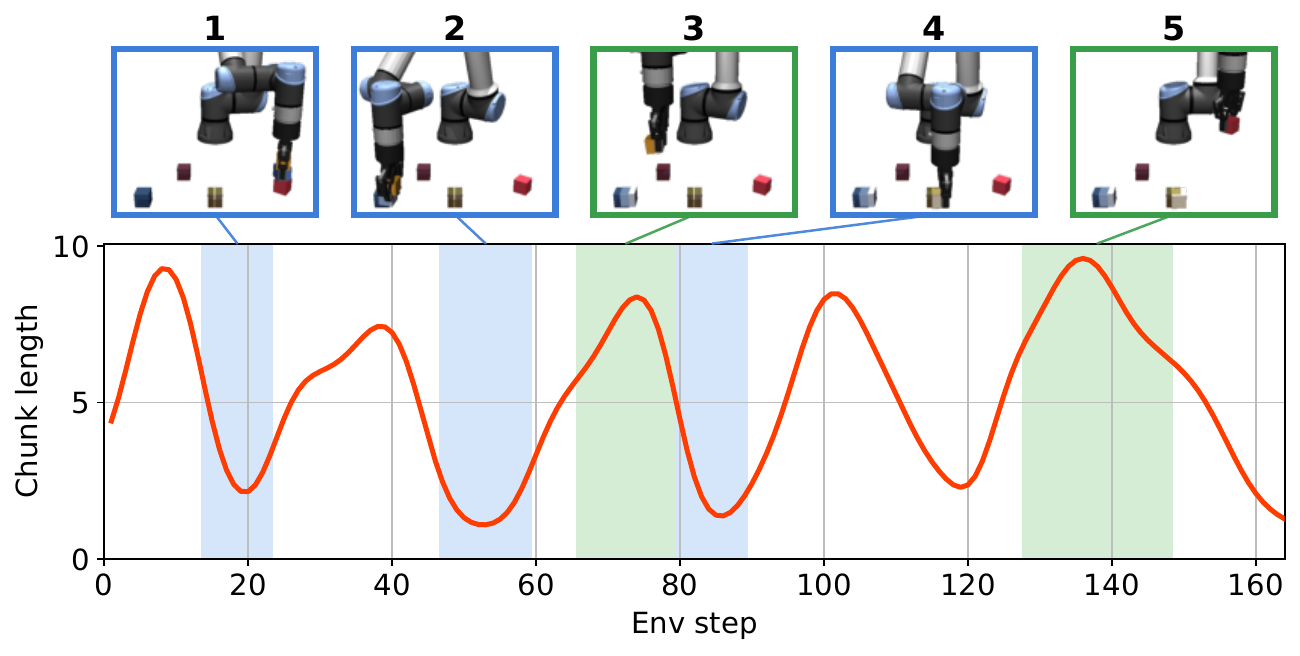}
    \vspace{-2ex}
    \caption{Chunk lengths selected by ACH over environment timesteps during a \texttt{cube-triple-task3} episode. Shaded regions correspond to the subfigures above. ACH selects shorter chunks (blue) when grasping or placing a cube and longer chunks (green) while moving between cubes.}
    \label{fig:manipulation}
\end{figure}
To examine whether the behavior in Figure \ref{fig:intro} extends beyond navigation, we analyze the chunk lengths selected by ACH during an episode of \texttt{cube-triple-task3}, where the agent must pick and place multiple cubes.
As shown in Figure \ref{fig:manipulation}, ACH repeatedly alternates between short and long chunks over the course of the episode.
When the gripper interacts with a cube, such as grasping or placing it, ACH reduces the chunk length to one or two and re-selects its actions from the latest observations during precise manipulation.
In contrast, when the gripper simply moves toward the next cube or carries a grasped cube to its target location, ACH selects long chunks, up to the maximum length, as these motions do not require precise control.
These results indicate that ACH adapts the chunk length to the precision required at each state not only in navigation but also in manipulation.

\subsection{Action Values across Chunk Lengths}
\label{appendix:motivation}
\begin{figure}[h!]
    \centering    \includegraphics[width=\linewidth]{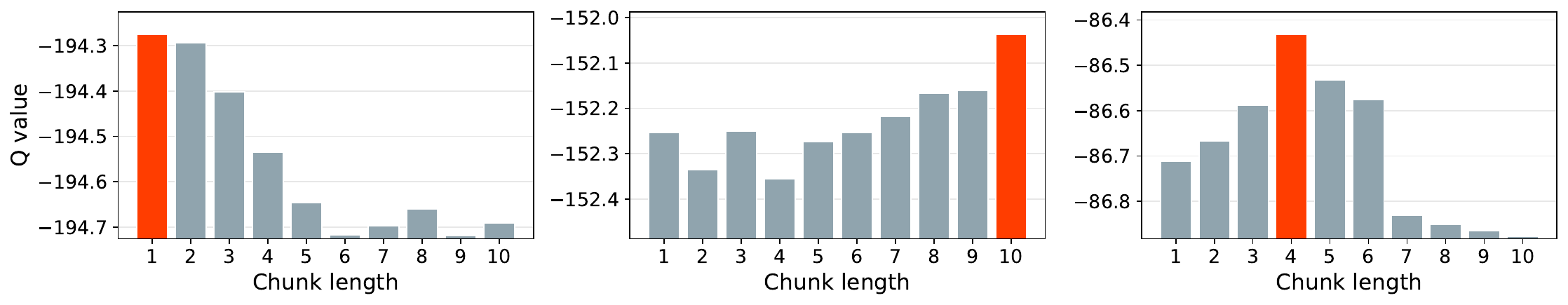}
    \vspace{-1ex}
    \caption{Action value as a function of chunk length (\texttt{antmaze-giant-navigate} with maximum chunk length set to 10): (left) fast-changing states, (middle) strongly correlated states over time, and (right) trade-off situation, which correspond to Subfigures 1, 3, and 5 in Figure \ref{fig:intro}, respectively. The minimal value difference stems from the reward design of the environment: a constant reward of -1 per step, complemented by a sparse reward (1 or 0).}
    \label{fig:motivation}
\end{figure}
Figure~\ref{fig:motivation} shows the action values of chunks with different lengths at three representative states in \texttt{antmaze-giant-navigate}, estimated by the Transformer-based value function trained with ACH (Section~\ref{subsec:transformer}).
In fast-changing states (left), where the uncertainty of future states dominates, the value is highest at a chunk length of one and decreases as the chunk becomes longer.
In states that are strongly correlated over time (middle), where policy stochasticity jittering dominates, the value is highest at the maximum chunk length.
When both factors are present (right), the value peaks at an intermediate chunk length.
These patterns confirm that the optimal chunk length varies across states and is determined by the trade-off discussed in Section~\ref{subsec:value_based_optimality}.

\subsection{Comparison with Fixed-Length Baselines}
\begin{table}[h!]
    \vspace{-2ex}
    \centering
    \footnotesize
    \vskip 0.1in
    \renewcommand{\arraystretch}{1.1}
    \caption{Comparison of ACH with fixed chunk length baselines on OGBench. H$n$ denotes the backbone trained and executed with a fixed chunk length of $n$, where H1 and H5 correspond to FQL and QC, respectively. All methods are evaluated after 1M steps of offline learning followed by 1M steps of online learning. We report the mean and 95\% confidence intervals across 10 seeds. Results for each environment are averaged over five tasks.}
    \vspace{2ex}
    \resizebox{\textwidth}{!}{%
    \begin{tabular}{l|*{5}{r@{\,$\rightarrow$\,}l}}
    \toprule
    Task
      & \multicolumn{2}{c}{H1 (FQL)}
      & \multicolumn{2}{c}{H3}
      & \multicolumn{2}{c}{H5 (QC)}
      & \multicolumn{2}{c}{H10}
      & \multicolumn{2}{c}{ACH} \\
    \midrule
    antmaze-giant-navigate & \res{10{\tiny{$\pm$5}}}{\textbf{74}{\tiny{$\pm$7}}} & \res{0{\tiny{$\pm$0}}}{\textbf{71}{\tiny{$\pm$7}}} & \res{0{\tiny{$\pm$0}}}{47{\tiny{$\pm$8}}} & \res{0{\tiny{$\pm$0}}}{20{\tiny{$\pm$6}}} & \res{1{\tiny{$\pm$1}}}{\textbf{72}{\tiny{$\pm$2}}} \\
    antsoccer-arena-navigate & \res{35{\tiny{$\pm$2}}}{70{\tiny{$\pm$5}}} & \res{50{\tiny{$\pm$3}}}{\textbf{82}{\tiny{$\pm$2}}} & \res{10{\tiny{$\pm$2}}}{67{\tiny{$\pm$2}}} & \res{0{\tiny{$\pm$0}}}{12{\tiny{$\pm$2}}} & \res{0{\tiny{$\pm$0}}}{\textbf{84}{\tiny{$\pm$1}}} \\
    cube-quadruple-play-10M & \res{0{\tiny{$\pm$0}}}{2{\tiny{$\pm$1}}} & \res{1{\tiny{$\pm$0}}}{8{\tiny{$\pm$5}}} & \res{1{\tiny{$\pm$1}}}{33{\tiny{$\pm$3}}} & \res{1{\tiny{$\pm$0}}}{41{\tiny{$\pm$3}}} & \res{0{\tiny{$\pm$0}}}{\textbf{52}{\tiny{$\pm$6}}} \\
    cube-triple-play & \res{2{\tiny{$\pm$1}}}{19{\tiny{$\pm$2}}} & \res{5{\tiny{$\pm$1}}}{35{\tiny{$\pm$3}}} & \res{4{\tiny{$\pm$2}}}{57{\tiny{$\pm$4}}} & \res{3{\tiny{$\pm$1}}}{\textbf{70}{\tiny{$\pm$4}}} & \res{1{\tiny{$\pm$1}}}{\textbf{67}{\tiny{$\pm$3}}} \\
    humanoidmaze-medium-navigate & \res{36{\tiny{$\pm$3}}}{56{\tiny{$\pm$3}}} & \res{53{\tiny{$\pm$3}}}{61{\tiny{$\pm$7}}} & \res{15{\tiny{$\pm$2}}}{32{\tiny{$\pm$4}}} & \res{0{\tiny{$\pm$0}}}{0{\tiny{$\pm$0}}} & \res{45{\tiny{$\pm$2}}}{\textbf{82}{\tiny{$\pm$1}}} \\
    puzzle-4x4-play & \res{14{\tiny{$\pm$2}}}{54{\tiny{$\pm$6}}} & \res{20{\tiny{$\pm$3}}}{50{\tiny{$\pm$8}}} & \res{11{\tiny{$\pm$2}}}{38{\tiny{$\pm$4}}} & \res{0{\tiny{$\pm$0}}}{20{\tiny{$\pm$5}}} & \res{13{\tiny{$\pm$3}}}{\textbf{90}{\tiny{$\pm$8}}} \\
    \bottomrule
    \end{tabular}%
    }
    \label{tab:chunk_length}
\end{table}
A natural question is whether the gains of ACH could be obtained simply by tuning a fixed chunk length for each environment.
To answer this question, we evaluate fixed-length baselines with four chunk lengths: 1, 3, 5, and 10.
As summarized in Table~\ref{tab:chunk_length}, their performance fluctuates significantly across environments depending on the chunk length.
Although certain fixed lengths perform comparably to ACH in some environments, they degrade substantially in others.
A fixed chunk length performs well only when that length is close to optimal for most states in an environment; otherwise, no single fixed length can cover the whole task.
For example, on \texttt{puzzle-4x4}, ACH achieves a success rate of 90, whereas the best fixed length reaches only 54, a gap that cannot be explained by a task-level choice of the chunk length.
{These results highlight the benefit of adapting the executed chunk length to each state, enabling ACH to achieve strong performance across environments.}

\begin{table}[h!]
        \vspace{-2ex}
    \centering
    \footnotesize
    \vskip 0.1in
    \renewcommand{\arraystretch}{1.1}
    \caption{Best fixed chunk length $H$ of Table~\ref{tab:chunk_length} with the MLP critic and with the Transformer critic of ACH. Transformer critic: mean and standard deviation over 4 seeds. Others: mean and 95\% confidence intervals over 10 seeds, taken from Table~\ref{tab:chunk_length}.}
    \vspace{2ex}
    \begin{tabular}{lc|*{3}{r@{\,$\rightarrow$\,}l}}
    \toprule
    Task & $H$
      & \multicolumn{2}{c}{MLP critic}
      & \multicolumn{2}{c}{Transformer critic}
      & \multicolumn{2}{c}{ACH} \\
    \midrule
    antmaze-giant-navigate & 1 & \res{10{\tiny{$\pm$5}}}{\textbf{74}{\tiny{$\pm$7}}} & \res{0{\tiny{$\pm$0}}}{\textbf{74}{\tiny{$\pm$6}}} & \res{1{\tiny{$\pm$1}}}{\textbf{72}{\tiny{$\pm$2}}} \\
    antsoccer-arena-navigate & 3 & \res{50{\tiny{$\pm$3}}}{\textbf{82}{\tiny{$\pm$2}}} & \res{1{\tiny{$\pm$0}}}{66{\tiny{$\pm$7}}} & \res{0{\tiny{$\pm$0}}}{\textbf{84}{\tiny{$\pm$1}}} \\
    cube-quadruple-play-10M & 10 & \res{1{\tiny{$\pm$0}}}{41{\tiny{$\pm$3}}} & \res{0{\tiny{$\pm$0}}}{43{\tiny{$\pm$9}}} & \res{0{\tiny{$\pm$0}}}{\textbf{52}{\tiny{$\pm$6}}} \\
    cube-triple-play & 10 & \res{3{\tiny{$\pm$1}}}{\textbf{70}{\tiny{$\pm$4}}} & \res{1{\tiny{$\pm$0}}}{\textbf{73}{\tiny{$\pm$8}}} & \res{1{\tiny{$\pm$1}}}{67{\tiny{$\pm$3}}} \\
    humanoidmaze-medium-navigate & 3 & \res{53{\tiny{$\pm$3}}}{61{\tiny{$\pm$7}}} & \res{37{\tiny{$\pm$6}}}{65{\tiny{$\pm$2}}} & \res{45{\tiny{$\pm$2}}}{\textbf{82}{\tiny{$\pm$1}}} \\
    puzzle-4x4-play & 1 & \res{14{\tiny{$\pm$2}}}{54{\tiny{$\pm$6}}} & \res{2{\tiny{$\pm$2}}}{55{\tiny{$\pm$10}}} & \res{13{\tiny{$\pm$3}}}{\textbf{90}{\tiny{$\pm$8}}} \\
    \bottomrule
    \end{tabular}
    \label{tab:transformer_fixed}
\end{table}
To isolate the benefit of adaptive chunk lengths, we also compare ACH with fixed-length baselines using the same Transformer critic architecture.
For each environment, we take the best fixed chunk length $H$ in Table~\ref{tab:chunk_length} and replace the MLP critic of the backbone with the Transformer critic of ACH, keeping its architecture and hyperparameters.
Table~\ref{tab:transformer_fixed} shows that replacing the MLP critic with the Transformer critic of ACH does not substantially improve the best fixed chunk length in any environment, and the gap to ACH remains on \texttt{cube-quadruple}, \texttt{humanoidmaze}, and \texttt{puzzle-4x4}.
These results indicate that the gains of ACH come from adapting the chunk length rather than from the Transformer critic.

\clearpage
\subsection{Computational Cost}
\label{appendix:compute}
\begin{figure}[h!]
    \centering
    \includegraphics[width=0.7\linewidth]{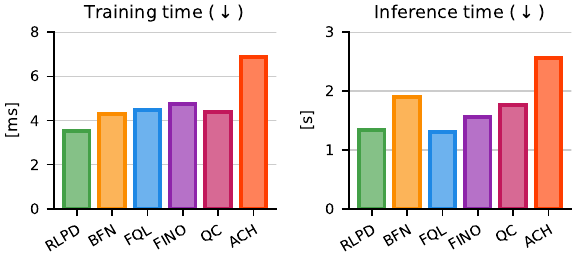}
    \vspace{-2ex}
    \caption{Comparison of computational costs between the proposed method and baselines on the \texttt{antmaze-giant-navigate} task. Training time represents the duration per step during online learning, and inference time denotes the time required for a single episode rollout. All measurements are conducted on a single NVIDIA RTX 3090 GPU.}
    \label{fig:computational_cost}
\end{figure}
In Figure \ref{fig:computational_cost}, we provide the computational costs of the proposed method and the baselines during both the training and inference phases.
Since ACH utilizes a relatively heavy transformer architecture and requires value estimation across all chunk lengths during inference, the computational time for ACH is higher than that of other baselines.
However, the increase is not prohibitive, as it remains 1.3 to 1.4 times the processing time of the baselines.
Considering the significant performance gains achieved by our method, this additional overhead is well within a justifiable limit.

\begin{figure}[h!]
    
    \centering
    \includegraphics[width=\linewidth]{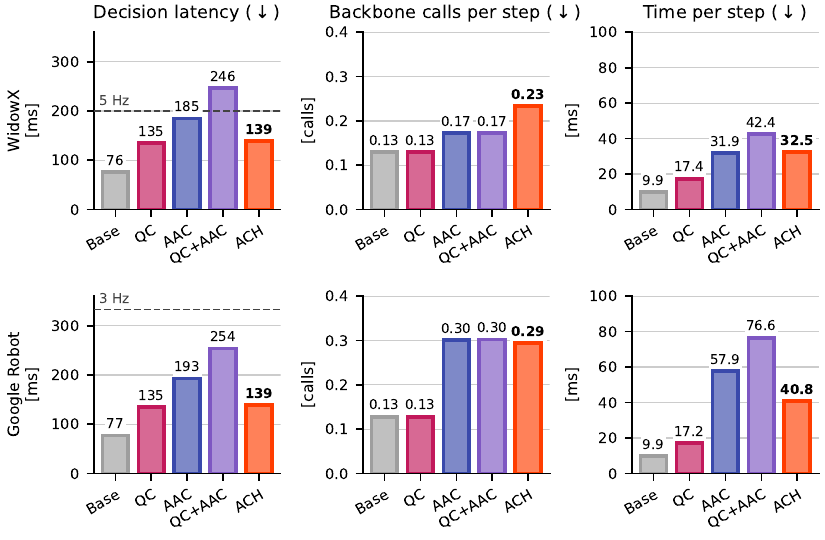}
    \vspace{-2ex}
    \caption{Inference cost in the VLA experiments on SimplerEnv, measured with a batch size of 1 on a single NVIDIA RTX 4000 Ada GPU. Dashed lines denote the shortest control period of each robot (5~Hz for WidowX and 3~Hz for the Google Robot).}
    \label{fig:vla_cost}
\end{figure}

We also measure the inference cost in the VLA experiments.
We report the decision latency, i.e., the time to produce the actions to execute from a single observation, the number of VLA backbone forward passes per environment step, and the policy computation time per environment step, excluding the simulator.
As shown in Figure~\ref{fig:vla_cost}, ACH has a decision latency similar to that of QC, with little additional overhead from evaluating all candidate chunk lengths with the Transformer-based value function in a single forward pass.
Most of the additional latency over the base policy comes from the one-step policy shared with QC.
ACH also has a lower decision latency than AAC and QC+AAC, which require 20 samples per decision.
Overall, ACH achieves higher success rates (Figure~\ref{fig:vla}) with decision latency comparable to QC and computation time per environment step comparable to or lower than that of AAC and QC+AAC.

\subsection{Evolution of Chunk Length}
\label{appendix:chunk_length}
\begin{figure}[h!]
    \centering
    \includegraphics[width=0.9\linewidth]{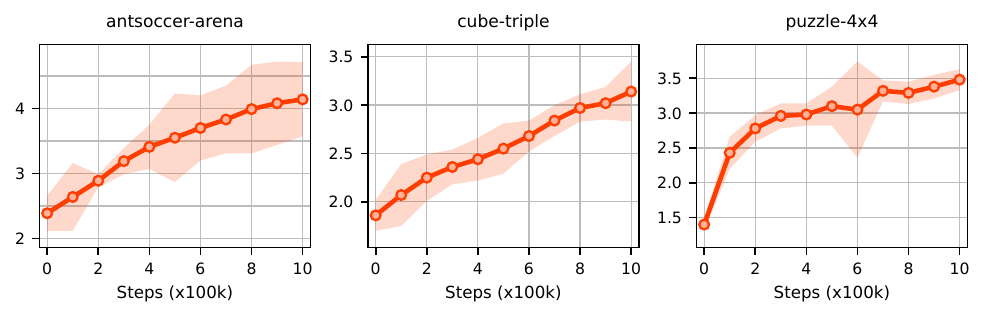}
    \vspace{-2ex}
    \caption{Evolution of average chunk length during the online learning phase on \texttt{antsoccer-arena}, \texttt{cube-triple}, and \texttt{puzzle-4x4}. The shaded region denotes one standard deviation across seeds. {The chunk lengths are those selected greedily at evaluation, averaged over the states visited by each checkpoint.}}
    \label{fig:chunk_length}
\end{figure}
In Off2On RL, the primary role of online learning is to enable the agent to refine its behavior through direct environment interaction.
While methods with a fixed chunk length are limited to adjusting individual actions, our approach facilitates both behavioral correction and chunk length optimization.
To investigate this, we analyze the evolution of the average chunk length during online learning.
As illustrated in Figure \ref{fig:chunk_length}, ACH exhibits a clear tendency to gradually increase the average chunk length as training progresses, and this trend holds consistently across both navigation (\texttt{antsoccer-arena}) and manipulation (\texttt{cube-triple} and \texttt{puzzle-4x4}) environments.
This trend is attributed to the stitching of trajectories within the value function{, together with a shift of the visited states, as we analyze below}.
As the policy optimizes based on this value function, it becomes capable of executing behaviors as single, extended sequences that previously required multiple short segments.
These results indicate that our approach goes beyond merely refining individual actions; by leveraging value-based trajectory stitching to progressively increase chunk lengths, it ultimately facilitates sample-efficient learning.

\begin{figure}[h!]
        \centering
    \includegraphics[width=0.85\linewidth]{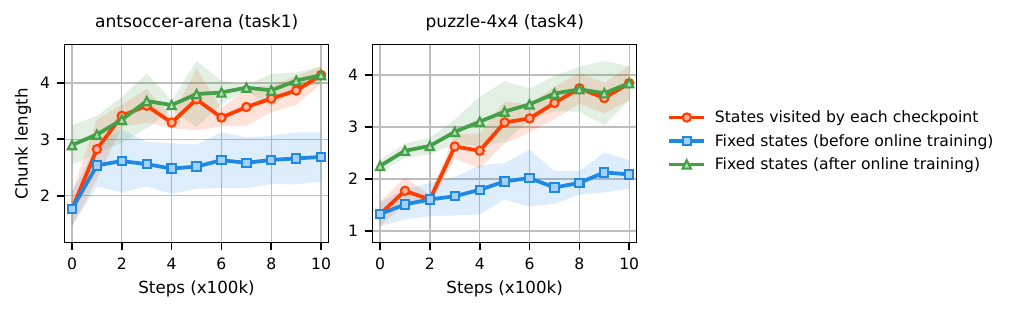}
    \vspace{-2ex}
    \caption{Chunk length selected greedily by each checkpoint during online learning on \texttt{antsoccer-arena} (task1) and \texttt{puzzle-4x4} (task4), measured on the states visited by each checkpoint and on two fixed sets of states, those visited before and after online training. The shaded region denotes one standard deviation across 4 seeds.}
    \label{fig:chunk_length_fixed}
\end{figure}
\par\noindent\textbf{Chunk Length at Fixed States}\quad\ignorespaces
Since the average in Figure~\ref{fig:chunk_length} is taken over the states visited by each checkpoint, its increase can reflect changes in the selected lengths at the same states as well as changes in the states visited by the policy.
To separate them, we fix two sets of states, those visited by the policy before online training and those visited after online training, and evaluate the selected chunk length of each checkpoint on both sets.
Figure~\ref{fig:chunk_length_fixed} shows that the selected chunk length increases during online learning even on fixed states, and differs substantially between the two state sets at the same checkpoint.
These results show that the selected chunk length depends on the states at which it is measured, but also increases through learning even when those states are held fixed.
Taken together, the results indicate that the trend in Figure~\ref{fig:chunk_length} reflects both changes in the visited states and learning of the policy and critic.

\subsection{Ablation Study on Robomimic}
\label{appendix:ablation_robomimic}
\begin{table}[h!]
    \centering
    \footnotesize
    \renewcommand{\arraystretch}{1.1}
    \caption{Ablation results on the \texttt{square} and \texttt{transport} tasks of robomimic, showing the performance immediately after 1M steps of offline learning and 1M steps of online learning. Results represent the mean $\pm$ 95\% confidence interval.}
    \vspace{1ex}
    \begin{tabular}{l|*{3}{r@{\,$\rightarrow$\,}l}}
    \toprule
    Task
      & \multicolumn{2}{c}{Uniform}
      & \multicolumn{2}{c}{Greedy}
      & \multicolumn{2}{c}{ACH} \\
    \midrule
    square & \res{44{\tiny{$\pm$4}}}{74{\tiny{$\pm$3}}} & \res{40{\tiny{$\pm$6}}}{80{\tiny{$\pm$5}}} & \res{42{\tiny{$\pm$5}}}{\textbf{94}{\tiny{$\pm$2}}} \\
    transport & \res{0{\tiny{$\pm$0}}}{11{\tiny{$\pm$1}}} & \res{0{\tiny{$\pm$0}}}{19{\tiny{$\pm$2}}} & \res{0{\tiny{$\pm$0}}}{\textbf{39}{\tiny{$\pm$4}}} \\
    \bottomrule
    \end{tabular}
    \label{tab:ablation_robomimic}
\end{table}
To examine whether the core components of ACH remain effective beyond OGBench, we additionally conduct the \texttt{Uniform} and \texttt{Greedy} ablations from Section \ref{subsec:analysis} on the two most challenging robomimic tasks, \texttt{square} and \texttt{transport}.
As shown in Table \ref{tab:ablation_robomimic}, the results are consistent with those in Figure \ref{fig:ablation}.
\texttt{Uniform} suffers from the misalignment between the training objective and the chunk length selection policy, and \texttt{Greedy} fails to explore sufficiently, which is particularly evident on the difficult \texttt{transport} task.
These results confirm that the core components of ACH are jointly necessary on robomimic as well.

\subsection{Results on Additional OGBench Environments}
\label{appendix:additional_envs}
\begin{table}[h!]
    
    \centering
    \footnotesize
    \renewcommand{\arraystretch}{1.1}
    \caption{Results of ACH and baselines on additional OGBench environments. All methods are evaluated after 1M steps of offline learning followed by 1M steps of online learning. We report the mean and 95\% confidence intervals across 10 seeds. Results for each environment are averaged over five tasks.}
    \vspace{1ex}
    \resizebox{\textwidth}{!}{%
    \begin{tabular}{l|*{8}{r@{\,$\rightarrow$\,}l}}
    \toprule
    Task
      & \multicolumn{2}{c}{RLPD}
      & \multicolumn{2}{c}{ReBRAC}
      & \multicolumn{2}{c}{BFN}
      & \multicolumn{2}{c}{FQL}
      & \multicolumn{2}{c}{FINO}
      & \multicolumn{2}{c}{QC-FQL}
      & \multicolumn{2}{c}{QC (best-of-$N$)}
      & \multicolumn{2}{c}{ACH} \\
    \midrule
    cube-double-play & \res{0{\tiny{$\pm$0}}}{34{\tiny{$\pm$10}}} & \res{12{\tiny{$\pm$5}}}{42{\tiny{$\pm$7}}} & \res{67{\tiny{$\pm$7}}}{74{\tiny{$\pm$8}}} & \res{28{\tiny{$\pm$6}}}{77{\tiny{$\pm$10}}} & \res{45{\tiny{$\pm$7}}}{84{\tiny{$\pm$8}}} & \res{43{\tiny{$\pm$7}}}{\textbf{100}{\tiny{$\pm$0}}} & \res{68{\tiny{$\pm$2}}}{\textbf{99}{\tiny{$\pm$1}}} & \res{15{\tiny{$\pm$6}}}{\textbf{99}{\tiny{$\pm$1}}} \\
    puzzle-3x3-play-sparse & \res{0{\tiny{$\pm$0}}}{\textbf{100}{\tiny{$\pm$0}}} & \res{65{\tiny{$\pm$3}}}{\textbf{100}{\tiny{$\pm$0}}} & \res{98{\tiny{$\pm$1}}}{\textbf{99}{\tiny{$\pm$1}}} & \res{100{\tiny{$\pm$0}}}{\textbf{100}{\tiny{$\pm$0}}} & \res{100{\tiny{$\pm$0}}}{\textbf{100}{\tiny{$\pm$0}}} & \res{64{\tiny{$\pm$13}}}{\textbf{100}{\tiny{$\pm$0}}} & \res{100{\tiny{$\pm$0}}}{\textbf{100}{\tiny{$\pm$0}}} & \res{91{\tiny{$\pm$3}}}{\textbf{100}{\tiny{$\pm$0}}} \\
    scene-play-sparse & \res{0{\tiny{$\pm$0}}}{93{\tiny{$\pm$3}}} & \res{20{\tiny{$\pm$4}}}{\textbf{99}{\tiny{$\pm$1}}} & \res{82{\tiny{$\pm$8}}}{\textbf{99}{\tiny{$\pm$1}}} & \res{54{\tiny{$\pm$7}}}{\textbf{96}{\tiny{$\pm$2}}} & \res{96{\tiny{$\pm$2}}}{\textbf{98}{\tiny{$\pm$1}}} & \res{85{\tiny{$\pm$6}}}{\textbf{99}{\tiny{$\pm$1}}} & \res{80{\tiny{$\pm$4}}}{\textbf{98}{\tiny{$\pm$1}}} & \res{96{\tiny{$\pm$3}}}{\textbf{99}{\tiny{$\pm$1}}} \\
    \bottomrule
    \end{tabular}%
    }
    \label{tab:additional_envs}
\end{table}
To further compare ACH with QC on the tasks used in its original evaluation \citep{qc}, we additionally evaluate all methods on \texttt{cube-double-play}, \texttt{puzzle-3x3-play-sparse}, and \texttt{scene-play-sparse}.
As shown in Table~\ref{tab:additional_envs}, QC-FQL, QC (best-of-$N$), and ACH all reach near-perfect success rates after online learning on these additional environments.
ACH performs on par with the best method in all three environments, showing that adaptive chunking maintains strong performance in environments where a fixed chunk length is already sufficient.

\subsection{Online RL without Offline Pre-training}
\label{appendix:online}
\begin{table}[h!]
    
    \centering
    \footnotesize
    \renewcommand{\arraystretch}{1.1}
    \caption{Results after 1M online steps without offline pre-training (ACH (Online)) and with it (ACH (Off2On)). ACH (Online): mean and standard deviation over 4 seeds. Others: mean and 95\% confidence intervals over 10 seeds, taken from Table~\ref{tab:main}.}
    \vspace{1ex}
    \begin{tabular}{l|ccc}
    \toprule
    Task & RLPD & ACH (Online) & ACH (Off2On) \\
    \midrule
    antmaze-giant-navigate       & \textbf{69}{\tiny{$\pm$9}}  & \textbf{71}{\tiny{$\pm$8}}  & \textbf{72}{\tiny{$\pm$2}} \\
    antsoccer-arena-navigate     & 64{\tiny{$\pm$4}}           & 62{\tiny{$\pm$8}}           & \textbf{84}{\tiny{$\pm$1}} \\
    humanoidmaze-medium-navigate & 29{\tiny{$\pm$9}}           & 61{\tiny{$\pm$13}}          & \textbf{82}{\tiny{$\pm$1}} \\
    puzzle-4x4-play              & 64{\tiny{$\pm$12}}          & 70{\tiny{$\pm$20}}          & \textbf{90}{\tiny{$\pm$8}} \\
    \bottomrule
    \end{tabular}
    \label{tab:online}
\end{table}
To examine the role of offline pre-training, we train ACH without the offline phase, following the protocol of RLPD \citep{rlpd}, in which the offline dataset is placed in the replay buffer and learning proceeds purely online.
As shown in Table~\ref{tab:online}, ACH (Online) performs comparably to or better than RLPD, and substantially outperforms it on \texttt{humanoidmaze-medium}, which shows that the benefit of adaptive action chunking carries over to the purely online setting.
On \texttt{antmaze-giant}, where RLPD already performs well with online learning alone, ACH (Online) is comparable to ACH (Off2On).
On the other environments, however, ACH (Off2On) achieves clearly higher performance.
We attribute this to the fact that the value-based chunk length selection requires reasonably accurate value estimates, which offline pre-training provides from the start of online learning, even when the success rate of the pre-trained policy itself is low.
This supports the Off2On setting as the natural choice for ACH.

\subsection{Comparison with Best-of-N QC}
\label{appendix:best_of_n}
\begin{table}[h!]
    
    \centering
    \footnotesize
    \renewcommand{\arraystretch}{1.1}
    \caption{Comparison with the best-of-$N$ variant of QC ($N=32$). QC (best-of-$N$): mean and standard deviation over 4 seeds. Others: mean and 95\% confidence intervals over 10 seeds, taken from Table~\ref{tab:main}.}
    \vspace{1ex}
    \begin{tabular}{l|*{3}{r@{\,$\rightarrow$\,}l}}
    \toprule
    Task
      & \multicolumn{2}{c}{QC-FQL}
      & \multicolumn{2}{c}{QC (best-of-$N$)}
      & \multicolumn{2}{c}{ACH} \\
    \midrule
    antmaze-giant-navigate & \res{0{\tiny{$\pm$0}}}{47{\tiny{$\pm$8}}} & \res{0{\tiny{$\pm$0}}}{\textbf{72}{\tiny{$\pm$11}}} & \res{1{\tiny{$\pm$1}}}{\textbf{72}{\tiny{$\pm$2}}} \\
    antsoccer-arena-navigate & \res{10{\tiny{$\pm$2}}}{67{\tiny{$\pm$2}}} & \res{30{\tiny{$\pm$10}}}{\textbf{80}{\tiny{$\pm$5}}} & \res{0{\tiny{$\pm$0}}}{\textbf{84}{\tiny{$\pm$1}}} \\
    cube-quadruple-play-10M & \res{1{\tiny{$\pm$1}}}{33{\tiny{$\pm$3}}} & \res{8{\tiny{$\pm$4}}}{42{\tiny{$\pm$9}}} & \res{0{\tiny{$\pm$0}}}{\textbf{52}{\tiny{$\pm$6}}} \\
    cube-triple-play & \res{4{\tiny{$\pm$2}}}{57{\tiny{$\pm$4}}} & \res{6{\tiny{$\pm$2}}}{61{\tiny{$\pm$5}}} & \res{1{\tiny{$\pm$1}}}{\textbf{67}{\tiny{$\pm$3}}} \\
    humanoidmaze-medium-navigate & \res{15{\tiny{$\pm$2}}}{32{\tiny{$\pm$4}}} & \res{39{\tiny{$\pm$3}}}{71{\tiny{$\pm$7}}} & \res{45{\tiny{$\pm$2}}}{\textbf{82}{\tiny{$\pm$1}}} \\
    puzzle-4x4-play & \res{11{\tiny{$\pm$2}}}{38{\tiny{$\pm$4}}} & \res{18{\tiny{$\pm$3}}}{52{\tiny{$\pm$9}}} & \res{13{\tiny{$\pm$3}}}{\textbf{90}{\tiny{$\pm$8}}} \\
    \bottomrule
    \end{tabular}
    \label{tab:best_of_n}
\end{table}
Our QC baseline uses the variant of \citet{qc} that distills the behavior policy into a one-step policy (QC-FQL), on which ACH is also built.
\citet{qc} also propose a best-of-$N$ variant, which samples $N$ action chunks from the behavior policy at each decision and executes the one with the highest estimated value.
We compare ACH with this variant using $N=32$ on all six environments of Table~\ref{tab:main}.
As shown in Table~\ref{tab:best_of_n}, the best-of-$N$ variant outperforms QC-FQL in all environments.
ACH performs comparably to the best-of-$N$ variant on \texttt{antmaze-giant} and \texttt{antsoccer-arena}, and outperforms it on the remaining four environments, with the largest gap on \texttt{puzzle-4x4} (90 vs.\ 52).
Note that ACH achieves this with a single action chunk per decision, whereas the best-of-$N$ variant samples and evaluates 32 action chunks at every decision.
These results show that the gains of ACH are not explained by a stronger policy extraction at a fixed chunk length.

\subsection{Comparison with a SEAR-style Variant}
\label{appendix:sear}
\begin{table}[h!]
        \vspace{-2ex}
    \centering
    \footnotesize
    \vskip 0.1in
    \renewcommand{\arraystretch}{1.1}
    \caption{Comparison with a SEAR-style variant evaluated with a fixed receding horizon $H$. SEAR-style: mean and standard deviation over 4 seeds. ACH: mean and 95\% confidence intervals over 10 seeds, taken from Table~\ref{tab:main}. The maximum chunk length on \texttt{humanoidmaze} is 5 (Appendix~\ref{appendix:ach}).}
    \vspace{2ex}
    \resizebox{\textwidth}{!}{%
    \begin{tabular}{l|*{5}{r@{\,$\rightarrow$\,}l}}
    \toprule
    Task
      & \multicolumn{2}{c}{H1}
      & \multicolumn{2}{c}{H3}
      & \multicolumn{2}{c}{H5}
      & \multicolumn{2}{c}{H10}
      & \multicolumn{2}{c}{ACH} \\
    \midrule
    cube-quadruple-play-10M & \res{0{\tiny{$\pm$0}}}{27{\tiny{$\pm$14}}} & \res{0{\tiny{$\pm$0}}}{28{\tiny{$\pm$14}}} & \res{0{\tiny{$\pm$0}}}{38{\tiny{$\pm$13}}} & \res{1{\tiny{$\pm$1}}}{45{\tiny{$\pm$13}}} & \res{0{\tiny{$\pm$0}}}{\textbf{52}{\tiny{$\pm$6}}} \\
    humanoidmaze-medium-navigate & \res{43{\tiny{$\pm$12}}}{60{\tiny{$\pm$13}}} & \res{27{\tiny{$\pm$8}}}{49{\tiny{$\pm$11}}} & \res{12{\tiny{$\pm$3}}}{30{\tiny{$\pm$4}}} & \multicolumn{2}{c}{--} & \res{45{\tiny{$\pm$2}}}{\textbf{82}{\tiny{$\pm$1}}} \\
    puzzle-4x4-play & \res{7{\tiny{$\pm$5}}}{50{\tiny{$\pm$16}}} & \res{22{\tiny{$\pm$6}}}{43{\tiny{$\pm$11}}} & \res{19{\tiny{$\pm$7}}}{36{\tiny{$\pm$12}}} & \res{15{\tiny{$\pm$4}}}{35{\tiny{$\pm$10}}} & \res{13{\tiny{$\pm$3}}}{\textbf{90}{\tiny{$\pm$8}}} \\
    \bottomrule
    \end{tabular}%
    }
    \label{tab:sear}
\end{table}
SEAR \citep{sear} also uses a causal Transformer critic that evaluates all prefixes of an action chunk. During training, it executes a prefix of uniformly random length, trains the critic on all prefixes with equal weights, and updates the actor with the value of the full chunk. At evaluation, it executes a fixed number of actions of each chunk before replanning, i.e., a fixed receding horizon.
Since the original SEAR builds on a soft actor-critic policy and learns purely online, we transfer its treatment of the chunk length to the backbone of ACH, keeping the policy, the critic architecture, and the Off2On protocol identical to ACH.
We train this SEAR-style variant once per environment and evaluate it with each receding horizon $H$.
As shown in Table~\ref{tab:sear}, the best receding horizon differs across environments, namely $H=10$ on \texttt{cube-quadruple} and $H=1$ on \texttt{humanoidmaze} and \texttt{puzzle-4x4}.
Even with the best horizon chosen for each environment after training, the variant remains below ACH.
These results indicate that a prefix critic trained with random prefix execution does not by itself yield the gains of ACH, which come from selecting the chunk length based on the prefix values during both training and evaluation.

\subsection{Reliability of the Critic's Ranking}
\label{appendix:mc_ranking}
\begin{table}[h!]
        \vspace{-2ex}
    \centering
    \footnotesize
    \vskip 0.1in
    \renewcommand{\arraystretch}{1.1}
    \caption{Pairwise accuracy (\%) of the ranking of chunk lengths against Monte-Carlo returns, computed over pairs of lengths whose returns differ significantly. Critic: mean and 95\% confidence intervals over 4 seeds. Always longer and always shorter rank the longer and the shorter length higher, respectively. A random ranking achieves 50\%.}
    \vspace{2ex}
    \begin{tabular}{l|ccc}
    \toprule
    Task & Critic & Always longer & Always shorter \\
    \midrule
    cube-triple-play (task3)            & \textbf{87}{\tiny{$\pm$12}}\% & 65\% & 35\% \\
    humanoidmaze-medium-navigate (task5) & \textbf{77}{\tiny{$\pm$13}}\% & 44\% & 56\% \\
    puzzle-4x4-play (task4)             & \textbf{91}{\tiny{$\pm$8}}\%  & 74\% & 26\% \\
    \bottomrule
    \end{tabular}
    \label{tab:mc_ranking}
\end{table}
To examine whether the critic ranks the chunk lengths correctly, we compare its values with Monte-Carlo (MC) returns.
For each task, we take the policy and the critic after online learning and sample states visited by the policy.
At each state, we generate a single full-length chunk and estimate the MC return of each prefix by executing it from that same state and then following the continuation policy, with chunk lengths sampled from $\pi_L$.
For each pair of prefix lengths, we compare the ordering of their estimated MC returns with the ordering of their critic values, keeping the state and candidate chunk fixed.
We consider only pairs for which the 95\% confidence interval of the paired MC return difference excludes zero.
For these pairs, we report how often the critic assigns a higher value to the prefix with the higher MC return (pairwise accuracy).
The always-longer and always-shorter baselines are evaluated on the same pairs by ranking the longer and shorter prefix higher, respectively.
Table~\ref{tab:mc_ranking} shows that the critic ranks the prefixes more accurately than a random ranking.
It also outperforms the fixed preferences for longer or shorter chunks in all tasks, including \texttt{humanoidmaze}, where neither preference dominates, and \texttt{puzzle-4x4}, where longer chunks are preferred in most pairs.
This shows that the critic distinguishes the states in which shorter chunks are better from those in which longer chunks are better, rather than following a single preference.

\subsection{Maximum Chunk Length on Humanoidmaze}
\label{appendix:max_chunk_length}
\begin{table}[h!]
        \centering
    \footnotesize
    \renewcommand{\arraystretch}{1.1}
    \caption{Effect of the chunk horizon on \texttt{humanoidmaze}: mean and 95\% CIs over 10 seeds per task, averaged over five tasks, after 1M offline steps and a further 1M online steps. QC-FQL executes the full chunk, whereas ACH selects a prefix up to the specified horizon. QC-FQL results are from Table~\ref{tab:chunk_length}; ACH results for horizon 5 are from Table~\ref{tab:main}.}
    \begin{tabular}{c|*{2}{r@{\,$\rightarrow$\,}l}}
    \toprule
    Chunk horizon & \multicolumn{2}{c}{QC-FQL} & \multicolumn{2}{c}{ACH} \\
    \midrule
    5 & \res{15{\tiny{$\pm$2}}}{32{\tiny{$\pm$4}}} & \res{45{\tiny{$\pm$2}}}{\textbf{82}{\tiny{$\pm$1}}} \\
    10 & \res{0{\tiny{$\pm$0}}}{0{\tiny{$\pm$0}}} & \res{2{\tiny{$\pm$1}}}{54{\tiny{$\pm$10}}} \\
    \bottomrule
    \end{tabular}
    \label{tab:max_chunk_length}
\end{table}
Table~\ref{tab:max_chunk_length} compares ACH with QC-FQL at the same chunk horizons.
ACH outperforms QC-FQL at both horizons, so its advantage over this baseline is not solely due to the choice of a shorter maximum chunk length.
At a horizon of 10, QC-FQL fails to learn a successful policy, whereas ACH achieves substantial success after online learning.
Both methods nevertheless perform better with a horizon of 5, supporting our choice for this environment.

\subsection{Sensitivity to Hyperparameters}
\label{appendix:sensitivity}
\begin{table}[h!]
        \centering
    \footnotesize
    \renewcommand{\arraystretch}{1.1}
    \caption{Sensitivity of ACH to the inverse temperature $\beta$ and the BC coefficient $\alpha$ on \texttt{antsoccer-arena} and \texttt{puzzle-4x4}: success rates averaged over five tasks, after 1M offline steps and a further 1M online steps. Default settings: mean and 95\% confidence intervals over 10 seeds, taken from Table~\ref{tab:main}. Others: mean and standard deviation over 4 seeds. The bottom table reports the average chunk length before and after online learning, where the default values are taken from Figure~\ref{fig:chunk_length}.}
    \begin{tabular}{l|*{4}{r@{\,$\rightarrow$\,}l}}
    \toprule
    Environment & \multicolumn{2}{c}{$\beta=0.1$} & \multicolumn{2}{c}{$\beta=1$ (default)} & \multicolumn{2}{c}{$\beta=10$} & \multicolumn{2}{c}{$\beta=100$} \\
    \midrule
    antsoccer-arena-navigate & \res{1{\tiny{$\pm$1}}}{\textbf{83}{\tiny{$\pm$3}}} & \res{0{\tiny{$\pm$0}}}{\textbf{84}{\tiny{$\pm$1}}} & \res{0{\tiny{$\pm$0}}}{78{\tiny{$\pm$3}}} & \res{0{\tiny{$\pm$0}}}{67{\tiny{$\pm$5}}} \\
    puzzle-4x4-play & \res{10{\tiny{$\pm$2}}}{80{\tiny{$\pm$5}}} & \res{13{\tiny{$\pm$3}}}{\textbf{90}{\tiny{$\pm$8}}} & \res{4{\tiny{$\pm$3}}}{\textbf{90}{\tiny{$\pm$5}}} & \res{5{\tiny{$\pm$3}}}{85{\tiny{$\pm$9}}} \\
    \bottomrule
    \end{tabular}

    \vspace{2ex}
    \begin{tabular}{l|*{3}{r@{\,$\rightarrow$\,}l}}
    \toprule
    Environment & \multicolumn{2}{c}{$\alpha/3$} & \multicolumn{2}{c}{$\alpha$ (default)} & \multicolumn{2}{c}{$3\alpha$} \\
    \midrule
    antsoccer-arena-navigate & \res{0{\tiny{$\pm$0}}}{36{\tiny{$\pm$13}}} & \res{0{\tiny{$\pm$0}}}{\textbf{84}{\tiny{$\pm$1}}} & \res{8{\tiny{$\pm$2}}}{68{\tiny{$\pm$3}}} \\
    puzzle-4x4-play & \res{16{\tiny{$\pm$5}}}{85{\tiny{$\pm$10}}} & \res{13{\tiny{$\pm$3}}}{\textbf{90}{\tiny{$\pm$8}}} & \res{15{\tiny{$\pm$4}}}{60{\tiny{$\pm$0}}} \\
    \bottomrule
    \end{tabular}

    \vspace{2ex}
    \begin{tabular}{l|*{2}{r@{\,$\rightarrow$\,}l}}
    \toprule
    Average chunk length & \multicolumn{2}{c}{$\alpha/3$} & \multicolumn{2}{c}{$\alpha$ (default)} \\
    \midrule
    puzzle-4x4-play & \res{1.42{\tiny{$\pm$0.14}}}{3.28{\tiny{$\pm$0.23}}} & \res{1.40{\tiny{$\pm$0.09}}}{3.48{\tiny{$\pm$0.15}}} \\
    \bottomrule
    \end{tabular}
    \label{tab:sensitivity}
\end{table}
Table~\ref{tab:sensitivity} shows the performance of ACH under different values of the inverse temperature $\beta$ in Eq.~(\ref{eq:sampling}) and the BC coefficient $\alpha$.
For $\beta$, we evaluate $\{0.1, 1, 10, 100\}$, where $\beta=1$ is the default.
The default $\beta=1$ performs best on both environments, and the performance changes moderately when $\beta$ is changed by a factor of 10.
However, with a large $\beta$, the length selection during training becomes close to greedy and explores fewer lengths, which lowers the performance on \texttt{antsoccer-arena}.
With a small $\beta$, the length selection becomes close to uniform and relies less on the values of the critic, which lowers the performance on \texttt{puzzle-4x4}.
For $\alpha$, we evaluate about one third and three times the default value in Table~\ref{tab:hyperparameter_alpha}.
The performance depends more on $\alpha$ than on $\beta$, as in the behavior-regularized methods that ACH builds on, where $\alpha$ is tuned for each environment \citep{fql,qc}.
On \texttt{puzzle-4x4}, where the performance with $\alpha/3$ is close to that of the default, the average chunk length before and after online learning is also close to that of the default, so the selected lengths are not sensitive to $\alpha$ in this range.

\newpage
\section{Complete Results}
\label{appendix:results}
\subsection{Main Experiments}
\begin{figure}[h!]
    \centering
    \includegraphics[width=\linewidth]{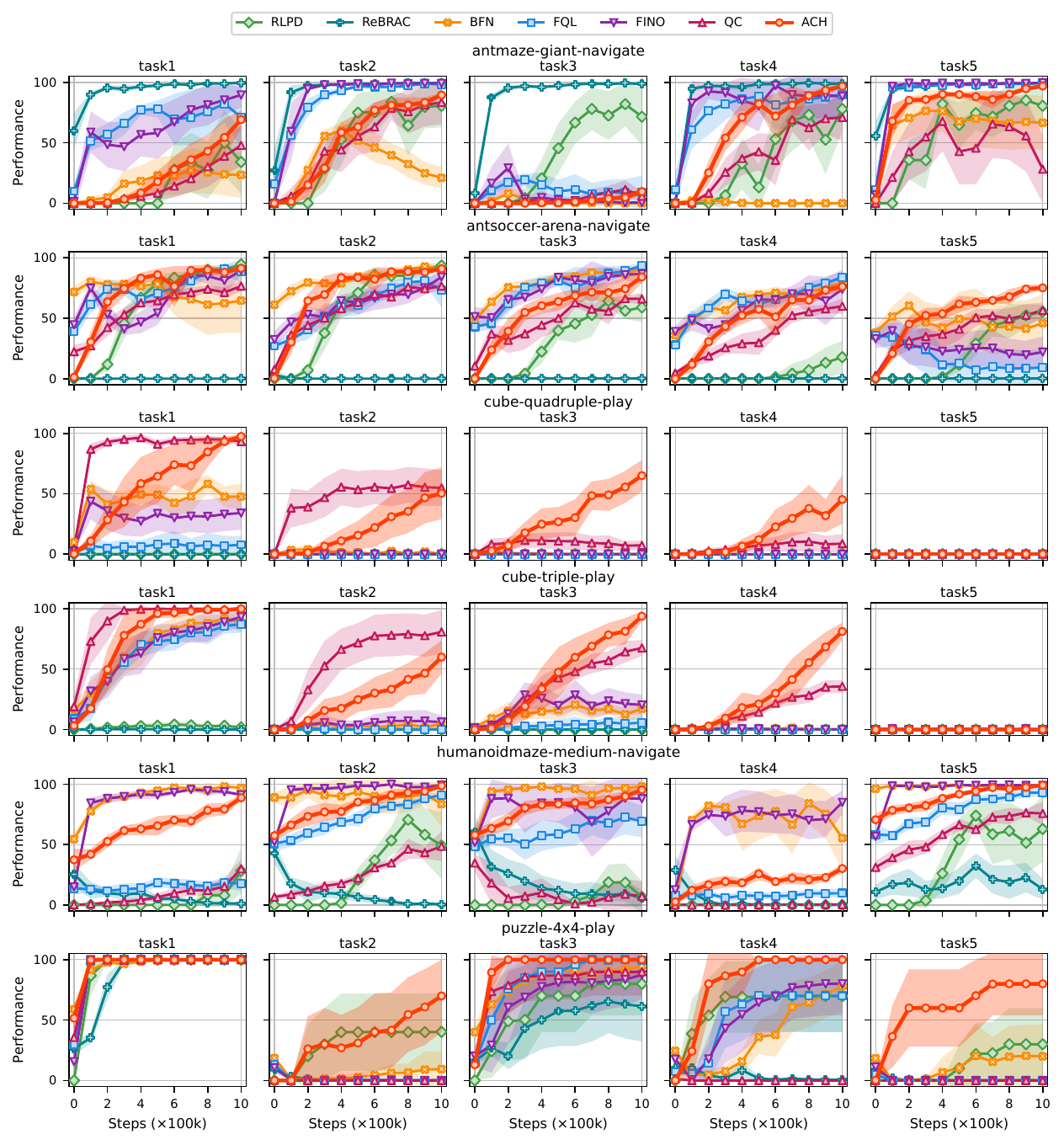}
    \caption{Learning curves during online learning for all tasks in OGBench. Solid lines represent the mean across 10 seeds, and shaded regions denote the 95\% confidence interval.}
    \label{fig:ogbench_full}
\end{figure}

We present the per-task results corresponding to Table \ref{tab:main} in Table \ref{tab:full_result_main}, and the corresponding learning curves in Figure \ref{fig:ogbench_full}.
For the ablation study shown in Figure \ref{fig:ablation}, we report the results across all tasks in Table \ref{tab:full_result_ablation}.

\subsection{VLA Experiments}
\label{appendix:vla_results}
We present the per-task results corresponding to Figure \ref{fig:vla} in Table \ref{tab:full_result_vla}.

\newcommand{\vres}[2]{#1{\tiny{$\pm$#2}}}
\begin{table}[h!]
    \centering
    \footnotesize
    \renewcommand{\arraystretch}{1.1}
    \caption{Complete results for Figure \ref{fig:vla}, showing the success rates (\%) of each method on SimplerEnv. Results represent the mean $\pm$ 95\% confidence interval calculated across 4 seeds, with 200 evaluation episodes per task.}
    \vspace{1ex}
    \begin{tabular}{ll|ccccc}
    \toprule
    Robot & Task & Base & QC & AAC & QC+AAC & ACH \\
    \midrule
    \multirow{8}{*}{WidowX}
      & \texttt{spoon\_on\_towel} & 81.5 & \vres{77.9}{2.0} & \vres{80.0}{4.7} & \vres{79.6}{3.6} & \vres{\textbf{91.1}}{4.7} \\
      & \texttt{carrot\_on\_plate} & 68.5 & \vres{68.8}{12.5} & \vres{62.6}{4.9} & \vres{67.0}{17.0} & \vres{\textbf{86.1}}{10.2} \\
      & \texttt{put\_eggplant\_in\_basket} & 25.0 & \vres{26.1}{11.1} & \vres{52.0}{3.2} & \vres{49.2}{13.8} & \vres{\textbf{67.6}}{6.5} \\
      & \texttt{stack\_cube} & 41.0 & \vres{39.6}{2.9} & \vres{34.9}{3.3} & \vres{35.0}{5.8} & \vres{\textbf{46.9}}{6.6} \\
      & \texttt{put\_eggplant\_in\_sink} & 2.0 & \vres{1.0}{0.6} & \vres{0.8}{0.5} & \vres{1.0}{1.3} & \vres{\textbf{5.3}}{2.1} \\
      & \texttt{close\_drawer} & \textbf{98.0} & \vres{\textbf{100.0}}{0.0} & \vres{\textbf{98.0}}{0.9} & \vres{\textbf{99.6}}{0.8} & \vres{\textbf{100.0}}{0.0} \\
      & \texttt{open\_drawer} & \textbf{100.0} & \vres{\textbf{99.9}}{0.4} & \vres{\textbf{99.6}}{0.4} & \vres{\textbf{100.0}}{0.0} & \vres{\textbf{100.0}}{0.0} \\
      \cmidrule{2-7}
      & Average & 59.4 & \vres{59.0}{2.4} & \vres{61.1}{1.2} & \vres{61.6}{3.3} & \vres{\textbf{71.0}}{2.1} \\
    \midrule
    \multirow{7}{*}{\shortstack{Google\\Robot}}
      & \texttt{pick\_coke\_can} & 64.0 & \vres{59.7}{5.3} & \vres{90.1}{1.5} & \vres{92.8}{2.9} & \vres{\textbf{98.2}}{3.8} \\
      & \texttt{pick\_object} & 38.5 & \vres{42.4}{4.6} & \vres{67.9}{4.7} & \vres{72.8}{3.5} & \vres{\textbf{87.0}}{4.0} \\
      & \texttt{move\_near} & 54.0 & \vres{54.2}{2.9} & \vres{80.7}{0.8} & \vres{86.1}{3.9} & \vres{\textbf{94.0}}{4.8} \\
      & \texttt{open\_drawer} & 35.0 & \vres{39.5}{7.0} & \vres{42.5}{8.4} & \vres{43.7}{5.6} & \vres{\textbf{55.0}}{7.0} \\
      & \texttt{close\_drawer} & 69.5 & \vres{\textbf{78.9}}{2.3} & \vres{65.6}{6.0} & \vres{68.8}{3.1} & \vres{\textbf{80.5}}{1.7} \\
      & \texttt{place\_in\_closed\_drawer} & 6.0 & \vres{7.9}{3.6} & \vres{8.7}{1.4} & \vres{7.9}{2.9} & \vres{\textbf{16.6}}{2.1} \\
      \cmidrule{2-7}
      & Average & 44.5 & \vres{47.1}{1.9} & \vres{59.2}{1.9} & \vres{62.0}{1.5} & \vres{\textbf{71.9}}{1.7} \\
    \bottomrule
    \end{tabular}
    \label{tab:full_result_vla}
\end{table}

\subsection{Behavior Analysis}
\label{appendix:analysis}
\begin{figure}[h!]
  \centering
  \includegraphics[width=\linewidth]{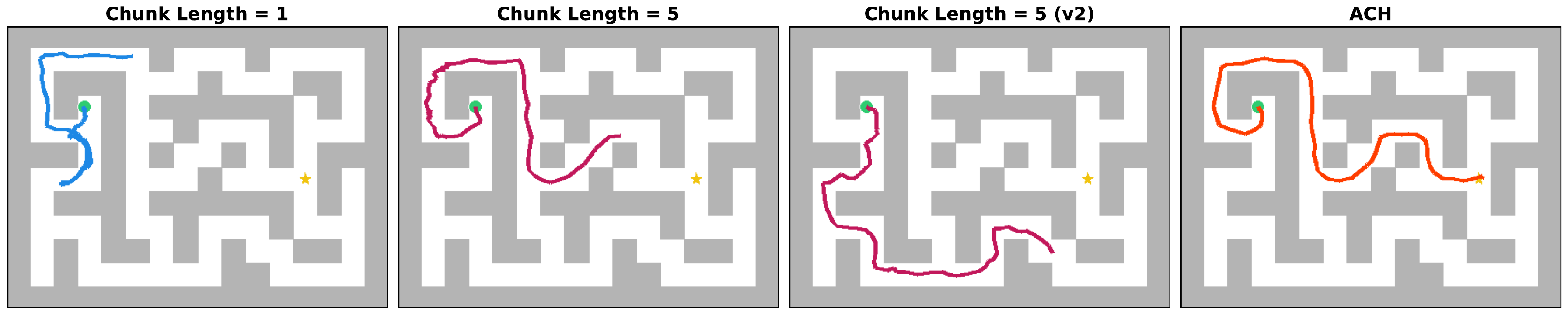}
  \vspace{-3ex}
  \caption{Behavioral comparison of ACH and fixed-length baselines in the \texttt{antmaze-giant} environment. Starting from a shared initial state, all algorithms execute the same number of total action steps. Agent trajectories are constructed by connecting sequential positions at each timestep. The initial state and goal state are marked by a green circle and a yellow star, respectively.}
  \vspace{-3ex}
  \label{fig:analysis_appendix}
\end{figure}
In Figure \ref{fig:analysis}, we validate the efficacy of the proposed method by analyzing the behaviors required to solve tasks in a practical environment, comparing our approach with fixed-length baselines.
In this section, Figure \ref{fig:analysis_appendix} presents the full layout of the environment and an additional behavioral pattern observed when the chunk length is fixed at 5 (v2).
The trajectory for this baseline reveals a significant detour instead of following the shortest path to the goal, which eventually leads to a failure to reach the goal within the time limit.
This outcome stems from the rigid nature of the fixed-length constraint. If the agent deviates from the optimal path during the early stages, the lack of flexibility prevents it from recovering and forces it to follow a suboptimal route.
These findings highlight the inherent limitation of conventional methods that rely on a static chunking strategy.

\newpage
\begin{table}[h!]
    \centering
    \footnotesize
    \vskip 0.1in
    \renewcommand{\arraystretch}{1.1}
    \caption{Complete results for Table \ref{tab:main} and Figure \ref{fig:robomimics}, showing the performance immediately after 1M steps of offline learning and 1M steps of online learning. Results represent the mean $\pm$ 95\% confidence interval calculated across 10 seeds.}
    \vspace{2ex}
    \resizebox{\textwidth}{!}{%
    \begin{tabular}{l|*{7}{r@{\,$\rightarrow$\,}l}}
    \toprule
    Environment
      & \multicolumn{2}{c}{RLPD}
      & \multicolumn{2}{c}{ReBRAC}
      & \multicolumn{2}{c}{BFN}
      & \multicolumn{2}{c}{FQL}
      & \multicolumn{2}{c}{FINO}
      & \multicolumn{2}{c}{QC}
      & \multicolumn{2}{c}{ACH} \\
    \midrule
    OGBench antmaze-giant-navigate-task1 & \res{0{\tiny{$\pm$0}}}{34{\tiny{$\pm$26}}} & \res{60{\tiny{$\pm$15}}}{\textbf{100}{\tiny{$\pm$0}}} & \res{0{\tiny{$\pm$0}}}{23{\tiny{$\pm$19}}} & \res{10{\tiny{$\pm$6}}}{71{\tiny{$\pm$25}}} & \res{1{\tiny{$\pm$9}}}{90{\tiny{$\pm$10}}} & \res{0{\tiny{$\pm$0}}}{48{\tiny{$\pm$17}}} & \res{0{\tiny{$\pm$0}}}{69{\tiny{$\pm$8}}} \\
    OGBench antmaze-giant-navigate-task2 & \res{0{\tiny{$\pm$0}}}{81{\tiny{$\pm$18}}} & \res{27{\tiny{$\pm$18}}}{\textbf{100}{\tiny{$\pm$1}}} & \res{0{\tiny{$\pm$0}}}{21{\tiny{$\pm$9}}} & \res{16{\tiny{$\pm$12}}}{\textbf{98}{\tiny{$\pm$1}}} & \res{0{\tiny{$\pm$0}}}{\textbf{99}{\tiny{$\pm$1}}} & \res{0{\tiny{$\pm$0}}}{83{\tiny{$\pm$18}}} & \res{0{\tiny{$\pm$0}}}{90{\tiny{$\pm$2}}} \\
    OGBench antmaze-giant-navigate-task3 & \res{0{\tiny{$\pm$0}}}{72{\tiny{$\pm$24}}} & \res{8{\tiny{$\pm$12}}}{\textbf{99}{\tiny{$\pm$1}}} & \res{0{\tiny{$\pm$0}}}{0{\tiny{$\pm$0}}} & \res{0{\tiny{$\pm$0}}}{9{\tiny{$\pm$14}}} & \res{0{\tiny{$\pm$0}}}{1{\tiny{$\pm$1}}} & \res{0{\tiny{$\pm$0}}}{6{\tiny{$\pm$4}}} & \res{0{\tiny{$\pm$0}}}{10{\tiny{$\pm$3}}} \\
    OGBench antmaze-giant-navigate-task4 & \res{0{\tiny{$\pm$0}}}{78{\tiny{$\pm$18}}} & \res{0{\tiny{$\pm$0}}}{\textbf{99}{\tiny{$\pm$1}}} & \res{0{\tiny{$\pm$0}}}{0{\tiny{$\pm$0}}} & \res{11{\tiny{$\pm$14}}}{90{\tiny{$\pm$16}}} & \res{0{\tiny{$\pm$0}}}{88{\tiny{$\pm$19}}} & \res{0{\tiny{$\pm$0}}}{71{\tiny{$\pm$20}}} & \res{0{\tiny{$\pm$0}}}{\textbf{97}{\tiny{$\pm$1}}} \\
    OGBench antmaze-giant-navigate-task5 & \res{0{\tiny{$\pm$0}}}{81{\tiny{$\pm$19}}} & \res{56{\tiny{$\pm$13}}}{\textbf{100}{\tiny{$\pm$1}}} & \res{2{\tiny{$\pm$2}}}{67{\tiny{$\pm$22}}} & \res{11{\tiny{$\pm$13}}}{\textbf{100}{\tiny{$\pm$0}}} & \res{7{\tiny{$\pm$8}}}{\textbf{100}{\tiny{$\pm$0}}} & \res{0{\tiny{$\pm$0}}}{28{\tiny{$\pm$28}}} & \res{3{\tiny{$\pm$5}}}{\textbf{97}{\tiny{$\pm$1}}} \\
    \midrule
    OGBench antsoccer-arena-navigate-task1 & \res{0{\tiny{$\pm$0}}}{\textbf{95}{\tiny{$\pm$2}}} & \res{0{\tiny{$\pm$0}}}{0{\tiny{$\pm$0}}} & \res{72{\tiny{$\pm$5}}}{65{\tiny{$\pm$27}}} & \res{39{\tiny{$\pm$6}}}{89{\tiny{$\pm$10}}} & \res{45{\tiny{$\pm$7}}}{90{\tiny{$\pm$4}}} & \res{22{\tiny{$\pm$6}}}{77{\tiny{$\pm$4}}} & \res{1{\tiny{$\pm$1}}}{\textbf{91}{\tiny{$\pm$2}}} \\
    OGBench antsoccer-arena-navigate-task2 & \res{0{\tiny{$\pm$0}}}{\textbf{94}{\tiny{$\pm$2}}} & \res{3{\tiny{$\pm$5}}}{0{\tiny{$\pm$0}}} & \res{61{\tiny{$\pm$4}}}{\textbf{91}{\tiny{$\pm$2}}} & \res{27{\tiny{$\pm$4}}}{73{\tiny{$\pm$14}}} & \res{32{\tiny{$\pm$7}}}{84{\tiny{$\pm$6}}} & \res{8{\tiny{$\pm$2}}}{77{\tiny{$\pm$4}}} & \res{0{\tiny{$\pm$1}}}{\textbf{91}{\tiny{$\pm$2}}} \\
    OGBench antsoccer-arena-navigate-task3 & \res{0{\tiny{$\pm$0}}}{59{\tiny{$\pm$12}}} & \res{0{\tiny{$\pm$0}}}{0{\tiny{$\pm$0}}} & \res{48{\tiny{$\pm$3}}}{\textbf{91}{\tiny{$\pm$3}}} & \res{43{\tiny{$\pm$5}}}{\textbf{93}{\tiny{$\pm$2}}} & \res{51{\tiny{$\pm$8}}}{87{\tiny{$\pm$9}}} & \res{10{\tiny{$\pm$3}}}{66{\tiny{$\pm$5}}} & \res{0{\tiny{$\pm$0}}}{84{\tiny{$\pm$5}}} \\
    OGBench antsoccer-arena-navigate-task4 & \res{0{\tiny{$\pm$0}}}{18{\tiny{$\pm$14}}} & \res{0{\tiny{$\pm$0}}}{0{\tiny{$\pm$0}}} & \res{36{\tiny{$\pm$6}}}{79{\tiny{$\pm$5}}} & \res{28{\tiny{$\pm$6}}}{\textbf{84}{\tiny{$\pm$4}}} & \res{39{\tiny{$\pm$5}}}{74{\tiny{$\pm$16}}} & \res{5{\tiny{$\pm$2}}}{60{\tiny{$\pm$7}}} & \res{0{\tiny{$\pm$0}}}{76{\tiny{$\pm$3}}} \\
    OGBench antsoccer-arena-navigate-task5 & \res{0{\tiny{$\pm$0}}}{55{\tiny{$\pm$6}}} & \res{0{\tiny{$\pm$0}}}{0{\tiny{$\pm$0}}} & \res{38{\tiny{$\pm$5}}}{46{\tiny{$\pm$24}}} & \res{36{\tiny{$\pm$7}}}{9{\tiny{$\pm$15}}} & \res{33{\tiny{$\pm$8}}}{22{\tiny{$\pm$13}}} & \res{3{\tiny{$\pm$2}}}{57{\tiny{$\pm$6}}} & \res{0{\tiny{$\pm$0}}}{\textbf{75}{\tiny{$\pm$3}}} \\
    \midrule
    OGBench cube-quadruple-play-task1-10M & \res{0{\tiny{$\pm$0}}}{0{\tiny{$\pm$0}}} & \res{0{\tiny{$\pm$0}}}{0{\tiny{$\pm$0}}} & \res{10{\tiny{$\pm$5}}}{48{\tiny{$\pm$11}}} & \res{1{\tiny{$\pm$1}}}{8{\tiny{$\pm$7}}} & \res{3{\tiny{$\pm$2}}}{34{\tiny{$\pm$14}}} & \res{4{\tiny{$\pm$3}}}{\textbf{93}{\tiny{$\pm$3}}} & \res{0{\tiny{$\pm$0}}}{\textbf{97}{\tiny{$\pm$2}}} \\
    OGBench cube-quadruple-play-task2-10M & \res{0{\tiny{$\pm$0}}}{0{\tiny{$\pm$0}}} & \res{0{\tiny{$\pm$0}}}{0{\tiny{$\pm$0}}} & \res{0{\tiny{$\pm$0}}}{0{\tiny{$\pm$0}}} & \res{0{\tiny{$\pm$0}}}{0{\tiny{$\pm$0}}} & \res{0{\tiny{$\pm$0}}}{0{\tiny{$\pm$0}}} & \res{0{\tiny{$\pm$0}}}{\textbf{55}{\tiny{$\pm$15}}} & \res{0{\tiny{$\pm$0}}}{50{\tiny{$\pm$21}}} \\
    OGBench cube-quadruple-play-task3-10M & \res{0{\tiny{$\pm$0}}}{0{\tiny{$\pm$0}}} & \res{0{\tiny{$\pm$0}}}{0{\tiny{$\pm$0}}} & \res{0{\tiny{$\pm$0}}}{0{\tiny{$\pm$0}}} & \res{0{\tiny{$\pm$0}}}{0{\tiny{$\pm$0}}} & \res{0{\tiny{$\pm$0}}}{0{\tiny{$\pm$0}}} & \res{0{\tiny{$\pm$0}}}{7{\tiny{$\pm$4}}} & \res{0{\tiny{$\pm$0}}}{\textbf{65}{\tiny{$\pm$13}}} \\
    OGBench cube-quadruple-play-task4-10M & \res{0{\tiny{$\pm$0}}}{0{\tiny{$\pm$0}}} & \res{0{\tiny{$\pm$0}}}{0{\tiny{$\pm$0}}} & \res{0{\tiny{$\pm$0}}}{0{\tiny{$\pm$0}}} & \res{0{\tiny{$\pm$0}}}{0{\tiny{$\pm$0}}} & \res{0{\tiny{$\pm$0}}}{0{\tiny{$\pm$0}}} & \res{0{\tiny{$\pm$0}}}{9{\tiny{$\pm$8}}} & \res{0{\tiny{$\pm$0}}}{\textbf{45}{\tiny{$\pm$20}}} \\
    OGBench cube-quadruple-play-task5-10M & \res{0{\tiny{$\pm$0}}}{0{\tiny{$\pm$0}}} & \res{0{\tiny{$\pm$0}}}{0{\tiny{$\pm$0}}} & \res{0{\tiny{$\pm$0}}}{0{\tiny{$\pm$0}}} & \res{0{\tiny{$\pm$0}}}{0{\tiny{$\pm$0}}} & \res{0{\tiny{$\pm$0}}}{0{\tiny{$\pm$0}}} & \res{0{\tiny{$\pm$0}}}{0{\tiny{$\pm$0}}} & \res{0{\tiny{$\pm$0}}}{0{\tiny{$\pm$0}}} \\
    \midrule
    OGBench cube-triple-play-task1 & \res{0{\tiny{$\pm$0}}}{2{\tiny{$\pm$1}}} & \res{0{\tiny{$\pm$0}}}{0{\tiny{$\pm$0}}} & \res{16{\tiny{$\pm$5}}}{89{\tiny{$\pm$9}}} & \res{9{\tiny{$\pm$4}}}{87{\tiny{$\pm$7}}} & \res{6{\tiny{$\pm$3}}}{93{\tiny{$\pm$9}}} & \res{19{\tiny{$\pm$8}}}{\textbf{100}{\tiny{$\pm$1}}} & \res{3{\tiny{$\pm$3}}}{\textbf{100}{\tiny{$\pm$0}}} \\
    OGBench cube-triple-play-task2 & \res{0{\tiny{$\pm$0}}}{0{\tiny{$\pm$0}}} & \res{0{\tiny{$\pm$0}}}{0{\tiny{$\pm$0}}} & \res{0{\tiny{$\pm$1}}}{5{\tiny{$\pm$4}}} & \res{0{\tiny{$\pm$0}}}{0{\tiny{$\pm$0}}} & \res{0{\tiny{$\pm$0}}}{6{\tiny{$\pm$5}}} & \res{0{\tiny{$\pm$0}}}{\textbf{81}{\tiny{$\pm$18}}} & \res{0{\tiny{$\pm$0}}}{60{\tiny{$\pm$14}}} \\
    OGBench cube-triple-play-task3 & \res{0{\tiny{$\pm$0}}}{0{\tiny{$\pm$0}}} & \res{0{\tiny{$\pm$0}}}{0{\tiny{$\pm$0}}} & \res{2{\tiny{$\pm$2}}}{17{\tiny{$\pm$8}}} & \res{0{\tiny{$\pm$0}}}{6{\tiny{$\pm$6}}} & \res{2{\tiny{$\pm$2}}}{20{\tiny{$\pm$9}}} & \res{0{\tiny{$\pm$0}}}{68{\tiny{$\pm$7}}} & \res{0{\tiny{$\pm$0}}}{\textbf{94}{\tiny{$\pm$4}}} \\
    OGBench cube-triple-play-task4 & \res{0{\tiny{$\pm$0}}}{0{\tiny{$\pm$0}}} & \res{0{\tiny{$\pm$0}}}{0{\tiny{$\pm$0}}} & \res{0{\tiny{$\pm$0}}}{0{\tiny{$\pm$1}}} & \res{0{\tiny{$\pm$0}}}{0{\tiny{$\pm$0}}} & \res{0{\tiny{$\pm$0}}}{0{\tiny{$\pm$0}}} & \res{0{\tiny{$\pm$0}}}{36{\tiny{$\pm$5}}} & \res{0{\tiny{$\pm$0}}}{\textbf{81}{\tiny{$\pm$7}}} \\
    OGBench cube-triple-play-task5 & \res{0{\tiny{$\pm$0}}}{0{\tiny{$\pm$0}}} & \res{0{\tiny{$\pm$0}}}{0{\tiny{$\pm$0}}} & \res{0{\tiny{$\pm$0}}}{0{\tiny{$\pm$0}}} & \res{0{\tiny{$\pm$0}}}{0{\tiny{$\pm$0}}} & \res{0{\tiny{$\pm$0}}}{0{\tiny{$\pm$0}}} & \res{0{\tiny{$\pm$0}}}{0{\tiny{$\pm$0}}} & \res{0{\tiny{$\pm$0}}}{0{\tiny{$\pm$0}}} \\
    \midrule
    OGBench humanoidmaze-medium-navigate-task1 & \res{0{\tiny{$\pm$0}}}{28{\tiny{$\pm$19}}} & \res{25{\tiny{$\pm$9}}}{1{\tiny{$\pm$2}}} & \res{55{\tiny{$\pm$5}}}{\textbf{97}{\tiny{$\pm$2}}} & \res{14{\tiny{$\pm$5}}}{18{\tiny{$\pm$6}}} & \res{15{\tiny{$\pm$4}}}{91{\tiny{$\pm$1}}} & \res{0{\tiny{$\pm$0}}}{30{\tiny{$\pm$8}}} & \res{37{\tiny{$\pm$9}}}{89{\tiny{$\pm$3}}} \\
    OGBench humanoidmaze-medium-navigate-task2 & \res{0{\tiny{$\pm$0}}}{49{\tiny{$\pm$28}}} & \res{43{\tiny{$\pm$9}}}{1{\tiny{$\pm$1}}} & \res{89{\tiny{$\pm$4}}}{84{\tiny{$\pm$19}}} & \res{51{\tiny{$\pm$6}}}{91{\tiny{$\pm$2}}} & \res{52{\tiny{$\pm$2}}}{\textbf{99}{\tiny{$\pm$1}}} & \res{7{\tiny{$\pm$2}}}{49{\tiny{$\pm$12}}} & \res{57{\tiny{$\pm$4}}}{\textbf{99}{\tiny{$\pm$1}}} \\
    OGBench humanoidmaze-medium-navigate-task3 & \res{0{\tiny{$\pm$0}}}{7{\tiny{$\pm$13}}} & \res{61{\tiny{$\pm$13}}}{6{\tiny{$\pm$7}}} & \res{54{\tiny{$\pm$11}}}{\textbf{98}{\tiny{$\pm$2}}} & \res{48{\tiny{$\pm$5}}}{69{\tiny{$\pm$13}}} & \res{52{\tiny{$\pm$12}}}{88{\tiny{$\pm$18}}} & \res{35{\tiny{$\pm$7}}}{7{\tiny{$\pm$12}}} & \res{58{\tiny{$\pm$4}}}{\textbf{95}{\tiny{$\pm$2}}} \\
    OGBench humanoidmaze-medium-navigate-task4 & \res{0{\tiny{$\pm$0}}}{0{\tiny{$\pm$0}}} & \res{29{\tiny{$\pm$12}}}{0{\tiny{$\pm$0}}} & \res{4{\tiny{$\pm$7}}}{55{\tiny{$\pm$26}}} & \res{8{\tiny{$\pm$3}}}{10{\tiny{$\pm$4}}} & \res{13{\tiny{$\pm$20}}}{\textbf{85}{\tiny{$\pm$10}}} & \res{0{\tiny{$\pm$0}}}{0{\tiny{$\pm$1}}} & \res{3{\tiny{$\pm$2}}}{30{\tiny{$\pm$4}}} \\
    OGBench humanoidmaze-medium-navigate-task5 & \res{0{\tiny{$\pm$0}}}{63{\tiny{$\pm$27}}} & \res{11{\tiny{$\pm$6}}}{13{\tiny{$\pm$9}}} & \res{96{\tiny{$\pm$3}}}{\textbf{99}{\tiny{$\pm$1}}} & \res{58{\tiny{$\pm$8}}}{93{\tiny{$\pm$4}}} & \res{57{\tiny{$\pm$1}}}{\textbf{99}{\tiny{$\pm$1}}} & \res{31{\tiny{$\pm$5}}}{76{\tiny{$\pm$10}}} & \res{71{\tiny{$\pm$5}}}{\textbf{99}{\tiny{$\pm$1}}} \\
    \midrule
    OGBench puzzle-4x4-play-task1 & \res{0{\tiny{$\pm$0}}}{\textbf{100}{\tiny{$\pm$0}}} & \res{27{\tiny{$\pm$6}}}{\textbf{100}{\tiny{$\pm$0}}} & \res{59{\tiny{$\pm$4}}}{\textbf{100}{\tiny{$\pm$0}}} & \res{30{\tiny{$\pm$6}}}{\textbf{100}{\tiny{$\pm$0}}} & \res{16{\tiny{$\pm$9}}}{\textbf{100}{\tiny{$\pm$0}}} & \res{36{\tiny{$\pm$8}}}{\textbf{100}{\tiny{$\pm$0}}} & \res{52{\tiny{$\pm$11}}}{\textbf{100}{\tiny{$\pm$0}}} \\
    OGBench puzzle-4x4-play-task2 & \res{0{\tiny{$\pm$0}}}{40{\tiny{$\pm$32}}} & \res{9{\tiny{$\pm$2}}}{0{\tiny{$\pm$0}}} & \res{19{\tiny{$\pm$2}}}{9{\tiny{$\pm$14}}} & \res{13{\tiny{$\pm$3}}}{0{\tiny{$\pm$0}}} & \res{11{\tiny{$\pm$4}}}{0{\tiny{$\pm$0}}} & \res{0{\tiny{$\pm$0}}}{0{\tiny{$\pm$0}}} & \res{0{\tiny{$\pm$0}}}{\textbf{70}{\tiny{$\pm$30}}} \\
    OGBench puzzle-4x4-play-task3 & \res{0{\tiny{$\pm$0}}}{80{\tiny{$\pm$26}}} & \res{16{\tiny{$\pm$3}}}{61{\tiny{$\pm$29}}} & \res{40{\tiny{$\pm$5}}}{\textbf{95}{\tiny{$\pm$5}}} & \res{15{\tiny{$\pm$3}}}{\textbf{99}{\tiny{$\pm$1}}} & \res{20{\tiny{$\pm$8}}}{87{\tiny{$\pm$20}}} & \res{16{\tiny{$\pm$4}}}{90{\tiny{$\pm$20}}} & \res{13{\tiny{$\pm$7}}}{\textbf{100}{\tiny{$\pm$0}}} \\
    OGBench puzzle-4x4-play-task4 & \res{0{\tiny{$\pm$0}}}{70{\tiny{$\pm$30}}} & \res{11{\tiny{$\pm$2}}}{0{\tiny{$\pm$1}}} & \res{25{\tiny{$\pm$3}}}{78{\tiny{$\pm$21}}} & \res{7{\tiny{$\pm$2}}}{70{\tiny{$\pm$30}}} & \res{18{\tiny{$\pm$5}}}{80{\tiny{$\pm$26}}} & \res{1{\tiny{$\pm$1}}}{0{\tiny{$\pm$0}}} & \res{0{\tiny{$\pm$0}}}{\textbf{100}{\tiny{$\pm$0}}} \\
    OGBench puzzle-4x4-play-task5 & \res{0{\tiny{$\pm$0}}}{30{\tiny{$\pm$30}}} & \res{9{\tiny{$\pm$3}}}{0{\tiny{$\pm$0}}} & \res{18{\tiny{$\pm$5}}}{20{\tiny{$\pm$26}}} & \res{5{\tiny{$\pm$2}}}{0{\tiny{$\pm$0}}} & \res{11{\tiny{$\pm$6}}}{0{\tiny{$\pm$0}}} & \res{0{\tiny{$\pm$0}}}{0{\tiny{$\pm$0}}} & \res{0{\tiny{$\pm$0}}}{\textbf{80}{\tiny{$\pm$26}}} \\
    \midrule
    robomimic lift-mh & \res{0{\tiny{$\pm$0}}}{\textbf{100}{\tiny{$\pm$1}}} & \res{0{\tiny{$\pm$0}}}{60{\tiny{$\pm$25}}} & \res{90{\tiny{$\pm$3}}}{\textbf{98}{\tiny{$\pm$1}}} & \res{83{\tiny{$\pm$5}}}{94{\tiny{$\pm$3}}} & \res{87{\tiny{$\pm$3}}}{\textbf{99}{\tiny{$\pm$1}}} & \res{97{\tiny{$\pm$2}}}{\textbf{100}{\tiny{$\pm$0}}} & \res{96{\tiny{$\pm$1}}}{\textbf{100}{\tiny{$\pm$0}}} \\
    robomimic can-mh & \res{0{\tiny{$\pm$0}}}{70{\tiny{$\pm$5}}} & \res{0{\tiny{$\pm$0}}}{\textbf{97}{\tiny{$\pm$2}}} & \res{64{\tiny{$\pm$6}}}{80{\tiny{$\pm$4}}} & \res{39{\tiny{$\pm$7}}}{71{\tiny{$\pm$6}}} & \res{53{\tiny{$\pm$9}}}{88{\tiny{$\pm$5}}} & \res{89{\tiny{$\pm$3}}}{93{\tiny{$\pm$2}}} & \res{88{\tiny{$\pm$2}}}{\textbf{99}{\tiny{$\pm$1}}} \\
    robomimic square-mh & \res{0{\tiny{$\pm$0}}}{72{\tiny{$\pm$3}}} & \res{0{\tiny{$\pm$0}}}{14{\tiny{$\pm$5}}} & \res{14{\tiny{$\pm$2}}}{28{\tiny{$\pm$3}}} & \res{4{\tiny{$\pm$2}}}{13{\tiny{$\pm$3}}} & \res{8{\tiny{$\pm$3}}}{29{\tiny{$\pm$4}}} & \res{35{\tiny{$\pm$5}}}{81{\tiny{$\pm$3}}} & \res{42{\tiny{$\pm$5}}}{\textbf{94}{\tiny{$\pm$2}}} \\
    robomimic transport-mh & \res{0{\tiny{$\pm$0}}}{0{\tiny{$\pm$0}}} & \res{0{\tiny{$\pm$0}}}{0{\tiny{$\pm$0}}} & \res{1{\tiny{$\pm$1}}}{1{\tiny{$\pm$1}}} & \res{0{\tiny{$\pm$0}}}{0{\tiny{$\pm$0}}} & \res{0{\tiny{$\pm$0}}}{0{\tiny{$\pm$0}}} & \res{13{\tiny{$\pm$4}}}{22{\tiny{$\pm$6}}} & \res{0{\tiny{$\pm$0}}}{\textbf{39}{\tiny{$\pm$4}}} \\
    \bottomrule
    \end{tabular}%
    }
    \label{tab:full_result_main}
\end{table}

\newpage
\begin{table}[h!]
    \centering
    \footnotesize
    \vskip 0.1in
    \renewcommand{\arraystretch}{1.1}
    \caption{Complete results for Figure \ref{fig:ablation}, showing the performance immediately after 1M steps of offline learning and 1M steps of online learning. Results represent the mean $\pm$ 95\% confidence interval calculated across 10 seeds.}
    \vspace{2ex}
    \resizebox{\textwidth}{!}{%
    \begin{tabular}{l|*{4}{r@{\,$\rightarrow$\,}l}}
    \toprule
    Environment
      & \multicolumn{2}{c}{Random}
      & \multicolumn{2}{c}{Uniform}
      & \multicolumn{2}{c}{Greedy}
      & \multicolumn{2}{c}{ACH}\\
    \midrule
    OGBench antmaze-giant-navigate-task1 & \res{0{\tiny{$\pm$0}}}{25{\tiny{$\pm$18}}} & \res{0{\tiny{$\pm$0}}}{26{\tiny{$\pm$8}}} & \res{0{\tiny{$\pm$0}}}{\textbf{77}{\tiny{$\pm$7}}} & \res{0{\tiny{$\pm$0}}}{69{\tiny{$\pm$8}}} \\
    OGBench antmaze-giant-navigate-task2 & \res{0{\tiny{$\pm$0}}}{68{\tiny{$\pm$18}}} & \res{0{\tiny{$\pm$0}}}{\textbf{88}{\tiny{$\pm$2}}} & \res{0{\tiny{$\pm$0}}}{\textbf{86}{\tiny{$\pm$4}}} & \res{0{\tiny{$\pm$0}}}{\textbf{90}{\tiny{$\pm$2}}} \\
    OGBench antmaze-giant-navigate-task3 & \res{0{\tiny{$\pm$0}}}{13{\tiny{$\pm$8}}} & \res{0{\tiny{$\pm$0}}}{2{\tiny{$\pm$1}}} & \res{0{\tiny{$\pm$0}}}{\textbf{40}{\tiny{$\pm$7}}} & \res{0{\tiny{$\pm$0}}}{10{\tiny{$\pm$3}}} \\
    OGBench antmaze-giant-navigate-task4 & \res{0{\tiny{$\pm$0}}}{45{\tiny{$\pm$25}}} & \res{0{\tiny{$\pm$0}}}{91{\tiny{$\pm$3}}} & \res{0{\tiny{$\pm$0}}}{\textbf{92}{\tiny{$\pm$3}}} & \res{0{\tiny{$\pm$0}}}{\textbf{97}{\tiny{$\pm$1}}} \\
    OGBench antmaze-giant-navigate-task5 & \res{0{\tiny{$\pm$0}}}{27{\tiny{$\pm$26}}} & \res{4{\tiny{$\pm$3}}}{91{\tiny{$\pm$4}}} & \res{6{\tiny{$\pm$5}}}{\textbf{94}{\tiny{$\pm$3}}} & \res{3{\tiny{$\pm$5}}}{\textbf{97}{\tiny{$\pm$1}}} \\
    \midrule
    OGBench antsoccer-arena-navigate-task1 & \res{23{\tiny{$\pm$6}}}{70{\tiny{$\pm$6}}} & \res{1{\tiny{$\pm$1}}}{83{\tiny{$\pm$4}}} & \res{1{\tiny{$\pm$1}}}{\textbf{89}{\tiny{$\pm$3}}} & \res{1{\tiny{$\pm$1}}}{\textbf{91}{\tiny{$\pm$2}}} \\
    OGBench antsoccer-arena-navigate-task2 & \res{7{\tiny{$\pm$2}}}{73{\tiny{$\pm$4}}} & \res{0{\tiny{$\pm$0}}}{80{\tiny{$\pm$3}}} & \res{0{\tiny{$\pm$0}}}{\textbf{90}{\tiny{$\pm$3}}} & \res{0{\tiny{$\pm$1}}}{\textbf{91}{\tiny{$\pm$2}}} \\
    OGBench antsoccer-arena-navigate-task3 & \res{1{\tiny{$\pm$1}}}{23{\tiny{$\pm$7}}} & \res{0{\tiny{$\pm$0}}}{73{\tiny{$\pm$4}}} & \res{0{\tiny{$\pm$1}}}{78{\tiny{$\pm$10}}} & \res{0{\tiny{$\pm$0}}}{\textbf{84}{\tiny{$\pm$5}}} \\
    OGBench antsoccer-arena-navigate-task4 & \res{6{\tiny{$\pm$2}}}{55{\tiny{$\pm$7}}} & \res{0{\tiny{$\pm$0}}}{63{\tiny{$\pm$4}}} & \res{0{\tiny{$\pm$0}}}{\textbf{75}{\tiny{$\pm$3}}} & \res{0{\tiny{$\pm$0}}}{\textbf{76}{\tiny{$\pm$3}}} \\
    OGBench antsoccer-arena-navigate-task5 & \res{4{\tiny{$\pm$2}}}{55{\tiny{$\pm$6}}} & \res{0{\tiny{$\pm$0}}}{67{\tiny{$\pm$4}}} & \res{0{\tiny{$\pm$0}}}{\textbf{72}{\tiny{$\pm$6}}} & \res{0{\tiny{$\pm$0}}}{\textbf{75}{\tiny{$\pm$3}}} \\
    \midrule
    OGBench cube-quadruple-play-10M-task1 & \res{2{\tiny{$\pm$1}}}{95{\tiny{$\pm$3}}} & \res{0{\tiny{$\pm$0}}}{13{\tiny{$\pm$20}}} & \res{0{\tiny{$\pm$0}}}{36{\tiny{$\pm$29}}} & \res{0{\tiny{$\pm$0}}}{\textbf{97}{\tiny{$\pm$2}}} \\
    OGBench cube-quadruple-play-10M-task2 & \res{0{\tiny{$\pm$0}}}{55{\tiny{$\pm$14}}} & \res{0{\tiny{$\pm$0}}}{1{\tiny{$\pm$2}}} & \res{0{\tiny{$\pm$0}}}{0{\tiny{$\pm$0}}} & \res{0{\tiny{$\pm$0}}}{\textbf{50}{\tiny{$\pm$21}}} \\
    OGBench cube-quadruple-play-10M-task3 & \res{0{\tiny{$\pm$0}}}{7{\tiny{$\pm$4}}} & \res{0{\tiny{$\pm$0}}}{31{\tiny{$\pm$20}}} & \res{0{\tiny{$\pm$0}}}{27{\tiny{$\pm$16}}} & \res{0{\tiny{$\pm$0}}}{\textbf{65}{\tiny{$\pm$13}}} \\
    OGBench cube-quadruple-play-10M-task4 & \res{0{\tiny{$\pm$0}}}{6{\tiny{$\pm$5}}} & \res{0{\tiny{$\pm$0}}}{0{\tiny{$\pm$0}}} & \res{0{\tiny{$\pm$0}}}{14{\tiny{$\pm$17}}} & \res{0{\tiny{$\pm$0}}}{\textbf{45}{\tiny{$\pm$20}}} \\
    OGBench cube-quadruple-play-10M-task5 & \res{0{\tiny{$\pm$0}}}{0{\tiny{$\pm$0}}} & \res{0{\tiny{$\pm$0}}}{0{\tiny{$\pm$0}}} & \res{0{\tiny{$\pm$0}}}{0{\tiny{$\pm$0}}} & \res{0{\tiny{$\pm$0}}}{0{\tiny{$\pm$0}}} \\
    \midrule
    OGBench cube-triple-play-task1 & \res{14{\tiny{$\pm$6}}}{\textbf{100}{\tiny{$\pm$0}}} & \res{4{\tiny{$\pm$4}}}{7{\tiny{$\pm$8}}} & \res{2{\tiny{$\pm$2}}}{92{\tiny{$\pm$12}}} & \res{3{\tiny{$\pm$3}}}{\textbf{100}{\tiny{$\pm$0}}} \\
    OGBench cube-triple-play-task2 & \res{0{\tiny{$\pm$0}}}{\textbf{75}{\tiny{$\pm$17}}} & \res{0{\tiny{$\pm$0}}}{15{\tiny{$\pm$19}}} & \res{0{\tiny{$\pm$0}}}{44{\tiny{$\pm$29}}} & \res{0{\tiny{$\pm$0}}}{60{\tiny{$\pm$14}}} \\
    OGBench cube-triple-play-task3 & \res{1{\tiny{$\pm$1}}}{52{\tiny{$\pm$11}}} & \res{1{\tiny{$\pm$1}}}{80{\tiny{$\pm$19}}} & \res{0{\tiny{$\pm$0}}}{81{\tiny{$\pm$13}}} & \res{0{\tiny{$\pm$0}}}{\textbf{94}{\tiny{$\pm$4}}} \\
    OGBench cube-triple-play-task4 & \res{0{\tiny{$\pm$0}}}{27{\tiny{$\pm$10}}} & \res{0{\tiny{$\pm$0}}}{20{\tiny{$\pm$25}}} & \res{0{\tiny{$\pm$0}}}{50{\tiny{$\pm$18}}} & \res{0{\tiny{$\pm$0}}}{\textbf{81}{\tiny{$\pm$7}}} \\
    OGBench cube-triple-play-task5 & \res{0{\tiny{$\pm$0}}}{0{\tiny{$\pm$0}}} & \res{0{\tiny{$\pm$0}}}{0{\tiny{$\pm$0}}} & \res{0{\tiny{$\pm$0}}}{0{\tiny{$\pm$0}}} & \res{0{\tiny{$\pm$0}}}{0{\tiny{$\pm$0}}} \\
    \midrule
    OGBench humanoidmaze-medium-navigate-task1 & \res{0{\tiny{$\pm$0}}}{30{\tiny{$\pm$12}}} & \res{49{\tiny{$\pm$5}}}{31{\tiny{$\pm$2}}} & \res{38{\tiny{$\pm$5}}}{\textbf{88}{\tiny{$\pm$6}}} & \res{37{\tiny{$\pm$9}}}{\textbf{89}{\tiny{$\pm$3}}} \\
    OGBench humanoidmaze-medium-navigate-task2 & \res{8{\tiny{$\pm$2}}}{54{\tiny{$\pm$5}}} & \res{52{\tiny{$\pm$4}}}{78{\tiny{$\pm$7}}} & \res{63{\tiny{$\pm$3}}}{\textbf{94}{\tiny{$\pm$3}}} & \res{57{\tiny{$\pm$4}}}{\textbf{99}{\tiny{$\pm$1}}} \\
    OGBench humanoidmaze-medium-navigate-task3 & \res{23{\tiny{$\pm$10}}}{9{\tiny{$\pm$16}}} & \res{42{\tiny{$\pm$7}}}{50{\tiny{$\pm$5}}} & \res{51{\tiny{$\pm$5}}}{\textbf{89}{\tiny{$\pm$4}}} & \res{58{\tiny{$\pm$4}}}{\textbf{95}{\tiny{$\pm$2}}} \\
    OGBench humanoidmaze-medium-navigate-task4 & \res{0{\tiny{$\pm$0}}}{1{\tiny{$\pm$1}}} & \res{6{\tiny{$\pm$3}}}{5{\tiny{$\pm$2}}} & \res{3{\tiny{$\pm$4}}}{22{\tiny{$\pm$4}}} & \res{3{\tiny{$\pm$2}}}{\textbf{30}{\tiny{$\pm$4}}} \\
    OGBench humanoidmaze-medium-navigate-task5 & \res{29{\tiny{$\pm$6}}}{85{\tiny{$\pm$6}}} & \res{67{\tiny{$\pm$8}}}{82{\tiny{$\pm$7}}} & \res{70{\tiny{$\pm$4}}}{\textbf{96}{\tiny{$\pm$1}}} & \res{71{\tiny{$\pm$5}}}{\textbf{99}{\tiny{$\pm$1}}} \\
    \midrule
    OGBench puzzle-4x4-play-task1 & \res{35{\tiny{$\pm$7}}}{\textbf{99}{\tiny{$\pm$1}}} & \res{63{\tiny{$\pm$15}}}{\textbf{100}{\tiny{$\pm$0}}} & \res{49{\tiny{$\pm$6}}}{\textbf{100}{\tiny{$\pm$0}}} & \res{52{\tiny{$\pm$11}}}{\textbf{100}{\tiny{$\pm$0}}} \\
    OGBench puzzle-4x4-play-task2 & \res{0{\tiny{$\pm$0}}}{20{\tiny{$\pm$26}}} & \res{0{\tiny{$\pm$0}}}{0{\tiny{$\pm$0}}} & \res{0{\tiny{$\pm$0}}}{20{\tiny{$\pm$26}}} & \res{0{\tiny{$\pm$0}}}{\textbf{70}{\tiny{$\pm$30}}} \\
    OGBench puzzle-4x4-play-task3 & \res{14{\tiny{$\pm$7}}}{80{\tiny{$\pm$26}}} & \res{28{\tiny{$\pm$11}}}{50{\tiny{$\pm$33}}} & \res{9{\tiny{$\pm$4}}}{\textbf{100}{\tiny{$\pm$0}}} & \res{13{\tiny{$\pm$7}}}{\textbf{100}{\tiny{$\pm$0}}} \\
    OGBench puzzle-4x4-play-task4 & \res{0{\tiny{$\pm$0}}}{10{\tiny{$\pm$20}}} & \res{1{\tiny{$\pm$1}}}{30{\tiny{$\pm$30}}} & \res{0{\tiny{$\pm$1}}}{90{\tiny{$\pm$20}}} & \res{0{\tiny{$\pm$0}}}{\textbf{100}{\tiny{$\pm$0}}} \\
    OGBench puzzle-4x4-play-task5 & \res{0{\tiny{$\pm$0}}}{0{\tiny{$\pm$0}}} & \res{0{\tiny{$\pm$0}}}{0{\tiny{$\pm$0}}} & \res{0{\tiny{$\pm$0}}}{20{\tiny{$\pm$26}}} & \res{0{\tiny{$\pm$0}}}{\textbf{80}{\tiny{$\pm$26}}} \\
    \bottomrule
    \end{tabular}%
    }
    \label{tab:full_result_ablation}
\end{table}
\end{document}

%% file: math_commands.tex
\usepackage{amsmath,amsfonts,bm}

\def\eqref#1{equation~\ref{#1}}

\def\1{\bm{1}}

\def\va{{\bm{a}}}

\DeclareMathAlphabet{\mathsfit}{\encodingdefault}{\sfdefault}{m}{sl}
\SetMathAlphabet{\mathsfit}{bold}{\encodingdefault}{\sfdefault}{bx}{n}

\newcommand{\R}{\mathbb{R}}